\documentclass{article}

\usepackage{arxiv}

\usepackage{graphicx} 
\usepackage{subcaption} 
\usepackage{amsmath}
\renewcommand{\vec}[1]{\boldsymbol{#1}}
\usepackage{amssymb}
\usepackage{algorithm}
\usepackage{algpseudocode}
\usepackage[dvipsnames,table,xcdraw]{xcolor}
\usepackage{hyperref}
\usepackage[normalem]{ulem}
\usepackage{comment}
\usepackage{biblatex}

\usepackage{siunitx}
\usepackage{pifont} 

\usepackage{array} 
\usepackage{tabularx}
\newcolumntype{Y}{>{\centering\arraybackslash\hsize=\hsize}X}

\newcommand{\cmark}{\ding{51}} 
\newcommand{\xmark}{\ding{55}} 

\newif\ifrevision
\revisionfalse

\usepackage{multirow} 

\usepackage{pgfplots} 
\usepgfplotslibrary{groupplots,fillbetween}
\usepackage{pgfplotstable}
\pgfplotsset{compat=1.16}

\definecolor{lightblue}{RGB}{173, 216, 230}

\NewDocumentCommand{\orcid}{m}{%
  \href{https://orcid.org/#1}{%
    \includegraphics[height=1.6ex]{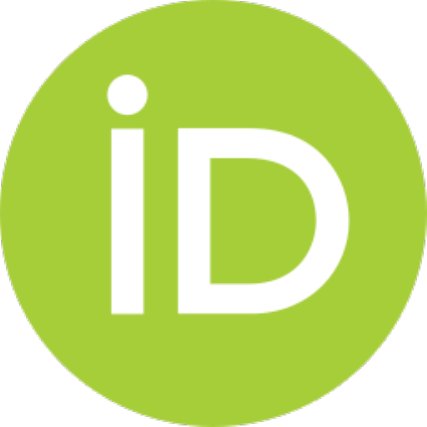}%
  }%
}

\NewDocumentCommand{\mail}{m}{%
  \href{mailto:#1}{\texttt{#1}}%
}

\title{DISCERN: Distributional Synthetic Replay for Retain-Free Incremental Machine Unlearning}

\author{\orcid{0009-0002-1413-2277}
    Nicolás Segura-Kunsagi$^{*,a}$ \\
    \mail{nkunsagi@ugr.es}
    \And
    \orcid{0000-0001-8587-4345}
    Juan Luis Suárez-Díaz$^{a}$ \\
    \mail{jlsuarezdiaz@ugr.es}
    \AND
    \orcid{0000-0002-5029-9106}
    Iván Sevillano-García$^{a}$ \\
    \mail{isevillano@ugr.es}
    \And
    \orcid{0000-0002-7283-312X}
    Francisco Herrera$^{a}$ \\
    \mail{herrera@decsai.ugr.es}
}

\newcommand{\affiliations}{
  \vspace{-3em}
  \begin{center}
    \textsuperscript{*} Corresponding author \\
    \textsuperscript{a} \href{https://dasci.es}{Department of Computer Science and Artificial Intelligence (DECSAI), Andalusian Institute of Data Science and Computational Intelligence (DaSCI),  University of Granada, Granada, Spain}
    \vspace{3.5em}
  \end{center}
}

\begin{document}

\maketitle
\affiliations
\begin{abstract}
Machine unlearning aims to remove the influence of specific training data from an already trained model without the cost of retraining it from scratch. In real-world deployments, however, deletion requests rarely arrive only once: regulatory changes and evolving content policies call for incremental unlearning, in which classes are removed sequentially, previously forgotten knowledge must remain forgotten, and the remaining classes must be preserved. In this work, we present DIstributional SynthetiC rEplay for Retain-free iNcremental unlearning (DISCERN), an intervention-based class unlearning method that replaces access to the retain set with synthetic embeddings sampled from per-class parametric distributions. Unlike prototype-based replay, which assumes an isotropic Gaussian around each class centroid, DISCERN fits Euclidean (Gaussian, Laplace, Student's $t$) and directional (von Mises--Fisher, Fisher--Bingham) families to each class and selects among them through statistical tests and, when none of them accepts, cross-validated log-likelihood. This design is motivated by the observation that latent geometry is strongly architecture-dependent: transformer embeddings lie approximately on a hypersphere and concentrate around class directions, and unlearning methods consistently concentrate the representation of the forgotten class in transformers while dispersing it in convolutional networks. We evaluate DISCERN against seven retain-free baselines on MNIST, CIFAR-10, and CIFAR-100 with two convolutional and two transformer backbones, under single-class, multi-class, and incremental class unlearning (ICU).
The results show that the benefit of explicitly modeling the embedding distribution is not uniform across settings but emerges mainly in ICU, where forgetting must be sustained along a sequence. DISCERN is competitive when classes are removed at once and is the most stable method in ICU on transformers: on ViT-B/16, after 20 sequential deletions it retains a harmonic mean of $0.797$, whereas the strongest method in the simultaneous setting drops from $0.918$ to $0.048$. A paired ablation shows that per-class distribution selection substantially reduces the resurfacing of previously forgotten classes along the incremental sequence, suggesting that it acts as a stabilizer of accumulated forgetting rather than as a uniform improvement in isolated deletion.
\end{abstract}

\section{Introduction}

Modern deep neural networks are trained on, or have access to, massive amounts of data. These data may be collected from users or obtained through unfiltered, indiscriminate web scraping, which can give rise to several problems. In the former case, users may exercise their ``right to be forgotten'' \cite{voigt2017eu}; in the latter, the data may be poisoned, degrading model performance or introducing backdoors \cite{northcutt2021pervasive, roychowdhury2024confusedpilot}. Techniques for removing the effect of specific subsets of the training data from a model, known as machine unlearning (MU) \cite{cao2015towards}, are therefore of vital importance. While existing MU methods have been designed with static unlearning in mind, that is, forgetting a small set of classes or data in a single step, real-world scenarios require incremental unlearning (IU) \cite{zhao2024continual}. Changes in regulations (e.g., the EU AI Act) and in restricted content call for the ability to forget data or concepts dynamically while avoiding the high cost of retraining \cite{bourtoule2021machine}. IU seeks not only to remove the influence of some data in a single step, but also to preserve the original performance of the model on the data to be retained and to ensure that previously performed forgetting persists. IU methods are mostly applied in two scenarios: iteratively removing sensitive concepts (nudity, violence, etc.) from generative models \cite{Gandikota_2023_ICCV}, and removing instances or classes from classification models fine-tuned from pretrained backbones \cite{zhao2024continual}. 

Effective MU methods exist when the model is trained with future unlearning in mind \cite{wu2020deltagrad, ALMAHMUD2025107879}, but they are not applicable to models trained in the standard way. We are therefore interested in post hoc methods, i.e., methods applicable to any already-trained network. In MU, the training set is split into two subsets: the data to be forgotten (forget set) and the data to be retained (retain set). Unlearning techniques typically employ one loss function per subset, and they can be divided into two types: retain-free methods, which may use at most statistics extracted from the dataset and do not access the retain set during unlearning \cite{zhang2025toward}, and methods with partial or full access to the retain set \cite{10113700}. Within retain-free methods, we further distinguish intervention methods, which compute statistics on the dataset before unlearning \cite{foster2024fast}, from those that do not \cite{ijcai2025p39}. Most of these methods, however, are designed and evaluated for a single deletion request. This work addresses the second IU scenario described above, incremental class unlearning (ICU), in which classes are removed one at a time from image classification models. ICU is not a mere repetition of class unlearning: a method may correctly remove a class in an isolated step and yet progressively degrade the retained knowledge, or even recover information that had been removed in earlier steps. Our objectives are therefore threefold: to forget the target class at every step, to keep previously forgotten classes forgotten, and to maintain performance on the remaining classes. The challenge addressed in this work is to meet these three objectives jointly when the retain set is unavailable during unlearning. Our hypothesis is that a parametric representation of the geometry of each class can provide a replay mechanism informative enough to stabilize the process without storing the original samples.

Since models are not always prepared to facilitate forgetting, and access to the retain set may be unavailable when a deletion request arrives \cite{ahmed2025towards}, this work aims to develop a retain-free unlearning method for class-based tasks. To this end, we frame ICU within continual learning (CL) \cite{wang2024comprehensive}: a model that has learned a first task and whose subsequent tasks consist of forgetting part of it. In CL, replaying data from previous tasks has proven particularly beneficial for preserving prior knowledge \cite{rolnick2019experience}. Since such data are not always available, an alternative is to use synthetic data generated from class prototypes. Prototype Augmentation and Self-Supervision (PASS) \cite{zhu2021prototype} follows this philosophy: when a new class is incorporated, its centroid in the representation space is computed, and in subsequent tasks synthetic samples of that class are drawn from a Gaussian distribution whose mean is the centroid and whose covariance is the identity matrix.

Inspired by PASS, we propose a retain-free, intervention-based MU method designed to be effective in ICU. The proposed method, DIstributional SynthetiC rEplay for Retain-free iNcremental unlearning (DISCERN), not only adapts PASS but extends it by considering several families of parametric distributions and selecting, for each class, the one that best fits its data.

Moving beyond a single Gaussian template is not a matter of fine-tuning PASS, but a response to how modern backbones organize their latent space. Transformer architectures are increasingly the backbone of choice in vision, and their embedding geometry departs markedly from the Gaussian assumption. In both transformers we evaluate, the embeddings of every class lie approximately on a hypersphere and concentrate around a mean direction, so directional distributions are selected for all classes, whereas convolutional backbones yield Gaussian embeddings on CIFAR-10 and MNIST. Unlearning itself reshapes this geometry in an architecture-dependent way: on transformers, unlearning methods consistently concentrate the cluster of the forgotten class, whereas on convolutional networks they tend to disperse it relative to the original model. A replay mechanism that assumes one fixed shape for every class and every backbone is therefore misspecified precisely on the architectures where unlearning is increasingly deployed, and in ICU this misspecification compounds over successive steps. DISCERN addresses this by letting the data determine the family of each class, and our ablation confirms that this choice matters most in ICU, where it limits the resurfacing of previously forgotten classes.

The remainder of this paper is organized as follows. Section~\ref{sec:background} reviews related work on MU, CL and ICU, latent-space distributions, and distribution selection, and introduces the notation and formalizes the class unlearning and ICU problems. Section~\ref{sec:method} describes DISCERN: the candidate distributions, the per-class selection procedure, and the unlearning process. Section~\ref{sec:setup} describes the experimental setup, and Section~\ref{sec:results} presents and discusses the results in the three unlearning scenarios, the ablation study, and the analysis of cohesion and separation in the latent space. Finally, Section~\ref{sec:conclusions} concludes the paper and discusses limitations and future work.

Our main contributions are as follows:
\begin{itemize}
    \item a \textit{retain-free} class unlearning method that reconstructs retained knowledge through synthetic sampling in the representation space;
    \item a per-class distribution selection procedure that combines Euclidean and directional families to adapt to architecture-dependent latent geometries;
    \item a controlled evaluation of the effect of incrementality, comparing simultaneous and sequential deletion of the same set of classes;
    \item experimental evidence that per-class distribution selection reduces the reversal of accumulated forgetting, especially on the evaluated transformer architectures.
\end{itemize}

\section{Background}\label{sec:background}

This section positions DISCERN within the literature and sets out the formal framework of the paper. Section~\ref{sec:related} reviews related work, and Section~\ref{sec:prelim} introduces the notation and formalizes the unlearning problems we address.

\subsection{Related work}\label{sec:related}

\textbf{MU.} This field seeks methods that remove the influence of a set of data from a model already trained on it. Post hoc methods, applicable to any already-trained model, have proliferated in recent years under diverse strategies. Methods based on decision-boundary manipulation modify the classifier \textit{logits} to push the samples of the forget set out of their original class \cite{chen2023boundary}. Intervention-based approaches collect statistics of the model or the data in a stage prior to unlearning and use them as guidance during it, thereby dispensing with access to the original data at that time \cite{zhou2025decoupled}. Finally, repair-based approaches formulate unlearning as a network repair problem: PRUNE learns lightweight patch networks that redirect the model's predictions on the forget set while preserving performance on the remaining data \cite{LI2026108897}.

\textbf{CL and ICU.} ICU can be framed within CL \cite{wang2024comprehensive}. If $k$ classes are forgotten sequentially, the tasks are Task $0$, Task $1$, \dots, Task $k$, where Task $0$ is the original learning with all the data, Task $1$ consists of forgetting one class while remembering the rest, and Task $2$ consists of forgetting another class while remembering the rest, excluding those already forgotten; the subsequent tasks up to Task $k$ are analogous. Along these lines, several works apply CL techniques to forget a class \cite{devalapally2023simple}. Krawczyk et al. \cite{krawczyk2025continual} present a scenario that interleaves learning new classes with forgetting previously learned ones. In the same direction, we take PASS \cite{zhu2021prototype}, a class-incremental learning technique, and apply it to ICU.

\textbf{Latent-space distributions.} Several works model the distribution of each class in the representation space and use it as a substitute for the original data. PASS \cite{zhu2021prototype} stores a single prototype per class, the mean of its embeddings, and, while learning new tasks, replays pseudo-features obtained by adding isotropic Gaussian noise to that prototype, with a scale estimated from the average variance of the features; this keeps the old classes represented in the classifier without storing exemplars, and is combined with self-supervised label augmentation (image rotations) to learn more transferable features. In MU, SCAR \cite{bonato2024retain} relies on the feature distribution of each class, through a Mahalanobis distance, to move the samples to be forgotten towards the nearest wrong class, and Dehghani Tafti et al. \cite{dehghani2026universal} generate random embeddings in the intermediate space, pseudo-labeled by the model itself, so that existing unlearning methods can be applied without access to the training data, although without modeling the geometry of each class. When a parametric model of the class is used, it is therefore the Gaussian, but this assumption does not always hold. In natural language processing models, for instance, embeddings of related concepts are known not to follow a normal distribution \cite{li2020sentence}; rather, they cluster along directions.

\textbf{Distribution selection.} Since embeddings may follow non-Gaussian distributions, the range of candidate distributions must be broadened. On the one hand, one may consider additional distributions defined on Euclidean space, such as those discussed in \cite{kotz2019continuous}. On the other hand, embeddings may cluster along directions, which calls for distributions defined on directions, as studied in directional statistics \cite{mardia2009directional}. Once the candidates have been chosen, the one that best fits the data must be selected. In this regard, Claeskens and Hjort \cite{claeskens2008model} compare several selection criteria, such as the \emph{Bayesian information criterion} (BIC), which penalizes the number of parameters, and cross-validation, which directly estimates how well each model predicts unseen data. Since BIC relies on maximum likelihood estimates and on asymptotics with a fixed number of parameters, neither of which holds for shrinkage estimates of the covariance matrices of high-dimensional embeddings, we rely on cross-validation and compare families with the corrected resampled $t$-test of Nadeau and Bengio \cite{nadeau2003inference}.

Taken together, these works separately address three elements relevant to our problem: post hoc unlearning without full access to the retain set, the use of synthetic \textit{replay} to preserve knowledge in CL, and the modeling of class distributions in the latent space. However, none of these approaches explicitly combines the three elements in an ICU scenario: reconstructing the retained knowledge through synthetic sampling, adapting this sampling to the class-specific latent geometry, and assessing whether previously forgotten knowledge remains removed as new deletion requests arrive. DISCERN lies at this intersection.

\subsection{Preliminaries}\label{sec:prelim}

In this subsection, we introduce the notation used throughout the paper and formalize the unlearning problems addressed in our study. Let $\mathcal{D} = \{(\vec{x}_i, y_i)\}_{i=1}^N$ denote a dataset, where $\vec{x}_i$ is an input and $y_i \in \{1,\dots,C\}$ is its associated label, with $C$ the total number of classes and $N_c$ the number of samples belonging to class $c$. A classification model $f_\theta$ is trained on the training split of $\mathcal{D}$ to predict the correct label for each input image. If the output of the model is the logit vector $f_\theta(\vec{x}_i)=(z_i^1, \dots, z_i^j, \dots, z_i^C)$, then the predicted label is given by $\hat y_i = \underset{j}{\text{argmax}}\, z_i^j$, and the prediction is correct when $\hat y_i = y_i$. The quality of $f_\theta$ is evaluated on the test set $\mathcal{D}^t$.

The model $f_\theta$ can be decomposed into a backbone $h_{\theta_2}$ and a classifier $g_{\theta_1}$, such that $f_\theta = g_{\theta_1} \circ h_{\theta_2}$. For a given input $\vec{x}_i$, we denote its representation under $h_{\theta_2}$ by $\vec{e}_i$, that is, $h_{\theta_2}(\vec{x}_i) = \vec{e}_i$ $\in\mathbb{R}^d$, and refer to it as the embedding of $\vec{x}_i$, with $d$ the dimension of the embedding space. The set of embeddings associated with class $c$ is denoted by $E_c = \{\vec{e}_i : (\vec{x}_i,y_i) \in \mathcal{D},\ \vec{e}_i = h_{\theta_2}(\vec{x}_i),\ y_i = c\}$. Moreover, if these embeddings are assumed to follow a parametric distribution $P_c$ with parameters $\Lambda_c$, we write $E_c \sim P_c(\Lambda_c)$ as shorthand for the elements of $E_c$ being independent draws from $P_c(\Lambda_c)$.

In MU, the dataset $\mathcal{D}$ is partitioned into two disjoint subsets: the retain subset $\mathcal{D}_r$ and the forget subset $\mathcal{D}_f$. The objective is to obtain a new model $f_{\theta^U}$ such that the influence of $\mathcal{D}_f$ has been removed from the updated parameters $\theta^U$, while maintaining performance on $\mathcal{D}_r$ comparable to that of $f_\theta$. The ideal solution is to retrain a model from scratch on $\mathcal{D}_r$, obtaining $f_{{\theta^*}^U}$. Although full retraining is typically impractical, an effective unlearning method should produce outputs from $f_{\theta^U}$ that resemble those of $f_{{\theta^*}^U}$. In this work, we consider two unlearning scenarios: class unlearning \cite{bonato2024retain}; and ICU \cite{krawczyk2025continual}.

In class unlearning, we select one or more classes $\{c_1, c_2, \dots, c_k\} = C_f$, with $1 \leq k < C$, to be forgotten. The retain set contains no samples from the classes to be forgotten, that is, $\mathcal{D}_r = \{(\vec{x}_i,y_i)\in\mathcal{D}: y_i\notin C_f\}$, whereas all samples in $\mathcal{D}_f = \{(\vec{x}_i,y_i)\in\mathcal{D}: y_i\in C_f\}$ belong to one of the classes in $C_f$. For evaluation, the test set is split into a retain subset $\mathcal{D}^t_r$, which excludes classes in $C_f$, and a forget subset $\mathcal{D}^t_f$, which contains only classes in $C_f$. The unlearned model $f_{\theta^U}$ is expected to maximize accuracy on $\mathcal{D}^t_r$ while minimizing it on $\mathcal{D}^t_f$.

The second scenario is ICU. As in class removal, we consider a set of classes $\{c_1, c_2, \dots, c_k\} = C_f$, with $1 < k < C$, that must be forgotten. However, the classes are removed sequentially. At unlearning step $1\leq j \leq k$, we define $\mathcal{D}_{r,j} = \{(\vec{x}_i,y_i)\in\mathcal{D}: y_i\notin \{c_1,\dots,c_j\}\}$, $\mathcal{D}_{f,j} = \{(\vec{x}_i,y_i)\in\mathcal{D}: y_i = c_j\}$, and the current model $f_{{\theta}^U,j-1}$, where $f_{{\theta}^U, 0} = f_{\theta}$. At each step, the goal is to retain information from $\mathcal{D}_{r,j}$ while forgetting $\mathcal{D}_{f,j}$ in the model $f_{{\theta}^U,j-1}$, yielding an updated model $f_{{\theta}^U,j}$. This scenario evaluates how unlearning errors accumulate across successive forgetting steps, rather than how effectively a method removes multiple classes simultaneously.

An additional property of the incremental scenario is the persistence of forgetting. If $C_f^{(j)}=\{c_1,\ldots,c_j\}$ denotes the cumulative set of classes removed up to step $j$, an ICU method must not only reduce the information associated with the class $c_j$ removed at that step, but also maintain the forgetting of all classes in $C_f^{(j-1)}$. Hence, the quality of the incremental process depends jointly on three objectives: forgetting the current class, preventing the recovery of previously removed classes, and preserving performance on the remaining classes.

In general, unlearning methods rely on two distinct loss functions: one that preserves retained knowledge and is applied to $\mathcal{D}_r$, and another that promotes forgetting on $\mathcal{D}_f$. In practical settings, however, access to $\mathcal{D}_r$ may not always be available \cite{ahmed2025towards}. Accordingly, throughout this work we assume that $\mathcal{D}_r$ is inaccessible during unlearning, although statistics extracted from $\mathcal{D}$ prior to the unlearning stage remain available.

\section{DISCERN Method}\label{sec:method}

 DISCERN consists of two temporally separated stages. In an intervention phase prior to any deletion request, the original model is used to obtain the embeddings of each class and to estimate only the parameters of the distribution selected for it. The original samples and their embeddings need not be stored after this estimation. When a forgetting request is received, the method generates new embeddings from the stored parameters of the classes to be retained and uses these synthetic samples as a surrogate for the retain set. In this way, the method is retain-free during the unlearning process, although it requires statistics obtained beforehand through intervention.

This section is divided into three subsections: the first introduces the candidate distributions; the second explains the core of the proposal, namely fitting the best distribution to each class; and the last explains how forgetting is achieved.

\subsection{Candidate distributions}

DISCERN aims to simulate synthetic data that are representative of the classes to be retained. To obtain these data, we fit a distribution to each class and then sample from it. Only parametric distributions are considered, since nonparametric ones require much more information about the original embeddings to be fitted. For instance, kernel density estimation with a Gaussian kernel centers a Gaussian on each embedding; although such a nonparametric distribution may be more representative, it is questionable in a setting where $\mathcal{D}_r$ is inaccessible, since the embeddings can be reconstructed from the distribution itself.

\begin{figure}
    \centering
    \includegraphics[width=1.0\linewidth]{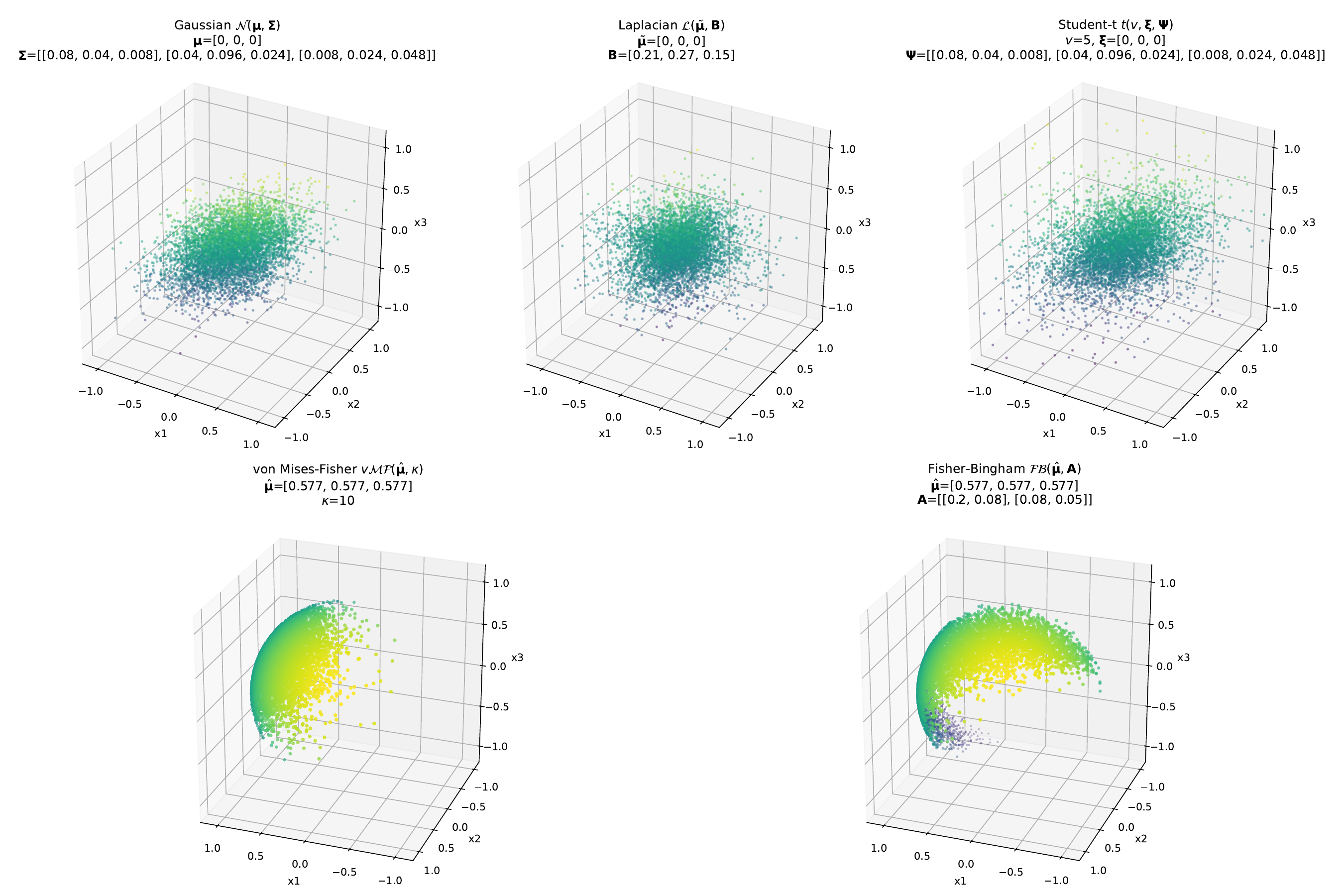}
    \caption{Samples from the parametric distributions; from left to right and top to bottom: Gaussian, Laplace, Student's $t$, von Mises--Fisher, and Fisher--Bingham.}
    \label{fig:distributions}
\end{figure}

The parametric distributions we consider are listed in Table~\ref{tab:distributions}. They fall into two groups: distributions defined on Euclidean space and directional distributions. To illustrate the differences, Figure~\ref{fig:distributions} shows samples from each of them in $\mathbb{R}^3$.

\begin{table}[t]
    
    \caption{Candidate distributions, indicating whether each one is directional and the rationale for its inclusion.}
    \centering
    \footnotesize
    \setlength{\tabcolsep}{4pt}
    \renewcommand{\arraystretch}{1.25}
    \begin{tabularx}{\textwidth}{%
        >{\centering\arraybackslash\hsize=1.0\hsize}Y|
        >{\centering\arraybackslash\hsize=0.4\hsize}Y|
        >{\centering\arraybackslash\hsize=1.6\hsize}Y
        >{\centering\arraybackslash\hsize=0\hsize}X}
        \hline
        Distribution & Directional & Rationale\\ \hline
        Gaussian, $\mathcal{N}(\vec\mu,\vec\Sigma)$ \cite{kotz2019continuous} & \xmark &
        Prototypes \cite{snell2017prototypical} are, a priori, ideal representatives of a class; all elements of the class may be distributed as noise around that prototype. It serves as a light-tailed elliptical reference. \\ \hline
        Student's $t$, $t(\nu, \vec\xi,\vec\Psi)$ \cite{kotz2004multivariate} & \xmark & If the embeddings of a class contain outliers with respect to a Gaussian that are still recognized as members of the class, the Student's $t$ distribution can capture them. Its heavier tails provide greater robustness to observations far from the center.\\ \hline
        Laplace, $\mathcal{L}(\tilde{\vec\mu}, \vec B)$ \cite{kotz2001laplace} & \xmark & The symmetric multivariate Laplace distribution is a Gaussian scale mixture with exponential mixing weights. Compared with the Gaussian, it places more mass both very close to the prototype and far from it, with exponential tails that are heavier than the Gaussian ones but lighter than the polynomial tails of the Student's $t$. It suits classes in which most embeddings are tightly clustered around the prototype while a minority lie farther away, a behavior intermediate between the Gaussian and the Student's $t$. \\ \hline
        von Mises--Fisher, $v\mathcal{MF}(\hat{\vec{\mu}},\kappa)$ \cite{mardia2009directional} & \cmark & Distribution on the unit hypersphere whose density decreases with the angle to a mean direction $\hat{\vec\mu}$, with a single concentration parameter $\kappa$ and hence the same dispersion in every direction around $\hat{\vec\mu}$. Directional latent spaces have proven effective in high-dimensional models; for instance, Xu and Durrett \cite{xu2018spherical} use von Mises--Fisher distributions on a hyperspherical latent space to stabilize variational autoencoders. It models representations whose information is mainly expressed by their direction. \\ \hline
        Fisher--Bingham, $\mathcal{FB}(\hat{\vec\mu},\vec\Sigma^T)$ \cite{mardia2009directional} & \cmark & Generalization of the von Mises--Fisher distribution that allows anisotropic concentration around the mean direction, so that the dispersion on the hypersphere may vary with the direction. In high dimension we use a tangent-plane approximation (Section~\ref{sec:selection}). \\ \hline
    \end{tabularx}
    \label{tab:distributions}
\end{table}

The two directional distributions considered are defined on the unit sphere around a mean direction $\hat{\vec\mu}$, which indicates the point around which the distribution is centered, while the remaining parameters describe the spread around it. The von Mises--Fisher density is proportional to $\exp(\kappa\hat{\vec\mu}^\top\vec x)$, where $\kappa\geq 0$ indicates how concentrated the points are around $\hat{\vec\mu}$; the spread is the same in every direction, that is, the distribution is isotropic. The Fisher--Bingham distribution generalizes it by adding a quadratic term: its density is proportional to $\exp(\kappa\hat{\vec\mu}^\top\vec x+\vec x^\top\vec A\vec x)$, so $\vec A$ controls the anisotropic spread around $\hat{\vec\mu}$, and $\vec A=\vec 0$ recovers the von Mises--Fisher distribution. The normalizing constant of the Fisher--Bingham distribution has no closed form, and evaluating it numerically becomes unreliable in the hundreds of dimensions of our embeddings. Owing to this high dimensionality, we do not fit this density but a normal distribution on the tangent plane at the mean direction (Section~\ref{sec:selection}): $\mathrm{Log}_{\hat{\vec\mu}}(\vec u)\sim\mathcal N(\vec 0,\vec\Sigma^T)$, where $\mathrm{Log}_{\hat{\vec\mu}}$ is the logarithmic map of the sphere at $\hat{\vec\mu}$, whose density on the sphere is
\begin{equation*}
    g(\vec u)=\mathcal N\!\left(\mathrm{Log}_{\hat{\vec\mu}}(\vec u);\vec 0,\vec\Sigma^T\right)\left(\frac{\theta}{\sin\theta}\right)^{d-2},
\end{equation*}
with $\theta$ the angle between $\vec u$ and $\hat{\vec\mu}$. Near $\hat{\vec\mu}$ this model approximates a Fisher--Bingham distribution whose precision on the tangent plane is $\kappa\vec I-2\vec A$, and we keep the name $\mathcal{FB}(\hat{\vec\mu},\vec\Sigma^T)$ for it. Its parameter $\vec\Sigma^T$ is therefore a covariance on the tangent plane, not the matrix $\vec A$ of the exact density. Both distributions are supported on the sphere, which is a strong restriction when fitting, since embeddings need not lie exactly on a sphere. Therefore, when modeling directional distributions we relax this condition and check whether the data lie approximately on a sphere. If so, we decompose the data into norm and direction: the angular part is fitted with a directional distribution, and the norm with a lognormal model whose mean may depend on the direction (Section~\ref{sec:selection}).

The families considered are not intended to be an exhaustive catalogue, but rather to cover qualitatively different latent geometries with a manageable number of parametric models. Nevertheless, other parametric families could be incorporated by adapting the selection procedure described in the next subsection accordingly.

\subsection{Distribution selection}\label{sec:selection}

\begin{figure}
    \centering

    \includegraphics[width=0.8\linewidth]{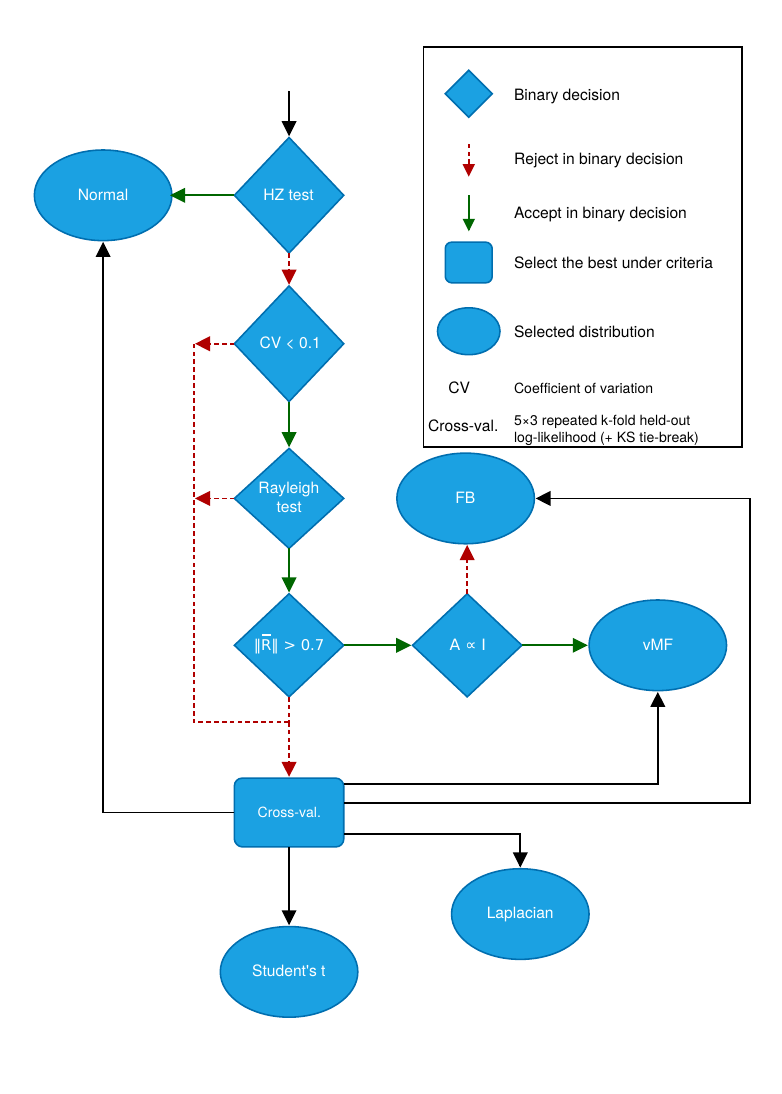}
    \caption{Flowchart for selecting the distribution of each class (Section~\ref{sec:selection}). Dashed red arrows indicate that the check is not passed. HZ: Henze--Zirkler test (passed if normality is not rejected, Step 1); $CV<0.1$, Rayleigh (passed if uniformity is rejected) and $||\overline{R}||>0.7$: directional checks (Step 2); $\vec\Sigma^T\propto\vec I$: sphericity test on the tangent covariance, which decides between $\mathcal{FB}$ and $v\mathcal{MF}$; Cross-val.: cross-validated log-likelihood comparison of the five families of Table~\ref{tab:distributions} (Step 3). The test of dependence between norm and direction is not shown, since it only affects the norm model of the directional families.}
    \label{fig:flowchart}
\end{figure}

This subsection explains how a distribution is selected for each class. The procedure, summarized in Figure~\ref{fig:flowchart}, has three steps. First, a normality test checks whether a Gaussian is enough to describe the class. Second, if it is not, a series of checks determines whether the class is compatible with a directional distribution. These two steps accept a specific distribution only when the data support it, in order to avoid selecting an overfitted one. Third, if neither accepts, the candidate families are compared by their cross-validated log-likelihood. If normality is rejected, the dependence between the norm and the direction of the embeddings is tested before the second step, since from then on a directional family may be selected and this dependence determines how its norm is modeled.

\textbf{Parameter estimation.} The mean of the Gaussian is the sample mean, and its covariance $\vec\Sigma_c$ is the Ledoit--Wolf shrinkage estimate \cite{ledoit2004well}, which remains well conditioned when $N_c$ is small relative to $d$. For the Student's $t$ and Laplace distributions, the location, the shape of the scatter matrix and, for the $t$, the degrees of freedom $\nu_c\in[2.05,1000]$ are estimated with EM-type algorithms (ECME for the $t$ \cite{liu1995ml}), applying Ledoit--Wolf shrinkage to the weighted scatter. After each iteration, the scale of $\vec\Psi_c$ and $\vec B_c$ is set so that the total variance (the trace of the covariance matrix) of the fitted distribution matches that of the Ledoit--Wolf covariance. For the $t$, $\mathrm{Cov}=\frac{\nu_c}{\nu_c-2}\vec\Psi_c$, so $\mathrm{tr}\,\vec\Psi_c=\frac{\nu_c-2}{\nu_c}\,\mathrm{tr}\,\vec\Sigma_c$, which requires $\nu_c>2$. For the Laplace distribution we use the parameterization $\vec x=\tilde{\vec\mu}_c+\sqrt{w}\,\vec B_c^{1/2}\vec\varepsilon$, with $w\sim\mathrm{Exp}(1)$ and $\vec\varepsilon\sim\mathcal N(\vec 0,\vec I)$, so $\mathrm{Cov}=\mathbb{E}[w]\,\vec B_c=\vec B_c$ and $\mathrm{tr}\,\vec B_c=\mathrm{tr}\,\vec\Sigma_c$. EM thus estimates the location, the shape of the matrix and $\nu_c$, while the scale is fixed by the total variance. A fit is flagged as degenerate if $\nu_c$ reaches its lower bound, if EM does not converge within 20 iterations, or, for the Laplace distribution, if the location collapses onto a sample; degenerate fits are never selected. For the von Mises--Fisher distribution, $\kappa_c$ is obtained with the approximation of Banerjee et al. \cite{banerjee2005clustering} refined by Newton iterations. The $\mathcal{FB}$ distribution is approximated by a Gaussian in the plane tangent to the sphere at the Karcher mean of the directions: the directions are mapped to that plane with the logarithmic map, $\vec\Sigma^T_c$ is their Ledoit--Wolf covariance, and samples are mapped back to the sphere with the exponential map. This approximation is forced by the high dimensionality of the embeddings, since the exact $\mathcal{FB}$ normalizing constant cannot be evaluated reliably there. It is accurate when the directions are concentrated around their mean, which Step~2 requires, because the tangent Gaussian then places negligible mass at angles close to $\pi$, where the exponential map stops being injective. It is also consistent with the von Mises--Fisher case: for large $\kappa_c$, a $v\mathcal{MF}$ is approximately a tangent Gaussian with covariance $\kappa_c^{-1}\vec I$, so an isotropic $\vec\Sigma^T_c$ corresponds to a $v\mathcal{MF}$. These estimators are therefore not all maximum likelihood estimators, but tractable approximations to them in high dimension.

\textbf{Step 1: normality.} The first step checks whether the data follow a Gaussian distribution. Since embeddings live in a high-dimensional space, we apply a test of multivariate normality, namely the Henze--Zirkler test \cite{henze1990class}, with $\alpha=0.05$ to $E_c$. If normality is not rejected, we assume $E_c \sim \mathcal{N}(\vec \mu_c, \vec \Sigma_c)$; otherwise, we proceed to the second step. The test uses the sample covariance of $E_c$. When $N_c\leq d$, as in CIFAR-100, this covariance is singular and the test cannot be computed, so the test is not run and, by convention, the class is treated as non-Gaussian and it proceeds to the next step. Non-rejection is not evidence of normality, and in high dimension the test may have limited power, so the Gaussian is adopted here as a parsimonious working model rather than as an identification of the true distribution.

\textbf{Norm--direction dependence.} If normality is rejected, the class may end up modeled by a directional family, either in Step 2 or in Step 3, so the dependence between the norm and the direction is tested at this point. Directional distributions are supported on the sphere, whereas the embeddings do not lie exactly on it. We therefore describe each embedding by its norm $r_i^c=||\vec e_i^c||$ and its direction $\hat{\vec e}_i^c=\vec e_i^c/r_i^c$, and model both components jointly. Rather than assuming that they are independent, we test it with three complementary tests: the Spearman correlation between $r_i^c$ and the cosine between $\hat{\vec e}_i^c$ and the mean direction of the class,
\begin{equation*}
    \hat{\vec\mu}_c= \dfrac{\sum\limits_{i=1}^{N_c} \hat{\vec e}_i^c}{\left|\left|\sum\limits_{i=1}^{N_c} \hat{\vec e}_i^c\right|\right|};
\end{equation*}
a ridge regression of $\log r_i^c$ on $\hat{\vec e}_i^c$, whose leave-one-out $R^2$ is compared with that obtained under 100 permutations of $\log r_i^c$; and the distance correlation \cite{szekely2007measuring} between $r_i^c$ and $\hat{\vec e}_i^c$, with 200 permutations. The three $p$-values are combined with the Benjamini--Hochberg procedure \cite{benjamini1995controlling} at $\alpha=0.05$, and dependence is declared if at least one test is rejected, which is equivalent to Simes' test of the global null hypothesis of independence. The outcome only affects the norm model of the directional families. Under independence, $\log r_i^c\sim\mathcal{N}(m_c,s_c^2)$, that is, the norm is lognormal; under dependence, $\log r_i^c\mid\hat{\vec e}_i^c\sim\mathcal{N}(\vec a_c^\top\hat{\vec e}_i^c+b_c,\,s_c^2)$, where $\vec a_c$ and $b_c$ come from the ridge regression above and $s_c^2$ is its leave-one-out error. To sample a directional embedding, a direction is drawn first, with Wood's algorithm \cite{wood1994simulation} for $v\mathcal{MF}$ or through the tangent Gaussian for $\mathcal{FB}$, and then its norm conditionally on that direction.

\textbf{Step 2: directional checks.} Next, we check whether $E_c$ follows an $\mathcal{FB}$ distribution through the following checks:
\begin{enumerate}
    \item We check whether $E_c$ is distributed approximately on a sphere by computing the coefficient of variation of the norms of the elements of $E_c$, i.e.,
    \begin{equation*}
        CV = \dfrac{\sigma}{\mu}, \quad \mu = \dfrac{1}{N_c}\sum_{i=1}^{N_c} \left|\left| \vec e_i^c\right|\right|, \quad \sigma^2 = \dfrac{1}{N_c}\sum_{i=1}^{N_c} \left(\left|\left| \vec e_i^c \right|\right| - \mu\right)^2.
    \end{equation*}
     $E_c$ is assumed to lie on a sphere if $CV$ is below $0.1$.
    \item Assuming that the data lie approximately on the surface of a sphere, the Rayleigh test \cite{mardia2009directional} is applied to rule out a uniform distribution. The checks continue if the test is rejected at a significance level of $0.05$; otherwise, the data are assumed not to follow an $\mathcal{FB}$ distribution.

    \item Once uniformity has been ruled out, we check whether the elements of $E_c$ point in the same direction by computing
    \begin{equation*}
        \overline{R} = \dfrac{1}{N_c} \sum_{i=1}^{N_c} \dfrac{\vec e_i^c}{||\vec e_i^c||},
    \end{equation*}
    If $||\overline{R}||$ exceeds $0.7$, we assume that the elements of $E_c$ point approximately in the same direction.
\end{enumerate}

The thresholds used in this filtering ($CV<0.1$ and $||\overline{R}||>0.7$) are operational criteria for identifying, respectively, an approximately spherical structure and a marked directional concentration, rather than universal constants of the representation space. Both have a direct geometric reading. $CV<0.1$ means that the typical spread of the norms is below $10\%$ of the mean norm, so the embeddings concentrate in a shell that is thin relative to its radius. In turn, $||\overline{R}||$ equals the mean cosine between each normalized embedding and the mean direction of the class, so the threshold $0.7\approx\cos 45^\circ$ corresponds roughly to a typical angular deviation of $45^\circ$ from that direction. This is a demanding value. If the directions were uniform, $\mathbb{E}||\overline{R}||^2=1/N_c$ for any $d$, so $||\overline{R}||$ would be of order $1/\sqrt{N_c}$; moreover, in high dimension the directions within $45^\circ$ of a given one cover an exponentially small fraction of the sphere. Both thresholds are kept fixed across all architectures and datasets to avoid scenario-specific tuning.

If all these checks are passed, $E_c$ is assigned a directional distribution, with the norm modeled as described before Step 2. To choose between the two directional families, a sphericity test is applied to the tangent-plane covariance: if it is not proportional to the identity matrix, the directions follow $\mathcal{FB}(\hat{\vec\mu}_c, \vec\Sigma^T_c)$; otherwise, they follow $v\mathcal{MF}(\hat{\vec\mu}_c, \kappa_c)$. For this test, the directions are rotated with a Householder reflection so that the mean direction becomes the first basis vector, and are mapped with the logarithmic map to the tangent plane, of dimension $p=d-1$, as in the fit of $\mathcal{FB}$. The sphericity of the sample covariance $\vec S_c$ of the projected directions is then tested against a $\chi^2$ distribution with $p(p+1)/2-1$ degrees of freedom. If $N_c\geq 1.1p$, we use the likelihood-ratio test with Bartlett correction, whose statistic is $(N_c-1)\,p\,\log(\mathrm{AM}/\mathrm{GM})\left[1-\frac{2p^2+p+2}{6p(N_c-1)}\right]$, where $\mathrm{AM}$ and $\mathrm{GM}$ are the arithmetic and geometric means of the eigenvalues of $\vec S_c$. If $p<N_c<1.1p$, we use John's test, with statistic $\frac{N_c\,p(p+1)}{2(p-1)}\left(\frac{p\,\mathrm{tr}\,\vec S_c^2}{(\mathrm{tr}\,\vec S_c)^2}-1\right)$. If $N_c\leq p$, we do not test sphericity and keep the more parsimonious $v\mathcal{MF}$. This is a design choice rather than a mathematical impossibility: $\vec S_c$ is singular, so the likelihood-ratio statistic is not defined, but trace-based statistics such as John's remain applicable when the dimension exceeds the sample size \cite{ledoit2002some}. For $\mathcal{FB}$, $\hat{\vec\mu}_c$ denotes the Karcher mean of the directions, which for concentrated directions is close to the normalized mean direction defined above.

\textbf{Step 3: cross-validation.} If any of the checks above fails, the five families of Table~\ref{tab:distributions} are compared by repeated $k$-fold cross-validation, with $k=5$ and three repetitions. In each of the resulting 15 folds, every family is fitted on the training folds and scored by its mean log-likelihood per point on the validation fold; the selection criterion is the average of this score over the 15 folds. For this comparison to be meaningful, all families are evaluated as densities on $\mathbb{R}^d$. For the directional ones, the density of an embedding $\vec e$ with norm $r$ and direction $\hat{\vec e}$ is
\begin{equation*}
    \log f(\vec e) = \log g(\hat{\vec e}) + \log \phi\left(\log r \mid \hat{\vec e}\right) - d\log r,
\end{equation*}
where $g$ is the directional density, $\phi$ is the normal density of the norm model described before Step 2, and the term $-d\log r$ collects the Jacobians of the change to polar coordinates, $-(d-1)\log r$, and of the logarithm of the norm, $-\log r$. For $\mathcal{FB}$, $g$ includes in addition the Jacobian of the exponential map, $(d-2)\log(\theta/\sin\theta)$, with $\theta$ the angle to the Karcher mean.

Let $f^\star$ be the family with the highest mean log-likelihood. A family $f$ is considered tied with $f^\star$ if it is not significantly worse according to the corrected resampled $t$-test, applied to the 15 per-fold differences (one-sided, $\alpha=0.05$), or if its mean log-likelihood is within a practical margin of $0.05$ nats per dimension of that of $f^\star$; expressing the margin per dimension makes it comparable across architectures. The margin is deliberately loose, so the likelihood acts as a coarse filter. Among the tied families, we favor the one whose samples best reproduce the norm of the embeddings, which determines the spherical shell seen by the classifier. This is measured by the Kolmogorov--Smirnov statistic between the norms of the validation embeddings and those of the same number of synthetic samples in each fold. The family with the lowest mean statistic wins, and the families whose per-fold statistics do not differ significantly from it (two-sided corrected resampled $t$-test) are tied again, in which case the highest log-likelihood decides. Degenerate fits score $-\infty$ in the corresponding fold. The winning family is finally refitted on all of $E_c$; if this refit is degenerate, the next family by log-likelihood is used, with the Gaussian as the last resort. Classes with fewer than ten embeddings are directly assigned a Gaussian.

\subsection{Unlearning process}

Unlearning relies on two loss functions: $L_f$, which forgets the classes in $\mathcal{D}_f$, and $L_r$, which prevents forgetting of $\mathcal{D}_r$. DISCERN is \textit{retain-free}, since it has no access to $\mathcal{D}_r$ at unlearning time, but it simulates it at the \textit{embedding} level. For each class $c \notin C_f$, $N_s^c$ elements are sampled from $P_c(\Lambda_c)$
\begin{equation*}
    \vec e_i^c \sim P_c(\Lambda_c), \ 1\leq i \leq N_s^c,
\end{equation*}

and the synthetic \textit{dataset} is defined as
    $\mathcal{D}_s = \left\{ (\vec e^c_i,c) :  \ 1\leq i \leq N_s^c, c \notin C_f \right\}.$

The synthetic embeddings are fed directly to the input of the classifier $g_{\theta_1}$, whereas the images of $\mathcal{D}_f$ go through the whole model. The distributions are fitted once on the backbone of the original model and reused at every step of ICU.

On the one hand, $L_r$ is the cross-entropy on $\mathcal{D}_s$, computed over the softmax of all $C$ classes, and zero on $\mathcal{D}_f$. On the other hand, $L_f$ is zero on $\mathcal{D}_s$, while on $\mathcal{D}_f$ it is a function that promotes forgetting. We use two such functions depending on the unlearning problem at hand. For class removal, we use the following function, which we call notop:

\begin{equation*}
    L_f(f_\theta(\vec x_i), y_i) = \text{}\ \max\left(0, f_\theta(\vec x_i)^{y_i} - \underset{j\neq y_i}{\max}\{f_\theta(\vec x_i)^j\} + m\right), \quad m \geq 0,
\end{equation*}

whereas in ICU we apply a confidence penalty:

\begin{equation*}
    L_f(f_\theta(\vec x_i), y_i) = \sum\limits_{c\notin C_f} p^c(\vec x_i) \log p^c(\vec x_i), \quad p^c = \text{}\ \dfrac{\exp{(z^c_i)}}{\sum\limits_{c'\notin C_f} \exp{(z^{c'}_i)}}.
\end{equation*}
At step $j$ of ICU, $C_f$ in this expression is the cumulative set $C_f^{(j)}$.

The former drives the model to misclassify the class to be forgotten, whereas the latter lowers the confidence of the forgotten class on the classes that will continue to be retained. The confidence penalty does not depend on the logits of the forgotten classes, which are excluded from its softmax. The direct pressure on those logits comes from $L_r$. Because the cross-entropy on $\mathcal{D}_s$ is computed over all $C$ classes and no synthetic samples of the forgotten classes are generated, its gradient with respect to the logit $z^f$ of a forgotten class $f$ is $p^f>0$ on every synthetic sample. Hence $L_r$ lowers that logit throughout the regions occupied by the retained classes, while $L_f$ flattens the distribution over the retained classes on the forgotten images.

The total loss is $L=L_r+L_f$, with unit weight for both terms, each averaged over its own part of the batch. The synthetic set is sampled once at the start of each unlearning process, and batches are formed by shuffling its embeddings together with $80\%$ of $\mathcal{D}_f$ (the remaining $20\%$ is used for validation).

\section{Experimental Setup}\label{sec:setup}

\textbf{Datasets.} MNIST \cite{lecun2002gradient}, CIFAR-10 and CIFAR-100 \cite{krizhevsky2009learning}. All three datasets are used to evaluate single-class unlearning, whereas simultaneous multi-class unlearning and ICU are evaluated on CIFAR-100, the only one with enough classes to define forget groups of increasing size and long unlearning sequences.

\textbf{Architectures.} We evaluate two convolutional architectures, EfficientNetV2-S \cite{tan2021efficientnetv2} and ResNet18 \cite{he2016deep}, and two transformer architectures, ViT-B/16 \cite{dosovitskiy2020image} and Swin-T \cite{liu2021swin}, in all three scenarios. The split into two families is sharper on paper than in practice: whereas ViT-B/16 keeps a single resolution and global attention in all its layers, Swin-T builds hierarchical feature maps through patch merging and computes attention within shifted local windows \cite{liu2021swin}, thereby reintroducing the locality and multiscale structure of convolutional networks. Swin-T thus occupies an intermediate position between the two families.

\textbf{Metrics.} To evaluate prediction-level performance, we use $Acc_f$, the accuracy of the model on the test set of the classes to be forgotten, $\mathcal{D}_f^t$, and $Acc_r$, the accuracy on the test set of the retained classes, $\mathcal{D}_r^t$. Since both objectives usually compete---a method that forgets aggressively tends to degrade $Acc_r$, and vice versa---we also report $H$, the harmonic mean of $Acc_r$ and $(1-Acc_f)$,
\begin{equation*}
    H = \dfrac{2\,Acc_r\,(1-Acc_f)}{Acc_r + 1 - Acc_f},
\end{equation*}
which summarizes both objectives in a single value and penalizes methods that excel at one at the expense of the other. In the single-class and multi-class tables, $H$ is computed from the mean $Acc_r$ and $Acc_f$ over runs. In ICU, $H$ is computed for each run and step and then averaged, so that its standard deviation across runs can be reported. Finally, to analyze the effect of unlearning at the representation level, we use cohesion and separation measures on the \textit{embeddings}, described in Section~\ref{sec:clustering-metrics}.

\textbf{Baselines.} We compare our proposal with other \textit{retain-free} methods: Boundary Shrink and Boundary Expand \cite{chen2023boundary}, Bad Teacher \cite{Chundawat_Tarun_Mandal_Kankanhalli_2023}, DELETE \cite{zhou2025decoupled}, EWC \cite{zhang2023machine}, LF-SSD \cite{DBLP:conf/iclr/FosterSB24}, and Random Labeling \cite{graves2021amnesiac, fan2024salun}, described in Table~\ref{tab:methods}. We also report the performance of the original model and of retraining; since the original model starts from a pretrained architecture, retraining corresponds to fine-tuning that architecture on $\mathcal{D}_r$. For each method, we use whenever possible the hyperparameters reported in the original publication or, failing that, those provided in the corresponding repository; when neither source is available, we perform a brief hyperparameter search to obtain adequate forgetting.

\begin{table}[]
    \centering
    \caption{Description of the compared methods.}
    \begin{tabularx}{\textwidth}{%
        >{\centering\arraybackslash\hsize=0.6\hsize}Y|
        >{\centering\arraybackslash\hsize=0.4\hsize}Y|
        >{\centering\arraybackslash\hsize=2.0\hsize}Y
        >{\centering\arraybackslash\hsize=0\hsize}X}
        \hline
        Method & Intervention & Description\\ \hline
        Boundary Shrink \cite{chen2023boundary} & \xmark & Reassigns each sample of the class to be forgotten to its nearest neighboring class, identified through an adversarial attack, and fine-tunes the model, thereby shrinking the decision boundary of that class. \\ \hline
        Boundary Expand \cite{chen2023boundary} & \xmark & Extends the classifier with an auxiliary class to which the samples to be forgotten are reassigned; after fine-tuning, the added neuron is removed, dispersing the knowledge of the target class. \\ \hline
        Bad Teacher \cite{Chundawat_Tarun_Mandal_Kankanhalli_2023} & \xmark & Distillation with two teachers: the original model guides the knowledge to be preserved, while an incompetent (randomly initialized) teacher guides the response on the samples to be forgotten. \\ \hline
        DELETE \cite{zhou2025decoupled} & \xmark & Decouples the teacher logits into target-class and non-target knowledge, masks the forgotten classes, and transfers only the residual knowledge to the student. \\ \hline
        EWC \cite{zhang2023machine} & \cmark & Applies a forgetting-promoting loss on the data to be forgotten, regularized with the Fisher information matrix, which penalizes changes to parameters relevant to the retained knowledge. \\ \hline
        LF-SSD \cite{DBLP:conf/iclr/FosterSB24} & \cmark & Estimates the importance of each parameter without labels, from the model outputs, and selectively dampens those that are disproportionately important for the forget set. \\ \hline
        Random Labeling \cite{graves2021amnesiac, fan2024salun} & \xmark & Assigns random incorrect labels to the samples of the class to be forgotten and fine-tunes the model on them. \\ \hline
    \end{tabularx}
    \label{tab:methods}
    
\end{table}

To keep the result tables compact, we abbreviate Boundary Shrink and Boundary Expand as Shrink and Expand, Bad Teacher as BT, and Random Labeling as RL. The two variants of DISCERN, which differ only in the forgetting loss $L_f$ (Section~\ref{sec:method}), are denoted Notop and Conf. (confidence) in those tables; both refer to the proposed method.

For the proposed method, each unlearning process generates a total of 1000 embeddings, randomly distributed among the retained classes, and notop uses a margin of 10. Both values are kept fixed across all experiments, and Section~\ref{sec:ablationHyper} analyzes the sensitivity of the method to both.

The original model for each dataset--architecture combination is obtained by fine-tuning ImageNet-1K pretrained weights on the corresponding dataset for 30 epochs, keeping the epoch with the best validation performance. For single-class unlearning, we perform 10 runs on CIFAR-10 and MNIST and 20 runs on CIFAR-100, forgetting a different class in each run. For multi-class unlearning, we perform 10 runs with different seeds on CIFAR-100, using groups of 5, 10, 20, and 50 randomly selected classes for each seed. For ICU, we perform 10 runs with the same seeds, forgetting 20 classes sequentially, one at a time, in a fixed order shared by all methods. For each seed, the 20-class group in ICU is exactly the 20-class group of the multi-class scenario: both scenarios start from the same original model and end with the same set of forgotten classes, and only the deletion schedule differs (simultaneous versus sequential).

During unlearning, each method has access only to $\mathcal{D}_f$ and, where applicable, to the data obtained through intervention. A small validation set is held out from $\mathcal{D}_f$, and the whole model is frozen except for the classifier and the last stage of the backbone, which produces the embeddings and is therefore part of $h_{\theta_2}$. In DISCERN, this stage only receives gradient from $L_f$, since the synthetic embeddings are injected after it (Section~\ref{sec:method}). Methods fine-tune the model for at most 30 epochs, with early stopping when the validation accuracy falls below chance level. We use a patience of 1 epoch because, as training progresses, the model becomes increasingly prone to catastrophic forgetting, which the methods cannot detect since they lack access to the retain set. Accordingly, we consider a class forgotten when its accuracy is at or below chance level.

Further experimental details are available in our code on GitHub \url{https://github.com/ari-dasci/S-DISCERN}.

\section{Results and Discussion}\label{sec:results}

This section presents and discusses the results obtained under the protocol of Section~\ref{sec:setup}. Before turning to the unlearning results, we examine which distributions DISCERN actually selects, since this choice conditions the interpretation of what follows. Section~\ref{sec:unlearningResults} then compares its unlearning performance with that of the baselines in the three scenarios: single-class unlearning, simultaneous multi-class unlearning, and ICU. Section~\ref{sec:ablation} reports the ablation study, Section~\ref{sec:runtime} compares their computational cost and, finally, Section~\ref{sec:clustering-metrics} analyzes how unlearning reshapes the cohesion and separation of the classes in the latent space.

Table~\ref{tab:selectedDist} reports, for each architecture--dataset pair, the distribution assigned to each class by the selection procedure of Figure~\ref{fig:flowchart}. The selection is largely determined by the architecture: the two transformers choose directional distributions for $100\%$ of the classes in all three datasets, whereas on CIFAR-10 and MNIST the two convolutional networks almost always end up with Euclidean families: the Gaussian for every class of EfficientNetV2-S and for 15 of the 20 classes of ResNet18, which additionally assigns Student's $t$ to four CIFAR-10 classes and the Fisher--Bingham distribution to a single MNIST class. This is consistent with the first directional check of the procedure, based on the coefficient of variation of the norms: the embeddings of the evaluated transformers lie approximately on a sphere, whereas those of the convolutional networks generally do not. CIFAR-100 is the exception: no class passes the normality test, and the convolutional networks spread their classes over several families. EfficientNetV2-S chooses directional distributions for 98 of its 100 classes, whereas on ResNet18 the Gaussian prevails (68/100), followed by the Fisher--Bingham distribution (24/100).

Among the directional families, the choice made in Step 2 depends on the number of classes. With ten classes the Fisher--Bingham distribution, which allows anisotropy, is selected, whereas with one hundred classes its isotropic special case, the von Mises--Fisher distribution, is systematically kept. This follows from the sphericity test that decides between them (Section~\ref{sec:selection}). CIFAR-100 provides only about $500$ samples per class, which is no more than the dimension $d-1$ of the tangent plane in any of the four architectures. In this regime we do not test sphericity and, by design, keep the von Mises--Fisher distribution. In CIFAR-10 and MNIST, by contrast, sphericity was rejected in every class that reached the test. The Fisher--Bingham classes that appear on CIFAR-100 with ResNet18 and EfficientNetV2-S do not come from this test but from the cross-validation of Step 3, applied to the classes that fail the directional checks. Finally, all five families in Table~\ref{tab:distributions} are selected for at least one class, although Student's $t$ and the Laplace distribution only marginally: 4 and 2 of the 480 classes, all on ResNet18. As for the remaining decisions of the procedure, the norm--direction test rejected independence in every class in which it was applied, so the norm of the directional families is always modeled conditionally on the direction. Moreover, removing the practical margin of Step 3 does not change the selected family for any of the 105 classes decided by cross-validation.

\begin{table}[]
    \centering
    \caption{Distribution selected per class for each architecture--dataset pair. In parentheses, the number of classes assigned to each family out of the total.}
    \label{tab:selectedDist}
    \begin{tabular}{l|c c c} \hline
        Architecture & CIFAR-10 & CIFAR-100 & MNIST \\ \hline
        ResNet18 & \begin{tabular}[c]{@{}c@{}}Gaussian (6/10)\\ Student's $t$ (4/10)\end{tabular} & \begin{tabular}[c]{@{}c@{}}Gaussian (68/100), $\mathcal{FB}$ (24/100)\\ $v\mathcal{MF}$ (6/100), Laplace (2/100)\end{tabular} & \begin{tabular}[c]{@{}c@{}}Gaussian (9/10)\\ $\mathcal{FB}$ (1/10)\end{tabular} \\ \hline
        EfficientNetV2-S & Gaussian (10/10) & \begin{tabular}[c]{@{}c@{}}$v\mathcal{MF}$ (93/100), $\mathcal{FB}$ (5/100)\\ Gaussian (2/100)\end{tabular} & Gaussian (10/10) \\ \hline
        Swin-T & $\mathcal{FB}$ (10/10) & $v\mathcal{MF}$ (100/100) & $\mathcal{FB}$ (10/10) \\
        ViT-B/16 & $\mathcal{FB}$ (10/10) & $v\mathcal{MF}$ (100/100) & $\mathcal{FB}$ (10/10) \\ \hline
    \end{tabular}
\end{table}

\subsection{Evaluating unlearning}\label{sec:unlearningResults}

This subsection is organized into three blocks, one per unlearning scenario: single-class unlearning, simultaneous multi-class unlearning, and ICU. In each of them, we report $Acc_r$, $Acc_f$, and $H$ for the four architectures, comparing DISCERN with the seven retain-free baselines described in Table~\ref{tab:methods}, the original model, and retraining.

\textbf{Single-class unlearning.} Tables~\ref{tab:VitSingleClassRemoval}, \ref{tab:SwinSingleClassRemoval}, \ref{tab:ResnetSingleClassRemoval}, and~\ref{tab:EffSingleClassRemoval} report the results for ViT-B/16, Swin-T, ResNet18, and EfficientNetV2-S, respectively, averaged over the runs in which each class is forgotten separately. For the original model, $Acc_r$ and $Acc_f$ coincide on CIFAR-10 and MNIST because each class is forgotten exactly once and the test set is balanced, so both equal the mean per-class accuracy. On CIFAR-100 only 20 of the 100 classes are forgotten, so the two values differ slightly.

\begin{table}[]
    \centering
    \caption{ViT-B/16 Single Class Removal}
    \label{tab:VitSingleClassRemoval}
    \begin{tabular}{c|S[table-format=0.4] S[table-format=0.4] S[table-format=0.4]|S[table-format=0.4] S[table-format=0.4] S[table-format=0.4]|S[table-format=0.4] S[table-format=0.4] S[table-format=0.4]} \hline
        \multirow{2}{*}{Methods} & \multicolumn{3}{c}{CIFAR-10} & \multicolumn{3}{c}{CIFAR-100} & \multicolumn{3}{c}{MNIST} \\ \cline{2-10}
         & {$Acc_r\uparrow$} & {$Acc_f\downarrow$} & {$H\uparrow$} & {$Acc_r\uparrow$} & {$Acc_f\downarrow$} & {$H\uparrow$} & {$Acc_r\uparrow$} & {$Acc_f\downarrow$} & {$H\uparrow$} \\ \hline
         Original & .9650000000000001 & .9650000000000001 & .0675 & .8502676767676768 & .8574999999999999 & .2441 & .9963 & .9963 & .0074 \\
         Retrain & .9697111111111111 & .0 & .9846 & .8490757575757575 & .0 & .9184 & .9967 & .0 & .9983 \\ \hline
         Shrink & .9666888888888888 & .23770000000000002 & .8524 & \bfseries .8507474747474747 & .6505 & .4955 & .9963222222222223 & .994 & .0119 \\
         Expand & \bfseries .9673 & .9012 & .1793 & .8506313131313131 & .7735 & .3577 & .9965333333333334 & .9856999999999999 & .0282 \\
         BT & .9547222222222222 & .0209 & .9668 & .8489545454545455 & .0035000000000000005 & .9168 & .9965999999999999 & .038400000000000004 & .9788 \\
         DELETE & .9590444444444444 & \bfseries .0 & .9791 & .8488737373737374 & .0015 & .9176 & .9960444444444445 & .0001 & .9980 \\
         EWC & .9254555555555555 & \bfseries .0 & .9613 & .8477121212121211 & \bfseries .0005 & .9174 & .9961666666666666 & \bfseries .0 & \bfseries .9981 \\
         LF-SSD & \bfseries .9673111111111112 & .0774 & .9444 & .8488636363636364 & .001 & \bfseries .9178 & \bfseries .9966555555555555 & .8432000000000001 & .2710 \\
         RL & .9667555555555556 & \bfseries .0 & \bfseries .9831 & .8485959595959596 & .0025 & .9170 & .9965999999999999 & .00030000000000000003 & \bfseries .9981 \\ \hline
         Notop & 0.9134000000 & \bfseries 0.0000000000 & 0.9547000000 & 0.8469949495 & \bfseries 0.0005000000 & 0.9169000000 & 0.9960555556 & \bfseries 0.0000000000 & 0.9980000000 \\
         Conf. & 0.9509333333 & 0.0528000000 & 0.9491000000 & 0.8489040404 & 0.1315000000 & 0.8586000000 & 0.9960777778 & 0.0464000000 & 0.9744000000 \\ \hline
    \end{tabular}

\end{table}

\begin{table}[]
    \centering
    \caption{Swin-T Single Class Removal}
    \label{tab:SwinSingleClassRemoval}
    \begin{tabular}{c|S[table-format=0.4] S[table-format=0.4] S[table-format=0.4]|S[table-format=0.4] S[table-format=0.4] S[table-format=0.4]|S[table-format=0.4] S[table-format=0.4] S[table-format=0.4]} \hline
        \multirow{2}{*}{Methods} & \multicolumn{3}{c}{CIFAR-10} & \multicolumn{3}{c}{CIFAR-100} & \multicolumn{3}{c}{MNIST} \\ \cline{2-10}
         & {$Acc_r\uparrow$} & {$Acc_f\downarrow$} & {$H\uparrow$} & {$Acc_r\uparrow$} & {$Acc_f\downarrow$} & {$H\uparrow$} & {$Acc_r\uparrow$} & {$Acc_f\downarrow$} & {$H\uparrow$} \\ \hline
         Original & 0.9702000000 & 0.9702000000 & 0.0578000000 & 0.8583989899 & 0.8585000000 & 0.2430000000 & 0.9958000000 & 0.9958000000 & 0.0084000000 \\
         Retrain & 0.9725444444 & 0.0000000000 & 0.9861000000 & 0.8552373737 & 0.0000000000 & 0.9220000000 & 0.9968000000 & 0.0000000000 & 0.9984000000 \\ \hline
         Shrink & 0.9711000000 & 0.3064000000 & 0.8092000000 & 0.8583282828 & 0.2860000000 & 0.7795000000 & \bfseries 0.9961111111 & 0.9865000000 & 0.0266000000 \\
         Expand & \bfseries 0.9724444444 & 0.9013000000 & 0.1792000000 & \bfseries 0.8588787879 & 0.6375000000 & 0.5098000000 & \bfseries 0.9961000000 & 0.9811000000 & 0.0371000000 \\
         BT & 0.9656888889 & 0.0829000000 & 0.9408000000 & 0.8524090909 & 0.0515000000 & 0.8979000000 & \bfseries 0.9961111111 & 0.0751000000 & 0.9592000000 \\
         DELETE & 0.9675111111 & \bfseries 0.0000000000 & 0.9835000000 & 0.8544090909 & 0.0050000000 & 0.9194000000 & 0.9944777778 & \bfseries 0.0000000000 & 0.9972000000 \\
         EWC & 0.9526000000 & \bfseries 0.0000000000 & 0.9757000000 & 0.8540606061 & 0.0015000000 & 0.9206000000 & 0.9922111111 & \bfseries 0.0000000000 & 0.9961000000 \\
         LF-SSD & 0.9258333333 & \bfseries 0.0000000000 & 0.9615000000 & 0.6693737374 & \bfseries 0.0000000000 & 0.8019000000 & 0.9947666667 & \bfseries 0.0000000000 & 0.9974000000 \\
         RL & 0.9701666667 & \bfseries 0.0000000000 & \bfseries 0.9849000000 & 0.8532777778 & 0.0410000000 & 0.9031000000 & \bfseries 0.9961333333 & \bfseries 0.0000000000 & \bfseries 0.9981000000 \\ \hline
         Notop & 0.9444333333 & \bfseries 0.0000000000 & 0.9714000000 & 0.8547424242 & 0.0010000000 & \bfseries 0.9213000000 & 0.9919666667 & \bfseries 0.0000000000 & 0.9960000000 \\
         Conf. & 0.9620888889 & 0.0200000000 & 0.9710000000 & 0.8479595960 & 0.6990000000 & 0.4443000000 & 0.9955777778 & 0.0003000000 & 0.9976000000 \\ \hline
    \end{tabular}

\end{table}

\begin{table}[]
    \centering
    \caption{ResNet18 Single Class Removal}
    \label{tab:ResnetSingleClassRemoval}
    \begin{tabular}{c|S[table-format=0.4] S[table-format=0.4] S[table-format=0.4]|S[table-format=0.4] S[table-format=0.4] S[table-format=0.4]|S[table-format=0.4] S[table-format=0.4] S[table-format=0.4]} \hline
        \multirow{2}{*}{Methods} & \multicolumn{3}{c}{CIFAR-10} & \multicolumn{3}{c}{CIFAR-100} & \multicolumn{3}{c}{MNIST} \\ \cline{2-10}
         & {$Acc_r\uparrow$} & {$Acc_f\downarrow$} & {$H\uparrow$} & {$Acc_r\uparrow$} & {$Acc_f\downarrow$} & {$H\uparrow$} & {$Acc_r\uparrow$} & {$Acc_f\downarrow$} & {$H\uparrow$} \\ \hline
         Original & 0.9538 & 0.9538 & 0.0881 & 0.7941515151515152 & 0.805 & 0.3131 & 0.9974000000000001 & 0.9974000000000001 & 0.0052 \\
         Retrain & 0.9545333333333333 & 0.0 & 0.9767 & 0.798060606060606 & 0.0 & 0.8877 & 0.9969888888888889 & 0.0 & 0.9985 \\ \hline
         Shrink & 0.9242222222222223 & 0.1329 & 0.8948 & 0.7100757575757576 & 0.0355 & 0.8180 & 0.9828888888888889 & 0.333 & 0.7947 \\
         Expand & 0.9320777777777778 & 0.1273 & 0.9014 & 0.7099494949494949 & 0.0335 & 0.8186 & 0.9854888888888889 & 0.1782 & 0.8962 \\
         BT & 0.8739666666666667 & 0.0632 & 0.9043 & 0.6720050505050505 & 0.037 & 0.7916 & 0.8122999999999999 & 0.053799999999999994 & 0.8742 \\
         DELETE & 0.9416777777777778 & 0.0009000000000000001 & \bfseries 0.9695 & 0.7272828282828283 & 0.01 & 0.8385 & \bfseries 0.9925444444444445 & \bfseries 0.0 & \bfseries 0.9963 \\
         EWC & 0.9209666666666667 & \bfseries 0.0 & 0.9589 & \bfseries 0.7598484848484849 & \bfseries 0.0 & \bfseries 0.8635 & 0.9842444444444445 & \bfseries 0.0 & 0.9921 \\
         LF-SSD & \bfseries 0.9518666666666666 & 0.016 & 0.9677 & 0.7591919191919192 & \bfseries 0.0 & 0.8631 & 0.9892555555555556 & \bfseries 0.0 & 0.9946 \\
         RL & 0.8918666666666667 & 0.0116 & 0.9377 & 0.6972727272727273 & 0.022 & 0.8141 & 0.9013777777777777 & 0.0001 & 0.9481 \\ \hline
         Notop & 0.8960444444 & \bfseries 0.0000000000 & 0.9452000000 & 0.7045909091 & 0.0045000000 & 0.8252000000 & 0.9414444444 & \bfseries 0.0000000000 & 0.9698000000 \\
         Conf. & 0.7635444444 & 0.1297000000 & 0.8134000000 & 0.6912424242 & 0.0495000000 & 0.8004000000 & 0.9174777778 & 0.0221000000 & 0.9467000000 \\ \hline
    \end{tabular}

\end{table}

\begin{table}[]
    \centering
    \caption{EfficientNetV2-S Single Class Removal}
    \label{tab:EffSingleClassRemoval}
    \begin{tabular}{c|S[table-format=0.4] S[table-format=0.4] S[table-format=0.4]|S[table-format=0.4] S[table-format=0.4] S[table-format=0.4]|S[table-format=0.4] S[table-format=0.4] S[table-format=0.4]} \hline
        \multirow{2}{*}{Methods} & \multicolumn{3}{c}{CIFAR-10} & \multicolumn{3}{c}{CIFAR-100} & \multicolumn{3}{c}{MNIST} \\ \cline{2-10}
         & {$Acc_r\uparrow$} & {$Acc_f\downarrow$} & {$H\uparrow$} & {$Acc_r\uparrow$} & {$Acc_f\downarrow$} & {$H\uparrow$} & {$Acc_r\uparrow$} & {$Acc_f\downarrow$} & {$H\uparrow$} \\ \hline
         Original & 0.9724000000 & 0.9724000000 & 0.0537000000 & 0.8481818182 & 0.8500000000 & 0.2549000000 & 0.9972000000 & 0.9972000000 & 0.0056000000 \\
         Retrain & 0.9727000000 & 0.0000000000 & 0.9862000000 & 0.8531969697 & 0.0000000000 & 0.9208000000 & 0.9969444444 & 0.0000000000 & 0.9985000000 \\ \hline
         Shrink & 0.9215555556 & 0.0997000000 & 0.9108000000 & 0.7603636364 & 0.0720000000 & 0.8359000000 & 0.9718666667 & 0.4994000000 & 0.6608000000 \\
         Expand & \bfseries 0.9545000000 & 0.2998000000 & 0.8078000000 & 0.7601919192 & 0.0705000000 & 0.8364000000 & 0.9763777778 & 0.3536000000 & 0.7778000000 \\
         BT & 0.8298777778 & 0.0312000000 & 0.8940000000 & 0.6593383838 & 0.0300000000 & 0.7851000000 & 0.9567555556 & 0.0507000000 & 0.9530000000 \\
         DELETE & 0.9520333333 & 0.0001000000 & \bfseries 0.9754000000 & \bfseries 0.7847777778 & 0.0150000000 & \bfseries 0.8736000000 & 0.9871222222 & 0.0001000000 & 0.9935000000 \\
         EWC & 0.9376222222 & \bfseries 0.0000000000 & 0.9678000000 & 0.7773131313 & 0.0055000000 & 0.8726000000 & 0.9845222222 & \bfseries 0.0000000000 & 0.9922000000 \\
         LF-SSD & 0.0968000000 & 0.1597000000 & 0.1736000000 & 0.0103131313 & \bfseries 0.0000000000 & 0.0204000000 & 0.0984333333 & 0.1000000000 & 0.1775000000 \\
         RL & 0.9468222222 & \bfseries 0.0000000000 & 0.9727000000 & 0.7453585859 & 0.0480000000 & 0.8361000000 & \bfseries 0.9897444444 & 0.0002000000 & \bfseries 0.9947000000 \\ \hline
         Notop & 0.9404222222 & \bfseries 0.0000000000 & 0.9693000000 & 0.7572070707 & 0.0055000000 & 0.8598000000 & 0.9857888889 & \bfseries 0.0000000000 & 0.9928000000 \\
         Conf. & 0.8553222222 & 0.2207000000 & 0.8155000000 & 0.4513484848 & 0.0445000000 & 0.6131000000 & 0.9751333333 & 0.3624000000 & 0.7710000000 \\ \hline
    \end{tabular}

\end{table}

In single-class unlearning, DISCERN consistently meets its primary objective: in virtually all architecture--dataset combinations it reduces $Acc_f$ to values close to 0, effectively forgetting the target class. In terms of $H$, it systematically ranks in the upper half of the comparison, typically between third and fifth among the nine methods, and is never the worst in any combination; on Swin-T with CIFAR-100 it is the best unlearning method, less than $0.001$ in $H$ below retraining. When another method obtains a higher $H$, mainly DELETE, Random Labeling, or Bad Teacher, the margin is small, at most about four hundredths, so DISCERN stays close to the best result in every combination even though it is designed for the incremental setting. Overall, DISCERN is a competitive and reliable method for single-class unlearning.

\begin{figure}
    \centering
    \includegraphics[width=\linewidth]{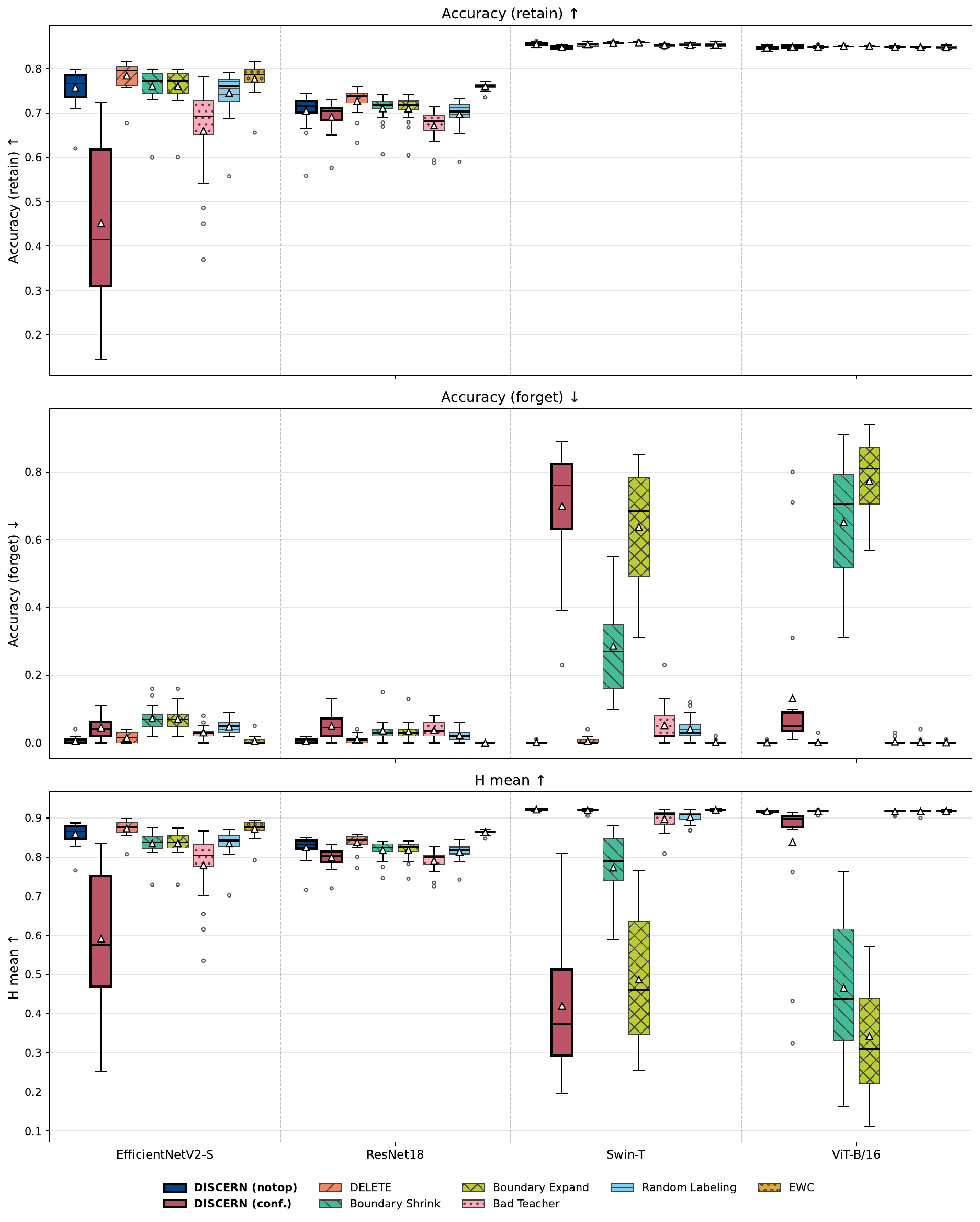}
    \caption{Distribution of $Acc_r$ and $Acc_f$ per method over the 20 individually forgotten classes of CIFAR-100, by architecture.}
    \label{fig:Boxplot}
\end{figure}

Figure~\ref{fig:Boxplot} shows, for CIFAR-100, the distribution of $Acc_r$ and $Acc_f$ of each method over the 20 single-class runs, broken down by architecture. The $Acc_f$ box of DISCERN stays close to 0 with very little spread across forgotten classes, while its $Acc_r$ box, although somewhat lower than that of methods such as DELETE, shows no extreme outliers, indicating consistent behavior across runs rather than merely a good average.

\textbf{Multi-class unlearning.} Tables~\ref{tab:MultipleClassRemoval}, \ref{tab:MultipleClassRemovalSwinT}, \ref{tab:MultipleClassRemovalResnet18}, and~\ref{tab:MultipleClassRemovalEffi} report, for each architecture, the effectiveness of the methods when removing groups of 5, 10, 20, and 50 classes at once from CIFAR-100, showing how each method degrades as the fraction of the dataset to be forgotten grows.
\sisetup{
    round-mode = places,
    round-precision = 3
}
\begin{table}[]
    \centering
    \setlength{\tabcolsep}{3.5pt}
    \caption{ViT-B/16 Multiple Class Removal}
    \label{tab:MultipleClassRemoval}
    \begin{tabular}{c|S[table-format=0.3] S[table-format=0.3] S[table-format=0.3]|S[table-format=0.3] S[table-format=0.3] S[table-format=0.3]|S[table-format=0.3] S[table-format=0.3] S[table-format=0.3]|S[table-format=0.3] S[table-format=0.3] S[table-format=0.3]} \hline
        \multirow{2}{*}{Methods} & \multicolumn{3}{c}{5 classes} & \multicolumn{3}{c}{10 classes} & \multicolumn{3}{c}{20 classes} & \multicolumn{3}{c}{50 classes} \\ \cline{2-13}
         & {$Acc_r\uparrow$} & {$Acc_f\downarrow$} & {$H\uparrow$} & {$Acc_r\uparrow$} & {$Acc_f\downarrow$} & {$H\uparrow$} & {$Acc_r\uparrow$} & {$Acc_f\downarrow$} & {$H\uparrow$} & {$Acc_r\uparrow$} & {$Acc_f\downarrow$} & {$H\uparrow$} \\ \hline
         Original & .8497473684210526 & .8608 & .2392 & .8505555555555556 & .8480000000000001 & .2579 & .8502625 & .85045 & .2544 & .84346 & .85714 & .2443 \\
         Retrain & .8516736842105264 & .0 & .9199 & .8574555555555555 & .0 & .9233 & .86365 & .0 & .9268 & .8896599999999999 & .0 & .9416 \\ \hline
         Shrink & .8489052631578946 & .4702 & .6524 & .8432000000000001 & .3711 & .7205 & .832825 & .38835000000000003 & .7053 & .73338 & .30304000000000003 & .7147 \\
         Expand & \bfseries .8513894736842106 & .7969999999999999 & .3278 & \bfseries .8531555555555557 & .8077 & .3139 & .853875 & .8272999999999999 & .2873 & .84866 & .84986 & .2551 \\
         BT & .8430105263157894 & .05 & .8933 & .7919111111111111 & .0374 & .8690 & .68765 & .04395 & .7999 & .42425999999999997 & .05132 & .5863 \\
         DELETE & .8439578947368421 & .0012000000000000001 & \bfseries .9149 & .8459888888888889 & .0008 & \bfseries .9162 & .848225 & \bfseries .00045000000000000004 & \bfseries .9177 & .8682000000000001 & \bfseries .00012000000000000002 & .9294 \\
         EWC & .823642105263158 & \bfseries .0 & .9033 & .7877222222222222 & \bfseries .0002 & .8812 & .7390625 & \bfseries .0 & .8500 & .32474 & .01332 & .4887 \\
         LF-SSD & \bfseries .8512736842105264 & .808 & .3133 & .8522000000000001 & .8305999999999999 & .2826 & .8503624999999999 & .85035 & .2545 & 0.84354 & 0.8572599999999999 & 0.2442 \\
         RL & 0.8482526315789475 & 0.0182 & 0.9102 & 0.8449333333333333 & 0.005 & 0.9138 & 0.8434250000000001 & 0.0006000000000000001 & 0.9148 & \bfseries 0.86866 & \bfseries 0.00014 & \bfseries 0.9297 \\ \hline
         Notop & 0.8161684211 & \bfseries 0.0000000000 & 0.8988000000 & 0.8003777778 & \bfseries 0.0000000000 & 0.8891000000 & 0.7185250000 & 0.0044000000 & 0.8347000000 & 0.3458000000 & 0.0125200000 & 0.5122000000 \\
         Conf. & \bfseries 0.8512526316 & 0.1470000000 & 0.8521000000 & \bfseries 0.8528333333 & 0.1347000000 & 0.8590000000 & \bfseries 0.8577750000 & 0.2313000000 & 0.8108000000 & 0.7412200000 & 0.0469800000 & 0.8339000000 \\ \hline
    \end{tabular}

\end{table}

\begin{table}[]
    \centering
    \setlength{\tabcolsep}{3.5pt}
    \caption{Swin-T Multiple Class Removal}
    \label{tab:MultipleClassRemovalSwinT}
    \begin{tabular}{c|S[table-format=0.3] S[table-format=0.3] S[table-format=0.3]|S[table-format=0.3] S[table-format=0.3] S[table-format=0.3]|S[table-format=0.3] S[table-format=0.3] S[table-format=0.3]|S[table-format=0.3] S[table-format=0.3] S[table-format=0.3]} \hline
        \multirow{2}{*}{Methods} & \multicolumn{3}{c}{5 classes} & \multicolumn{3}{c}{10 classes} & \multicolumn{3}{c}{20 classes} & \multicolumn{3}{c}{50 classes} \\ \cline{2-13}
         & {$Acc_r\uparrow$} & {$Acc_f\downarrow$} & {$H\uparrow$} & {$Acc_r\uparrow$} & {$Acc_f\downarrow$} & {$H\uparrow$} & {$Acc_r\uparrow$} & {$Acc_f\downarrow$} & {$H\uparrow$} & {$Acc_r\uparrow$} & {$Acc_f\downarrow$} & {$H\uparrow$} \\ \hline
         Original & 0.8583473684 & 0.8594000000 & 0.2416000000 & 0.8589000000 & 0.8539000000 & 0.2497000000 & 0.8591125000 & 0.8555500000 & 0.2473000000 & 0.8516600000 & 0.8651400000 & 0.2328000000 \\
         Retrain & 0.8592631579 & 0.0000000000 & 0.9243000000 & 0.8608888889 & 0.0000000000 & 0.9252000000 & 0.8688500000 & 0.0000000000 & 0.9298000000 & 0.8910400000 & 0.0000000000 & 0.9424000000 \\ \hline
         Shrink & 0.8547052632 & 0.2804000000 & 0.7814000000 & 0.8490555556 & 0.2827000000 & 0.7776000000 & 0.8274125000 & 0.3002500000 & 0.7582000000 & 0.6898200000 & 0.3333800000 & 0.6780000000 \\
         Expand & \bfseries 0.8612105263 & 0.7416000000 & 0.3975000000 & \bfseries 0.8626222222 & 0.7825000000 & 0.3474000000 & \bfseries 0.8646625000 & 0.8158500000 & 0.3036000000 & 0.8596800000 & 0.8527000000 & 0.2515000000 \\
         BT & 0.7744736842 & 0.0414000000 & 0.8568000000 & 0.6452222222 & 0.0510000000 & 0.7682000000 & 0.4253875000 & 0.0401000000 & 0.5895000000 & 0.1920600000 & 0.0395800000 & 0.3201000000 \\
         DELETE & 0.8388421053 & 0.0016000000 & \bfseries 0.9117000000 & 0.8377444444 & 0.0011000000 & \bfseries 0.9113000000 & 0.8394375000 & 0.0029500000 & \bfseries 0.9115000000 & \bfseries 0.8697000000 & 0.0009400000 & \bfseries 0.9299000000 \\
         EWC & 0.8095578947 & \bfseries 0.0000000000 & 0.8948000000 & 0.7873333333 & \bfseries 0.0002000000 & 0.8809000000 & 0.6940125000 & 0.0009500000 & 0.8191000000 & 0.1876000000 & 0.0018800000 & 0.3158000000 \\
         LF-SSD & 0.6475157895 & \bfseries 0.0002000000 & 0.7860000000 & 0.6599333333 & 0.0366000000 & 0.7833000000 & 0.6793125000 & 0.1662500000 & 0.7486000000 & 0.5871200000 & 0.4018200000 & 0.5926000000 \\
         RL & 0.8418315789 & 0.0198000000 & 0.9058000000 & 0.8283000000 & 0.0078000000 & 0.9029000000 & 0.7979250000 & \bfseries 0.0004000000 & 0.8874000000 & 0.8140000000 & \bfseries 0.0000000000 & 0.8975000000 \\ \hline
         Notop & 0.8141263158 & \bfseries 0.0000000000 & 0.8975000000 & 0.7773000000 & \bfseries 0.0001000000 & 0.8747000000 & 0.6472875000 & \bfseries 0.0003500000 & 0.7858000000 & 0.3212600000 & \bfseries 0.0000000000 & 0.4863000000 \\
         Conf. & 0.8215368421 & 0.2602000000 & 0.7785000000 & 0.8078111111 & 0.0771000000 & 0.8615000000 & 0.8035375000 & 0.0644000000 & 0.8646000000 & 0.8111600000 & 0.0490800000 & 0.8755000000 \\ \hline
    \end{tabular}

\end{table}

\begin{table}[]
    \centering
    \setlength{\tabcolsep}{3.5pt}
    \caption{ResNet18 Multiple Class Removal}
    \label{tab:MultipleClassRemovalResnet18}
    \begin{tabular}{c|S[table-format=0.3] S[table-format=0.3] S[table-format=0.3]|S[table-format=0.3] S[table-format=0.3] S[table-format=0.3]|S[table-format=0.3] S[table-format=0.3] S[table-format=0.3]|S[table-format=0.3] S[table-format=0.3] S[table-format=0.3]} \hline
        \multirow{2}{*}{Methods} & \multicolumn{3}{c}{5 classes} & \multicolumn{3}{c}{10 classes} & \multicolumn{3}{c}{20 classes} & \multicolumn{3}{c}{50 classes} \\ \cline{2-13}
         & {$Acc_r\uparrow$} & {$Acc_f\downarrow$} & {$H\uparrow$} & {$Acc_r\uparrow$} & {$Acc_f\downarrow$} & {$H\uparrow$} & {$Acc_r\uparrow$} & {$Acc_f\downarrow$} & {$H\uparrow$} & {$Acc_r\uparrow$} & {$Acc_f\downarrow$} & {$H\uparrow$} \\ \hline
         Original & 0.7940947368421052 & 0.7982 & 0.3218 & 0.7945 & 0.7925000000000001 & 0.3291 & 0.7953875 & 0.7899499999999999 & 0.3323 & 0.7881199999999999 & 0.80048 & 0.3184 \\
         Retrain & 0.7997578947368421 & 0.0 & 0.8887 & 0.8062444444444445 & 0.0 & 0.8927 & 0.8177749999999999 & 0.0 & 0.8998 & 0.84716 & 0.0 & 0.9173 \\ \hline
         Shrink & 0.734178947368421 & 0.0928 & 0.8116 & 0.7129444444444444 & 0.0759 & 0.8049 & 0.65045 & 0.0872 & 0.7596 & 0.43608 & 0.12928 & 0.5811 \\
         Expand & 0.7611263157894737 & 0.1514 & 0.8025 & 0.7747222222222223 & 0.21880000000000002 & 0.7779 & 0.789725 & 0.36725 & 0.7026 & 0.81716 & 0.63712 & 0.5026 \\
         BT & 0.7111368421052632 & 0.002 & 0.8305 & 0.6614666666666666 & 0.0031 & 0.7953 & 0.3772375 & 0.00795 & 0.5466 & 0.09586 & 0.018840000000000003 & 0.1747 \\
         DELETE & 0.7707894736842105 & 0.002 & \bfseries 0.8698 & 0.7792111111111111 & 0.004 & \bfseries 0.8744 & 0.7895 & 0.00545 & \bfseries 0.8802 & \bfseries 0.8236800000000001 & 0.00536 & \bfseries 0.9011 \\
         EWC & 0.7476947368421053 & \bfseries 0.0 & 0.8556 & 0.7542333333333333 & \bfseries 0.0 & 0.8599 & 0.6095125 & 0.0008 & 0.7572 & 0.10064 & 0.0028799999999999997 & 0.1828 \\
         LF-SSD & \bfseries 0.7973263157894737 & 0.3238 & 0.7318 & \bfseries 0.7945777777777778 & 0.7915 & 0.3303 & \bfseries 0.7952874999999999 & 0.78985 & 0.3325 & 0.78802 & 0.80038 & 0.3185 \\
         RL & 0.7259578947368421 & 0.004399999999999999 & 0.8397 & 0.7052555555555555 & 0.0031 & 0.8261 & 0.6564625 & \bfseries 0.0004 & 0.7925 & 0.58396 & \bfseries 0.0 & 0.7373 \\ \hline
         Notop & 0.7204526316 & \bfseries 0.0000000000 & 0.8375000000 & 0.6875111111 & \bfseries 0.0000000000 & 0.8148000000 & 0.6109000000 & \bfseries 0.0000000000 & 0.7585000000 & 0.5165800000 & \bfseries 0.0000000000 & 0.6812000000 \\
         Conf. & 0.5623052632 & 0.1156000000 & 0.6875000000 & 0.4912666667 & 0.0713000000 & 0.6426000000 & 0.4864500000 & 0.0419000000 & 0.6453000000 & 0.4769400000 & 0.0248400000 & 0.6406000000 \\ \hline
    \end{tabular}

\end{table}

\begin{table}[]
    \centering
    \setlength{\tabcolsep}{3.5pt}
    \caption{EfficientNetV2-S  Multiple Class Removal}
    \label{tab:MultipleClassRemovalEffi}
    \begin{tabular}{c|S[table-format=0.3] S[table-format=0.3] S[table-format=0.3]|S[table-format=0.3] S[table-format=0.3] S[table-format=0.3]|S[table-format=0.3] S[table-format=0.3] S[table-format=0.3]|S[table-format=0.3] S[table-format=0.3] S[table-format=0.3]} \hline
        \multirow{2}{*}{Methods} & \multicolumn{3}{c}{5 classes} & \multicolumn{3}{c}{10 classes} & \multicolumn{3}{c}{20 classes} & \multicolumn{3}{c}{50 classes} \\ \cline{2-13}
         & {$Acc_r\uparrow$} & {$Acc_f\downarrow$} & {$H\uparrow$} & {$Acc_r\uparrow$} & {$Acc_f\downarrow$} & {$H\uparrow$} & {$Acc_r\uparrow$} & {$Acc_f\downarrow$} & {$H\uparrow$} & {$Acc_r\uparrow$} & {$Acc_f\downarrow$} & {$H\uparrow$} \\ \hline
         Original & 0.8476000000 & 0.8596000000 & 0.2409000000 & 0.8476555556 & 0.8531000000 & 0.2504000000 & 0.8481625000 & 0.8483500000 & 0.2573000000 & 0.8420600000 & 0.8543400000 & 0.2484000000 \\
         Retrain & 0.8560105263 & 0.0000000000 & 0.9224000000 & 0.8589555556 & 0.0000000000 & 0.9241000000 & 0.8662125000 & 0.0000000000 & 0.9283000000 & 0.8876800000 & 0.0000000000 & 0.9405000000 \\ \hline
         Shrink & 0.8075578947 & 0.3264000000 & 0.7345000000 & 0.7702666667 & 0.2806000000 & 0.7440000000 & 0.5580000000 & 0.1614500000 & 0.6701000000 & 0.1777200000 & 0.0702000000 & 0.2984000000 \\
         Expand & \bfseries 0.8257684211 & 0.4450000000 & 0.6638000000 & \bfseries 0.8365555556 & 0.5708000000 & 0.5673000000 & \bfseries 0.8502750000 & 0.6945500000 & 0.4494000000 & \bfseries 0.8612400000 & 0.8035600000 & 0.3199000000 \\
         BT & 0.5251789474 & 0.0266000000 & 0.6823000000 & 0.3120888889 & 0.0307000000 & 0.4722000000 & 0.1502375000 & 0.0219000000 & 0.2605000000 & 0.0457600000 & 0.0194000000 & 0.0874000000 \\
         DELETE & 0.8209473684 & 0.0022000000 & \bfseries 0.9008000000 & 0.8238444444 & 0.0027000000 & \bfseries 0.9023000000 & 0.8323375000 & 0.0018500000 & \bfseries 0.9077000000 & 0.8537200000 & \bfseries 0.0004200000 & \bfseries 0.9209000000 \\
         EWC & 0.8115789474 & 0.0006000000 & 0.8957000000 & 0.8090666667 & 0.0027000000 & 0.8934000000 & 0.7766125000 & 0.0138000000 & 0.8689000000 & 0.4205200000 & 0.0118400000 & 0.5900000000 \\
         LF-SSD & 0.0098736842 & 0.0132000000 & 0.0196000000 & 0.0063222222 & 0.0489000000 & 0.0126000000 & 0.0065750000 & 0.0246000000 & 0.0131000000 & 0.0024600000 & 0.0176600000 & 0.0049000000 \\
         RL & 0.7609684211 & 0.0280000000 & 0.8536000000 & 0.7117444444 & 0.0230000000 & 0.8235000000 & 0.7143875000 & 0.0125000000 & 0.8290000000 & 0.4519000000 & \bfseries 0.0000200000 & 0.6225000000 \\ \hline
         Notop & 0.8064631579 & \bfseries 0.0000000000 & 0.8929000000 & 0.7965666667 & \bfseries 0.0001000000 & 0.8867000000 & 0.7504250000 & \bfseries 0.0000000000 & 0.8574000000 & 0.5756800000 & \bfseries 0.0000000000 & 0.7307000000 \\
         Conf. & 0.3576631579 & 0.0396000000 & 0.5212000000 & 0.2691333333 & 0.0512000000 & 0.4193000000 & 0.2449125000 & 0.0463000000 & 0.3897000000 & 0.1587400000 & 0.0522600000 & 0.2719000000 \\ \hline
    \end{tabular}

\end{table}

When moving from one to several classes, the pattern changes slightly: DELETE becomes the method with the highest $H$ in fifteen of the sixteen combinations of architecture and group size, the exception being ViT-B/16 with 50 classes, where Random Labeling exceeds it by $0.0003$. DISCERN remains in a stable leading group, with its best variant usually between second and fourth among the nine methods, and still achieves a near-zero $Acc_f$ even when 50 classes are forgotten simultaneously---something Boundary Expand and LF-SSD fail to do, as they barely reduce $Acc_f$ with respect to the original model on several architectures. As expected, the performance of all methods, DISCERN included, degrades as the number of forgotten classes increases; in particular, the notop variant suffers a notable drop in $Acc_r$. Overall, DISCERN is robust and competitive for simultaneous multi-class unlearning, with DELETE as the reference method in this scenario.

\textbf{ICU.} This scenario is where DISCERN shows its largest advantage (Table~\ref{tab:ICU}, ViT-B/16). At the last checkpoint (step 20), DISCERN (conf.) reaches $H=0.797$, second only to retraining ($H=0.926$), the upper bound of the problem, and ahead of every other method: Random Labeling, the second-best competitor, reaches $H=0.765$, whereas Bad Teacher, Boundary Shrink, LF-SSD, Boundary Expand, EWC, and DELETE fall below $H=0.62$, the last two not even exceeding $H=0.06$. This result is specific to the confidence variant: the notop variant degrades progressively as steps accumulate, which supports confidence as the appropriate configuration for this scenario, particularly on transformers.

Since the 20-class group is the same as in the multi-class scenario, comparing both provides a controlled test of the effect of incrementality. DELETE, the best method in the simultaneous setting ($H=0.9177$), collapses to $H=0.0482$ when the same deletion is split into twenty steps, and EWC follows the same path ($0.8500 \rightarrow 0.0580$); neither yields a usable model. Random Labeling absorbs the change with a moderate loss ($0.9148 \rightarrow 0.7653$), and DISCERN (conf.) is by far the most stable method ($0.8108 \rightarrow 0.7972$). The pattern repeats on Swin-T, where DELETE drops from $0.9115$ to $0.2009$ and DISCERN (conf.) only from $0.8646$ to $0.8571$. In other words, many state-of-the-art methods fail not because of the amount of knowledge they must remove, but because they must remove it piecemeal, and this is precisely where our proposal contributes.

This contrast holds only for the two transformers. On ResNet18 and EfficientNetV2-S, the incremental regime degrades all methods without exception, DISCERN included ($0.7585 \rightarrow 0.3654$ on ResNet18 and $0.8574 \rightarrow 0.5766$ on EfficientNetV2-S, both with notop), and twenty unlearning steps remain an open challenge for every evaluated method on these architectures.

\begin{table}[]
    \centering
    \setlength{\tabcolsep}{3.5pt}
    \caption{ViT-B/16 ICU}
    \label{tab:ICU}
    \begin{tabular}{c|S[table-format=0.3] S[table-format=0.3] S[table-format=0.3]|S[table-format=0.3] S[table-format=0.3] S[table-format=0.3]|S[table-format=0.3] S[table-format=0.3] S[table-format=0.3]|S[table-format=0.3] S[table-format=0.3] S[table-format=0.3]} \hline
        \multirow{2}{*}{Methods} & \multicolumn{3}{c}{Step 5} & \multicolumn{3}{c}{Step 10} & \multicolumn{3}{c}{Step 15} & \multicolumn{3}{c}{Step 20} \\ \cline{2-13}
         & {$Acc_r\uparrow$} & {$Acc_f\downarrow$} & {$H\uparrow$} & {$Acc_r\uparrow$} & {$Acc_f\downarrow$} & {$H\uparrow$} & {$Acc_r\uparrow$} & {$Acc_f\downarrow$} & {$H\uparrow$} & {$Acc_r\uparrow$} & {$Acc_f\downarrow$} & {$H\uparrow$} \\ \hline
         Retrain & .8529789269 & .0000000000 & .9206523150 & .8562666714 & .0000000000 & .9225655429 & .8604470789 & .0000000000 & .9249851214 & .8629375398 & .0000000000 & .9264202540 \\ \hline
         Shrink & .8419684052 & .2048000092 & .7995555431 & .7478777885 & .0223000006 & .8461463410 & .5258823693 & \bfseries .0008000000 & .6871873808 & .3282625139 & .0010500000 & .4924308524 \\
         Expand & \bfseries .8518736720 & .7634000421 & .3690020438 & \bfseries .8548888981 & .7504000425 & .3858502350 & \bfseries .8562000096 & .7576666474 & .3775433931 & \bfseries .8582125425 & .7489500284 & .3881410025 \\
         BT & .7724631488 & .1236000046 & .8151671907 & .5827888966 & .0898000042 & .6786142909 & .4875529557 & .0793333320 & .6074829510 & .5149625212 & .1376000036 & .6114687984 \\
         DELETE & .7655894637 & .0020000001 & .8633344471 & .2531000026 & .0030000001 & .3569981447 & .0444705890 & .0029999999 & .0839960029 & .0250000014 & .0013500001 & .0481624294 \\
         EWC & .5959157825 & .0040000001 & .7334050264 & .1370222233 & .0126000006 & .2245390165 & .0452941181 & .0196666669 & .0843723671 & .0317125017 & .0251500015 & .0580244610 \\
         LF-SSD & .7841894627 & .4254000187 & .6624912207 & .7771222293 & .5943000317 & .5325075003 & .7693294346 & .6551999927 & .4756345969 & .7656750441 & .6831500351 & .4478339901 \\
         RL & .8015684068 & .0904000028 & .8478187315 & .7507555574 & .0228000009 & .8368330555 & .7134353116 & \bfseries .0006000000 & .8174076253 & .6524375334 & \bfseries .0003000000 & .7652883707 \\ \hline
         Notop & 0.1873473644 & \bfseries 0.0000000000 & 0.2966159702 & 0.0138555557 & \bfseries 0.0000000000 & 0.0273199844 & 0.0115294120 & \bfseries 0.0009333333 & 0.0227592217 & 0.0120875002 & 0.0039000002 & 0.0237448471 \\
         Conf. & 0.8463684142 & 0.0274000019 & \bfseries 0.9049831911 & 0.8299889028 & 0.1023000065 & \bfseries 0.8615281591 & 0.8069059074 & 0.1657333292 & \bfseries 0.8187459030 & 0.7830125272 & 0.1818000093 & \bfseries 0.7972132053 \\ \hline
    \end{tabular}

\end{table}

Tables~\ref{tab:ICUswint}, \ref{tab:ICUresnet18}, and~\ref{tab:ICUefficientnetv2s} report the ICU results for the remaining architectures.

\begin{table}[]
    \centering
    \setlength{\tabcolsep}{3.5pt}
    \caption{Swin-T ICU}
    \label{tab:ICUswint}
    \begin{tabular}{c|S[table-format=0.3] S[table-format=0.3] S[table-format=0.3]|S[table-format=0.3] S[table-format=0.3] S[table-format=0.3]|S[table-format=0.3] S[table-format=0.3] S[table-format=0.3]|S[table-format=0.3] S[table-format=0.3] S[table-format=0.3]} \hline
        \multirow{2}{*}{Methods} & \multicolumn{3}{c}{Step 5} & \multicolumn{3}{c}{Step 10} & \multicolumn{3}{c}{Step 15} & \multicolumn{3}{c}{Step 20} \\ \cline{2-13}
         & {$Acc_r\uparrow$} & {$Acc_f\downarrow$} & {$H\uparrow$} & {$Acc_r\uparrow$} & {$Acc_f\downarrow$} & {$H\uparrow$} & {$Acc_r\uparrow$} & {$Acc_f\downarrow$} & {$H\uparrow$} & {$Acc_r\uparrow$} & {$Acc_f\downarrow$} & {$H\uparrow$} \\ \hline
         Retrain & 0.8597684145 & 0.0000000000 & 0.9245950560 & 0.8617888987 & 0.0000000000 & 0.9257570025 & 0.8635529578 & 0.0000000000 & 0.9267782080 & 0.8684375405 & 0.0000000000 & 0.9295818483 \\ \hline
         Shrink & 0.8453578889 & 0.0794000033 & 0.8801100162 & 0.7618555605 & 0.0082000004 & 0.8615768612 & 0.5711176574 & 0.0123999996 & 0.7205210819 & 0.4316750228 & 0.0108000004 & 0.5962400637 \\
         Expand & \bfseries 0.8616105020 & 0.6602000296 & 0.4852032964 & \bfseries 0.8645666778 & 0.6510000348 & 0.4962931983 & \bfseries 0.8666470885 & 0.6586666584 & 0.4891124229 & \bfseries 0.8699500501 & 0.6574000239 & 0.4911969849 \\
         BT & 0.8134210289 & 0.1024000048 & 0.8500034668 & 0.6790777802 & 0.1273000063 & 0.7620353532 & 0.4434000120 & 0.0985333337 & 0.5776216638 & 0.3449750155 & 0.1300500059 & 0.4804846743 \\
         DELETE & 0.8250526309 & 0.0010000000 & \bfseries 0.9036979808 & 0.7125444591 & 0.0009000000 & 0.8311571894 & 0.3924117759 & \bfseries 0.0001333333 & 0.5516436070 & 0.1183125047 & \bfseries 0.0000500000 & 0.2009127925 \\
         EWC & 0.8067368269 & \bfseries 0.0002000000 & 0.8925815830 & 0.4598444507 & \bfseries 0.0001000000 & 0.6158169325 & 0.0572117664 & \bfseries 0.0000000000 & 0.1058179300 & 0.0142750006 & \bfseries 0.0000000000 & 0.0281371740 \\
         LF-SSD & 0.3346315712 & 0.0584000029 & 0.4912069583 & 0.2229666680 & 0.0753000032 & 0.3583305766 & 0.1652705938 & 0.0799999986 & 0.2788540525 & 0.1339875057 & 0.0816500023 & 0.2336573709 \\
         RL & 0.8124947190 & 0.0014000001 & 0.8959257385 & 0.6848000050 & \bfseries 0.0004000000 & 0.8124083756 & 0.4842588246 & \bfseries 0.0000666667 & 0.6483509760 & 0.3628375173 & \bfseries 0.0000500000 & 0.5302176677 \\ \hline
         Notop & 0.7504315734 & \bfseries 0.0000000000 & 0.8526808655 & 0.1295555567 & \bfseries 0.0000000000 & 0.2126938904 & 0.0126235297 & \bfseries 0.0000000000 & 0.0249301360 & 0.0125250002 & \bfseries 0.0000000000 & 0.0247401255 \\
         Conf. & 0.8029684067 & 0.0098000004 & 0.8867917365 & 0.7785444498 & 0.0029000001 & \bfseries 0.8743378655 & 0.7671059012 & 0.0012000000 & \bfseries 0.8676913360 & 0.7505375266 & 0.0008000000 & \bfseries 0.8571291261 \\ \hline
    \end{tabular}

\end{table}

\begin{table}[]
    \centering
    \setlength{\tabcolsep}{3.5pt}
    \caption{ResNet18 ICU}
    \label{tab:ICUresnet18}
    \begin{tabular}{c|S[table-format=0.3] S[table-format=0.3] S[table-format=0.3]|S[table-format=0.3] S[table-format=0.3] S[table-format=0.3]|S[table-format=0.3] S[table-format=0.3] S[table-format=0.3]|S[table-format=0.3] S[table-format=0.3] S[table-format=0.3]} \hline
        \multirow{2}{*}{Methods} & \multicolumn{3}{c}{Step 5} & \multicolumn{3}{c}{Step 10} & \multicolumn{3}{c}{Step 15} & \multicolumn{3}{c}{Step 20} \\ \cline{2-13}
         & {$Acc_r\uparrow$} & {$Acc_f\downarrow$} & {$H\uparrow$} & {$Acc_r\uparrow$} & {$Acc_f\downarrow$} & {$H\uparrow$} & {$Acc_r\uparrow$} & {$Acc_f\downarrow$} & {$H\uparrow$} & {$Acc_r\uparrow$} & {$Acc_f\downarrow$} & {$H\uparrow$} \\ \hline
         Retrain & .8006526232 & .0000000000 & .8892871703 & .8053777814 & .0000000000 & .8921919571 & .8106470704 & .0000000000 & .8954184323 & .8175375342 & .0000000000 & .8996043214 \\ \hline
         Shrink & .7193578959 & .5786000222 & .5279438534 & .6896888971 & .6074000120 & .4970680782 & .6979411960 & .6629999816 & .4491267199 & .6848500311 & .6461000323 & .4600548257 \\
         Expand & \bfseries .7198105156 & .5784000307 & .5281580014 & \bfseries .6922222316 & .6097000301 & .4958651716 & \bfseries .7047529578 & .6669333279 & .4470168040 & \bfseries .6907500207 & .6500500321 & .4576997186 \\
         BT & .3752736717 & .2420000091 & .5012401472 & .1375333346 & .0631000024 & .2358999217 & .0960823543 & .0689333318 & .1728225273 & .0602000028 & .0506000025 & .1125431907 \\
         DELETE & .6764841974 & .3122000158 & .6789957230 & .3645777836 & .1416000042 & .4951730812 & .2322588269 & .1093333317 & .3569619450 & .1855875060 & .0958000029 & .3035036433 \\
         EWC & .6691578805 & \bfseries .0000000000 & \bfseries .8016130013 & .3663666688 & \bfseries .0030000001 & .5206098437 & .2185647126 & .0130666664 & .3482329337 & .1447250104 & .0119000003 & .2452889494 \\
         LF-SSD & .0277578941 & \bfseries .0000000000 & .0538211413 & .0342555559 & .0090000007 & .0659128416 & .0306588242 & .0200666658 & .0590798836 & .0322000016 & .0290500018 & .0617906329 \\
         RL & .6362210333 & .3902000129 & .6205922714 & .4758222282 & .1336000057 & .6099155597 & .4083294198 & .0369999994 & \bfseries .5680785812 & .3537250102 & .0105500005 & \bfseries .5185144246 \\ \hline
         Notop & 0.6549999893 & 0.1380000062 & 0.7430015312 & 0.4711000025 & 0.0475000027 & \bfseries 0.6286644866 & 0.3153058857 & 0.0311999993 & 0.4692225841 & 0.2266625121 & 0.0270500014 & 0.3654361701 \\
         Conf. & 0.5213789433 & 0.3726000205 & 0.5493931642 & 0.3589000016 & 0.0389000015 & 0.5203316353 & 0.3163294151 & \bfseries 0.0046666666 & 0.4760843154 & 0.2688625172 & \bfseries 0.0009000000 & 0.4198241993 \\ \hline
    \end{tabular}

\end{table}

\begin{table}[]
    \centering
    \setlength{\tabcolsep}{3.5pt}
    \caption{EfficientNetV2-S ICU}
    \label{tab:ICUefficientnetv2s}
    \begin{tabular}{c|S[table-format=0.3] S[table-format=0.3] S[table-format=0.3]|S[table-format=0.3] S[table-format=0.3] S[table-format=0.3]|S[table-format=0.3] S[table-format=0.3] S[table-format=0.3]|S[table-format=0.3] S[table-format=0.3] S[table-format=0.3]} \hline
        \multirow{2}{*}{Methods} & \multicolumn{3}{c}{Step 5} & \multicolumn{3}{c}{Step 10} & \multicolumn{3}{c}{Step 15} & \multicolumn{3}{c}{Step 20} \\ \cline{2-13}
         & {$Acc_r\uparrow$} & {$Acc_f\downarrow$} & {$H\uparrow$} & {$Acc_r\uparrow$} & {$Acc_f\downarrow$} & {$H\uparrow$} & {$Acc_r\uparrow$} & {$Acc_f\downarrow$} & {$H\uparrow$} & {$Acc_r\uparrow$} & {$Acc_f\downarrow$} & {$H\uparrow$} \\ \hline
         Retrain & 0.8544841886 & 0.0000000000 & 0.9215283700 & 0.8587666750 & 0.0000000000 & 0.9240096007 & 0.8623294294 & 0.0000000000 & 0.9260718049 & 0.8653875470 & 0.0000000000 & 0.9278282633 \\ \hline
         Shrink & 0.7753578782 & 0.6518000245 & 0.4776967304 & 0.7264333487 & 0.6562000275 & 0.4615932659 & 0.7355411887 & 0.6957999885 & 0.4163674858 & 0.6998125255 & 0.6531500340 & 0.4513222186 \\
         Expand & \bfseries 0.7766315699 & 0.6542000294 & 0.4754862456 & \bfseries 0.7366222382 & 0.6672000349 & 0.4545575687 & \bfseries 0.7539647222 & 0.7153999984 & 0.4047213791 & \bfseries 0.7317250311 & 0.6902500331 & 0.4259793651 \\
         BT & 0.0355052628 & 0.0394000013 & 0.0677880090 & 0.0222888888 & \bfseries 0.0061000002 & 0.0432481489 & 0.0236352947 & \bfseries 0.0078666665 & 0.0458967313 & 0.0191125011 & 0.0208000013 & 0.0369285917 \\
         DELETE & 0.5745157838 & 0.1770000063 & 0.6632600215 & 0.2051333349 & 0.0421000018 & 0.3229410213 & 0.0894823551 & 0.0193999994 & 0.1529668231 & 0.0368500020 & \bfseries 0.0055500003 & 0.0685805622 \\
         EWC & 0.7118420899 & 0.1236000046 & \bfseries 0.7848048233 & 0.5346555531 & 0.0779000042 & 0.6741986379 & 0.4459647149 & 0.0604666662 & 0.6015006262 & 0.3466000125 & 0.0428500018 & 0.5044587051 \\
         LF-SSD & 0.0105263153 & \bfseries 0.0000000000 & 0.0208333324 & 0.0033333334 & 0.0700000010 & 0.0065934068 & 0.0023529412 & 0.0533333302 & 0.0046511629 & 0.0025000000 & 0.0400000006 & 0.0049382717 \\
         RL & 0.6939157724 & 0.2910000116 & 0.6989578141 & 0.6134555638 & 0.1701000076 & \bfseries 0.7037727214 & 0.5722470671 & 0.1157333326 & \bfseries 0.6873850871 & 0.4872500151 & 0.0566000027 & \bfseries 0.6363597267 \\ \hline
         Notop & 0.7151684165 & 0.1942000091 & 0.7561439156 & 0.5798888892 & 0.1160000045 & 0.6977720435 & 0.4969411880 & 0.0839999983 & 0.6372644051 & 0.4178750217 & 0.0538000032 & 0.5765943896 \\
         Conf. & 0.0133368419 & 0.0012000000 & 0.0263033007 & 0.0097333335 & 0.0199000008 & 0.0192389796 & 0.0096823531 & 0.0134666659 & 0.0191338885 & 0.0100000001 & 0.0100000001 & 0.0197530867 \\ \hline
    \end{tabular}

\end{table}

\sisetup{
    round-mode = places,
    round-precision = 4
}

\begin{figure}[t]
    \centering
    \begin{subfigure}[b]{0.48\linewidth}
        \centering
        \includegraphics[width=\linewidth]{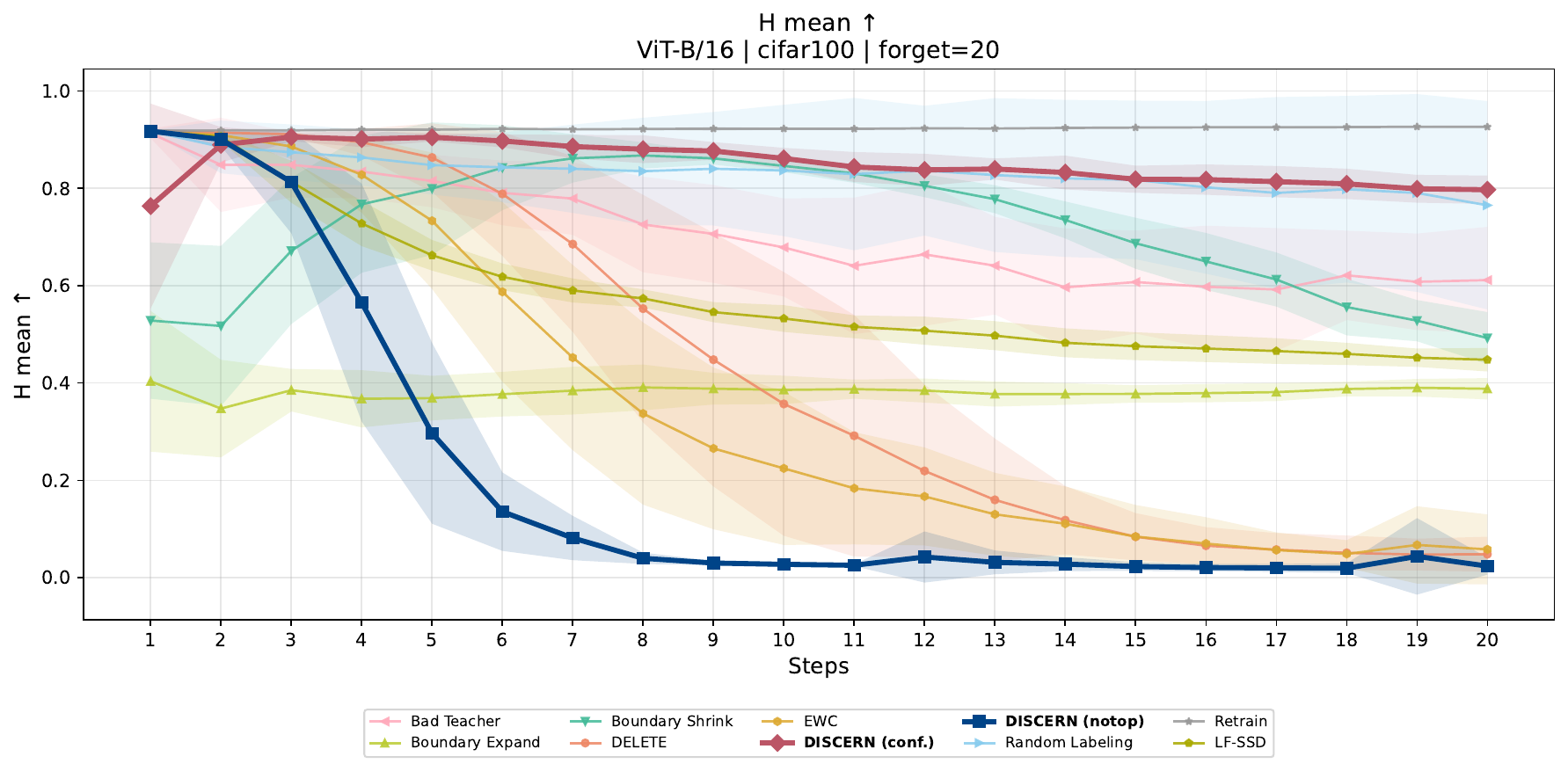}
        \caption{ViT-B/16}
    \end{subfigure}
    \hfill
    \begin{subfigure}[b]{0.48\linewidth}
        \centering
        \includegraphics[width=\linewidth]{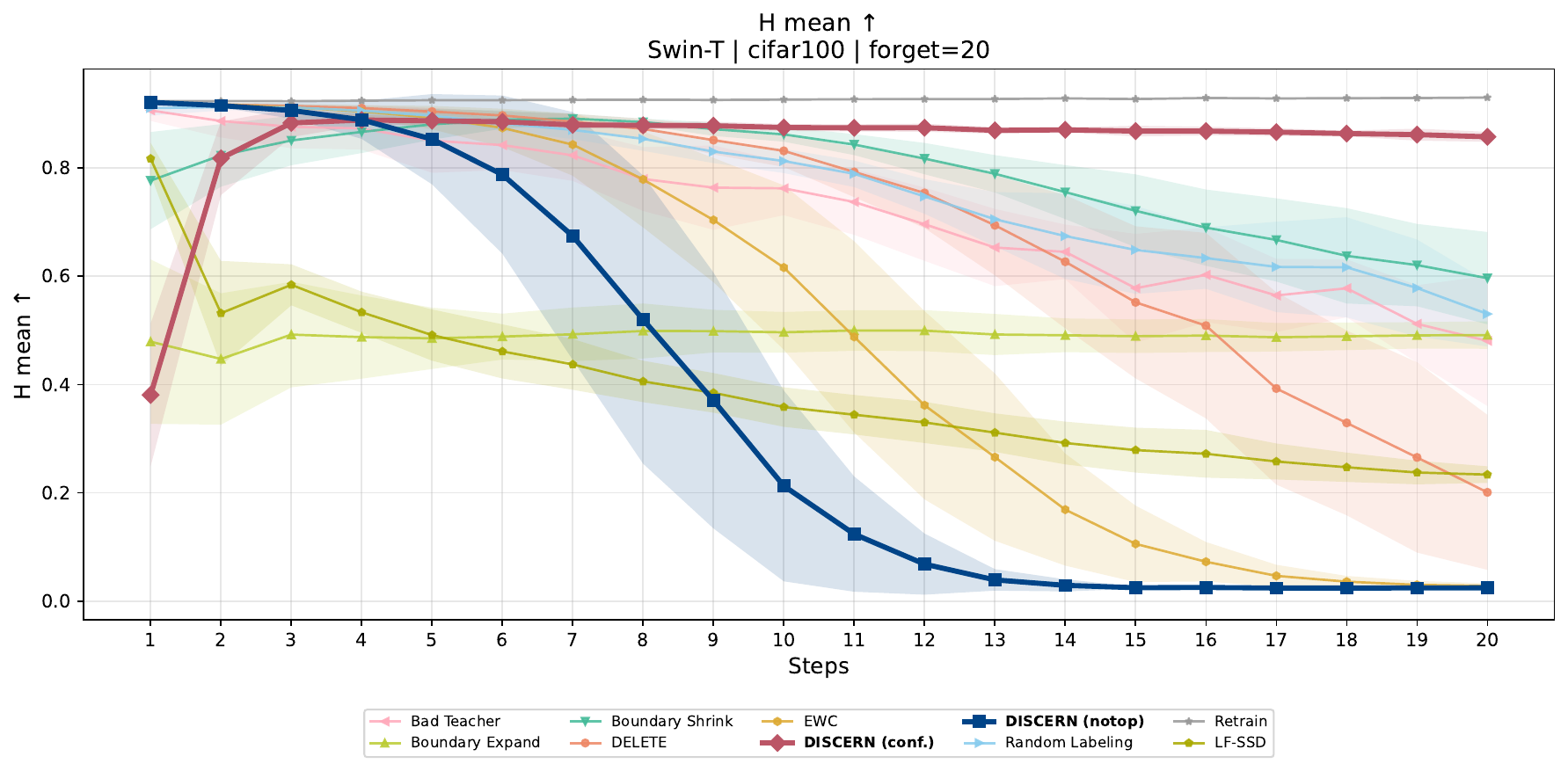}
        \caption{Swin-T}
    \end{subfigure}
    \\[1em]
    \begin{subfigure}[b]{0.48\linewidth}
        \centering
        \includegraphics[width=\linewidth]{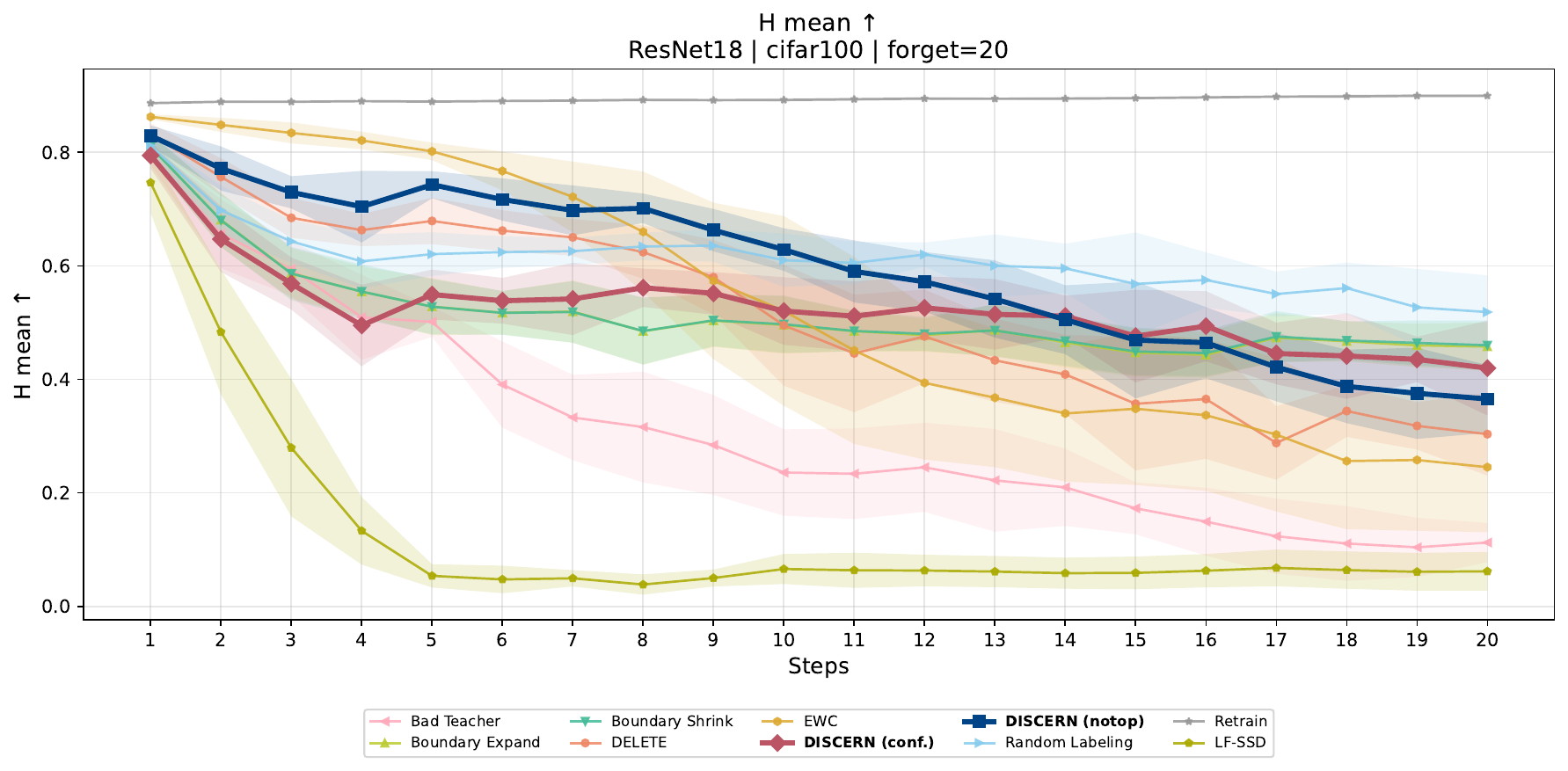}
        \caption{ResNet18}
    \end{subfigure}
    \hfill
    \begin{subfigure}[b]{0.48\linewidth}
        \centering
        \includegraphics[width=\linewidth]{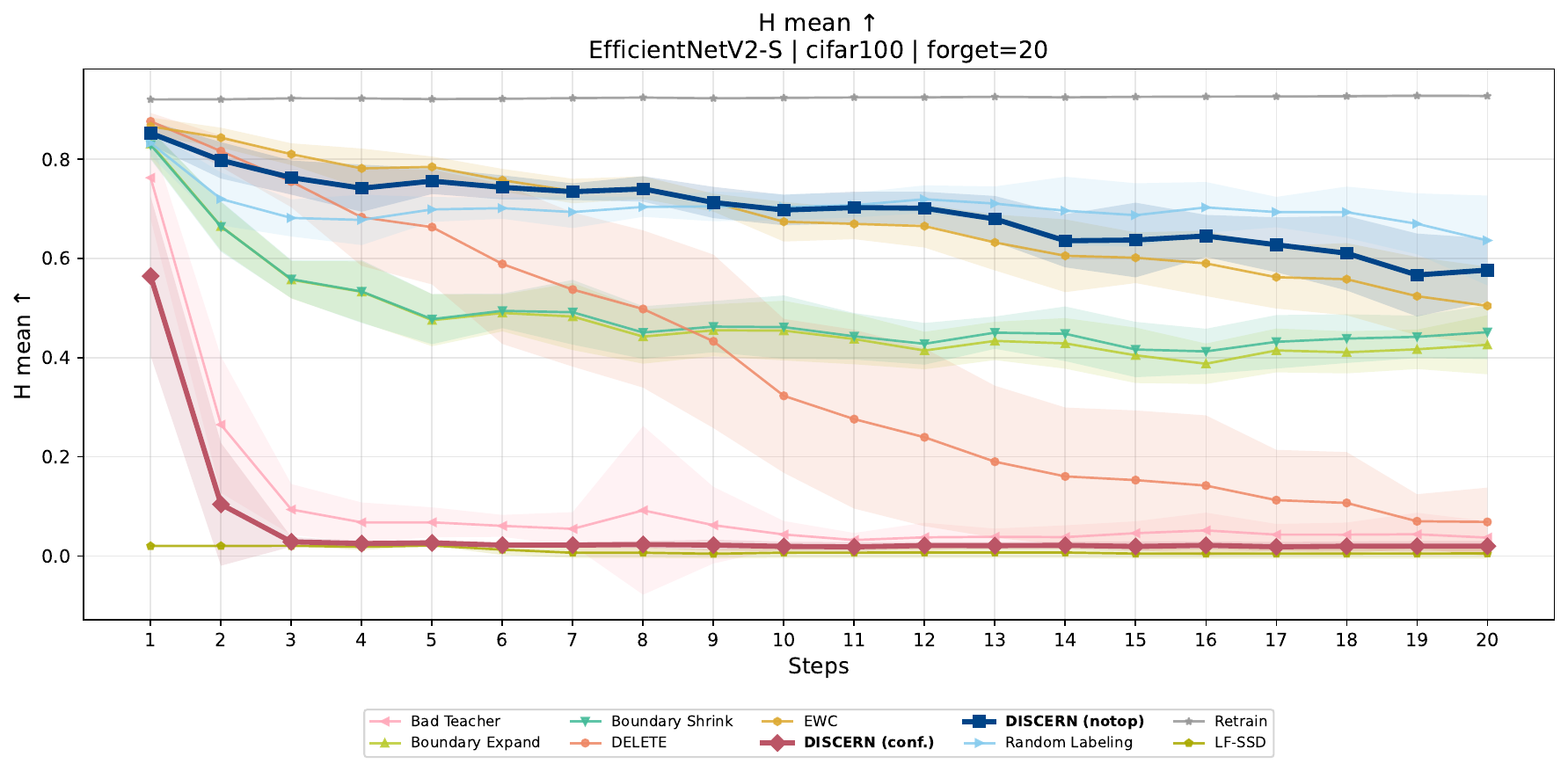}
        \caption{EfficientNetV2-S}
    \end{subfigure}
    \caption{Mean and standard deviation of $H$ over the 20 sequential unlearning steps in ICU (CIFAR-100, 20 classes), by architecture.}
    \label{fig:ICU}
\end{figure}

Figure~\ref{fig:ICU} shows the evolution of $H$ over the 20 sequential unlearning steps for the four architectures. On the transformers, DISCERN (conf.) remains stable and well above the other methods as steps accumulate, whereas on the convolutional networks all methods degrade substantially; on EfficientNetV2-S, the notop variant of DISCERN holds up much better than the confidence variant.

These results indicate that, under our experimental protocol, the difficulty of the incremental scenario cannot be explained solely by the total amount of knowledge removed. The simultaneous and incremental scenarios start from the same model and end with the same set of twenty removed classes, and retraining itself keeps its $H$ virtually unchanged between them. The sharp degradation of methods such as DELETE ($0.9177 \rightarrow 0.0482$) or EWC ($0.8500 \rightarrow 0.0580$) on ViT-B/16 therefore points to the sequential nature of the process and the accumulation of updates as the determining factors. DISCERN (conf.) is much less sensitive to this change of schedule on the transformer architectures evaluated ($0.8108 \rightarrow 0.7972$ on ViT-B/16 and $0.8646 \rightarrow 0.8571$ on Swin-T), although this advantage does not carry over to the convolutional architectures.

\subsection{Ablation study}\label{sec:ablation}

This section is organized in three blocks. The first studies the sensitivity of DISCERN to its two hyperparameters, the number of synthetic embeddings and the margin of the notop loss. The second isolates the contribution of the core of the method, the per-class distribution selection, by comparing it with a variant that assumes the Gaussian for every class. The third evaluates the selection procedure itself, comparing the test cascade of Figure~\ref{fig:flowchart} with a selection based solely on cross-validation.

\subsubsection{Hyperparameter sensitivity}\label{sec:ablationHyper}

DISCERN has two hyperparameters: the number of synthetic embeddings generated in each unlearning process, distributed among the retained classes, and the margin $m$ of the notop loss. We vary each one separately while keeping the other at its default value, $1000$ embeddings and $m=10$, in the single-class unlearning scenario over the four architectures and the three datasets, which amounts to $160$ runs per level. The number of embeddings is studied with both loss functions, whereas the margin only applies to notop.

The replicate is the forgotten class, not the seed: every level is trained on the same architecture, dataset, and class combinations, so the design is paired. As the primary result we report the mean over the forgotten classes with its $95\%$ bootstrap confidence interval, and we compare each configuration with the best one of its scope through the paired difference in $H$ and its $95\%$ bootstrap confidence interval. In addition, each level is compared with the default value through the two-sided paired Wilcoxon test, with Holm correction over the levels of each hyperparameter ($\alpha=0.05$). With $160$ pairs per pooled scope, differences of a few thousandths of $H$ are detectable, so we attend to magnitude and not only to sign.

\begin{figure}[t]
    \centering
    \includegraphics[width=\linewidth]{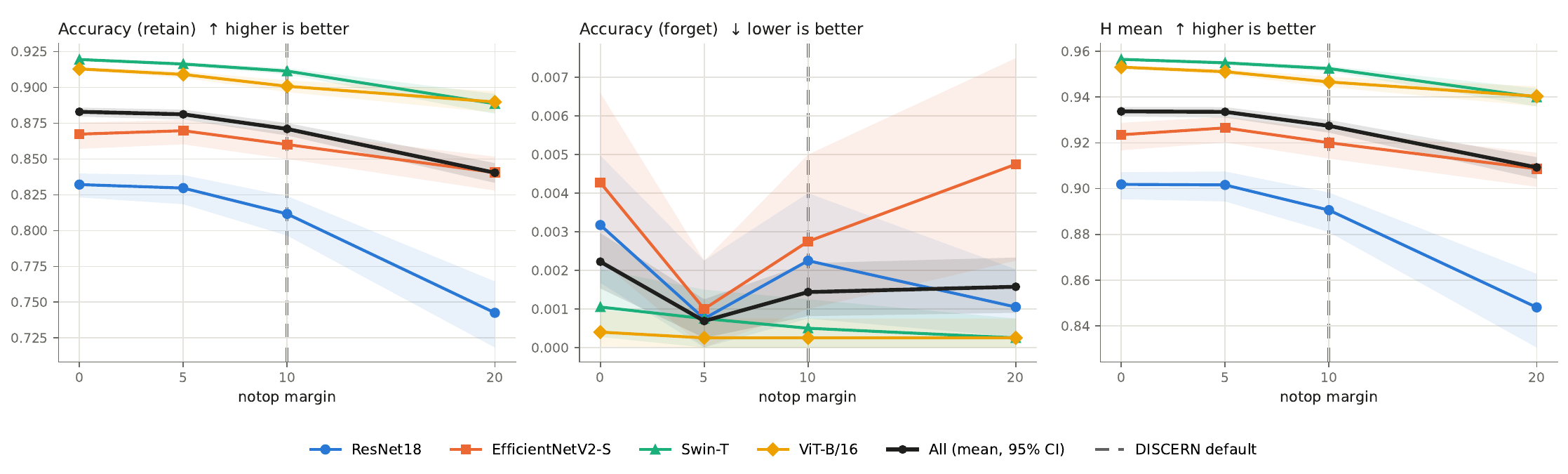}
    \caption{Sensitivity to the margin $m$ of the notop loss in single-class unlearning, pooled over the three datasets: $Acc_r$, $Acc_f$, and $H$ per architecture and their overall mean (black), with $95\%$ bootstrap confidence intervals over the forgotten classes. The dashed line marks the value used in the rest of this work ($m=10$).}
    \label{fig:AblationMargin}
\end{figure}

\textbf{Notop margin.} The margin has little influence on forgetting and somewhat more on retention (Figure~\ref{fig:AblationMargin}). $Acc_f$ is saturated near 0 at every margin; only $m=0$ leaves a slightly larger residual, $0.08$ percentage points above the default value. Beyond that, increasing the margin only costs retention: mean $Acc_r$ drops from $0.883$ at $m=0$ to $0.840$ at $m=20$, and $H$ from $0.934$ to $0.909$, with ResNet18 being the most affected architecture. The value used in the rest of this work ranks third of the four, $-0.006$ in $H$ below the best one ($[-0.008, -0.005]$): it is a conservative, near-optimal choice, although $m=5$ is slightly better and preserves the same level of forgetting.

\begin{figure}[t]
    \centering
    \begin{subfigure}[b]{\linewidth}
        \centering
        \includegraphics[width=\linewidth]{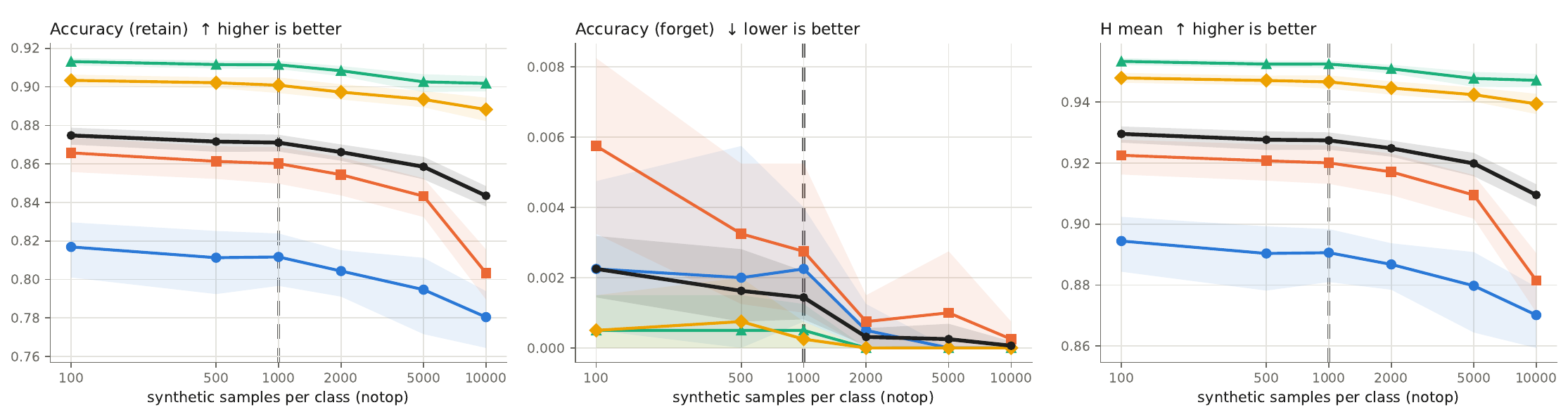}
        \caption{Notop loss.}
    \end{subfigure}
    \\[0.5em]
    \begin{subfigure}[b]{\linewidth}
        \centering
        \includegraphics[width=\linewidth]{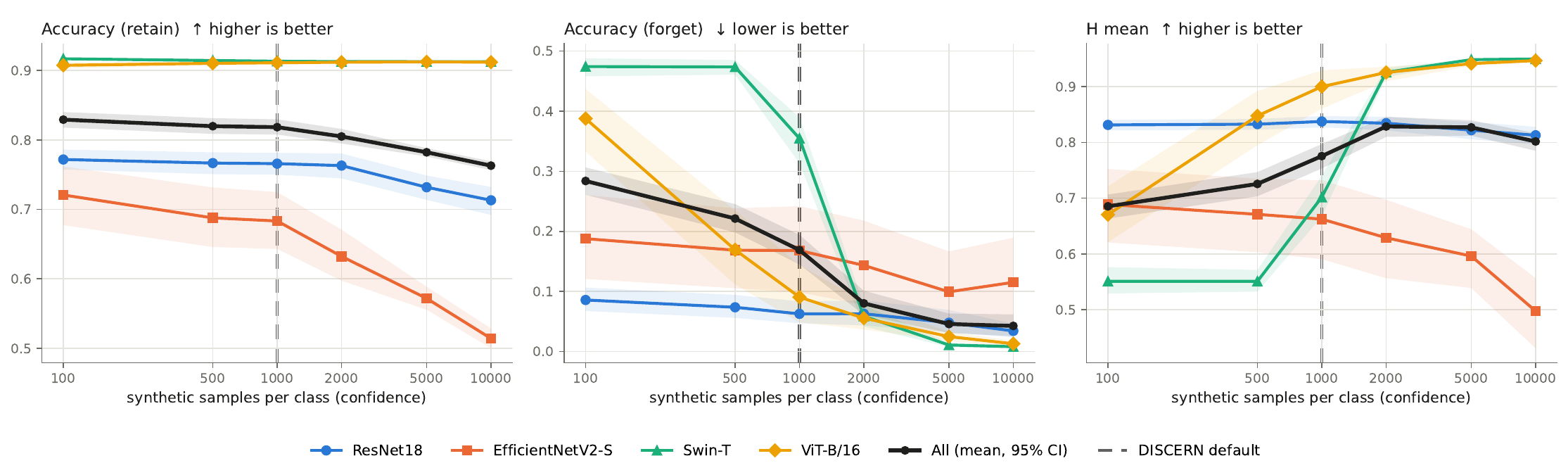}
        \caption{Confidence loss.}
    \end{subfigure}
    \caption{Sensitivity to the number of synthetic embeddings in single-class unlearning, pooled over the three datasets: $Acc_r$, $Acc_f$, and $H$ per architecture and their overall mean (black), with $95\%$ bootstrap confidence intervals over the forgotten classes. The dashed line marks the value used in the rest of this work (1000). With confidence, the two transformer architectures need larger synthetic sets in order to forget, whereas EfficientNetV2-S degrades as they grow.}
    \label{fig:AblationSynthArch}
\end{figure}

\textbf{Number of synthetic embeddings.} The effect of this hyperparameter depends entirely on the loss function (Figure~\ref{fig:AblationSynthArch}). With notop, performance is flat between 100 and 2000 embeddings and degrades only at the extreme values, with mean $Acc_r$ dropping from $0.875$ to $0.843$ at 10000. That degradation concentrates on CIFAR-100, and in particular on EfficientNetV2-S, whose $Acc_r$ falls from $0.757$ to $0.646$ between the default value and 10000. With respect to the best level, the default loses two thousandths of $H$ ($[-0.003, -0.001]$), a practically null difference, so 1000 lies within a wide plateau.

With confidence the situation is different and the default value is not the optimum. Here the size of the synthetic set governs forgetting itself: $Acc_f$ decreases monotonically, from $0.284$ with 100 embeddings to $0.043$ with 10000, at the cost of retention dropping from $0.829$ to $0.763$, which places the maximum of $H$ around 2000. The default ranks fourth of the six levels, with a difference of $-0.053$ in $H$ ($[-0.066, -0.041]$) with respect to that optimum. The gap is not evenly distributed across architectures (Figure~\ref{fig:AblationSynthArch}): on the two transformers with CIFAR-100, small values prevent forgetting, so with 1000 embeddings Swin-T remains at $H=0.419$ with $Acc_f = 0.699$ and ViT-B/16 at $H=0.839$, whereas with 2000 both exceed $0.86$ and with 10000 they reach $0.91$. That value of $Acc_f$ is exactly the one producing the largest discrepancy in Section~\ref{sec:ablationDist}, which indicates that the Swin-T failure does not stem from the distribution used but from a synthetic set that is too small for that combination. EfficientNetV2-S is the opposite case and constitutes a failure mode of its own: its retention collapses as the synthetic set grows, from $0.53$ to $0.12$ on CIFAR-100, dragging $H$ from $0.654$ down to $0.213$.

Overall, with notop the method is largely insensitive to both hyperparameters within a wide range ($m$ between $5$ and $10$, and between 100 and 2000 embeddings), and the default configuration lies within a few thousandths of $H$ of the best level. These differences are small but statistically significant: in $H$, every alternative margin differs significantly from $m=10$ ($m=0$ and $m=5$ are about $0.006$ better and $m=20$ is $0.018$ worse), and so do all numbers of embeddings with notop except 500, with differences between $+0.002$ and $-0.018$. With confidence, 100 and 500 embeddings are significantly worse than the default ($-0.090$ and $-0.050$) and 5000 significantly better ($+0.052$), whereas the gains of 2000 ($+0.053$) and 10000 ($+0.026$) do not survive the Holm correction ($p=0.055$ in both cases) owing to the large variability across classes. With confidence, 1000 embeddings delays forgetting on the transformers, which leaves a residual $Acc_f$ in one-shot removal; in ICU, the scenario in which this loss is used, forgetting completes within the first few steps (Figure~\ref{fig:AblationNoisyICUH}). Raising the value to 2000 is not uniformly better, since it costs retention and clearly harms EfficientNetV2-S. We therefore keep a single value for every architecture, dataset, and loss, without tuning it a posteriori on test results, so the results with confidence reflect a common parameterization rather than the best achievable.

\subsubsection{Distribution selection versus the Gaussian assumption}\label{sec:ablationDist}

To isolate the contribution of distribution selection, we compare DISCERN with an ablated variant, Gaussian, which assumes $E_c \sim \mathcal{N}(\vec\mu_c, \vec\Sigma_c)$ for every class, with $\vec\Sigma_c$ estimated by Ledoit--Wolf. This is the assumption of PASS \cite{zhu2021prototype} with a full covariance. The rest of the procedure is identical (Section~\ref{sec:setup}), and each DISCERN variant is compared with the Gaussian variant that uses the same loss, notop or confidence.

\textbf{Statistical design.} Both methods start from the same original model and forget the same classes in the same order, so the comparison is paired. The unit of analysis is the forgotten class in single-class unlearning ($n=160$: 10 CIFAR-10, 20 CIFAR-100, and 10 MNIST classes for each of the four architectures), the class group in multi-class unlearning ($n=160$: four architectures, four group sizes, and ten seeds), and the whole sequence in ICU ($n=40$: four architectures and ten seeds). In ICU, each sequence is summarized by the area under its metric--step curve, divided by the length of the step range so that it lies in $[0,1]$. For each scenario, loss, and metric we apply the two-sided paired Wilcoxon test, with Holm correction ($\alpha=0.05$) over the three metrics in the aggregate analysis and over metrics and architectures in the per-architecture breakdown. As effect size we report the mean paired difference.

\begin{table}[t]
    \centering
    \small
    \setlength{\tabcolsep}{3.5pt}
    \caption{Aggregate results of both ablations: mean paired difference of DISCERN (D) against Gaussian (G) and against CV-only (Section~\ref{sec:ablationFlow}). $\Delta Acc_r>0$, $\Delta H>0$, and $\Delta Acc_f<0$ favor DISCERN. In bold, differences that are significant under the two-sided paired Wilcoxon test with Holm correction over the three metrics of each comparison ($\alpha=0.05$). Columns D/G and D/CV-only give the number of pairs in which each variant is better in $H$. Since both comparisons share the DISCERN runs, subtracting their mean differences gives the mean difference between CV-only and Gaussian. In ICU the unit is the normalized area under the curve of each sequence.}
    \label{tab:AblationGlobal}
    \begin{tabular}{ll c|rrrc|rrrc} \hline
        \multirow{2}{*}{Scenario} & \multirow{2}{*}{Loss} & \multirow{2}{*}{$n$} & \multicolumn{4}{c|}{DISCERN vs. Gaussian} & \multicolumn{4}{c}{DISCERN vs. CV-only} \\ \cline{4-11}
         & & & $\Delta Acc_r$ & $\Delta Acc_f$ & $\Delta H$ & D/G & $\Delta Acc_r$ & $\Delta Acc_f$ & $\Delta H$ & D/CV-only \\ \hline
        \multirow{2}{*}{Single class} & notop & 160 & $0.000$ & $+0.001$ & $0.000$ & 83/66 & $0.000$ & $0.000$ & $0.000$ & 73/50 \\
         & confidence & 160 & $\mathbf{-0.026}$ & $\mathbf{+0.107}$ & $\mathbf{-0.099}$ & 47/113 & $+0.001$ & $\mathbf{-0.015}$ & $\mathbf{+0.014}$ & 97/55 \\ \hline
        \multirow{2}{*}{Multiple classes} & notop & 160 & $+0.052$ & $0.000$ & $+0.059$ & 88/71 & $0.000$ & $0.000$ & $0.000$ & 70/84 \\
         & confidence & 160 & $\mathbf{+0.058}$ & $\mathbf{+0.058}$ & $+0.022$ & 70/90 & $\mathbf{+0.015}$ & $\mathbf{-0.017}$ & $\mathbf{+0.025}$ & 114/46 \\ \hline
        \multirow{2}{*}{ICU} & notop & 40 & $+0.005$ & $+0.001$ & $+0.004$ & 25/15 & $\mathbf{+0.005}$ & $-0.001$ & $\mathbf{+0.006}$ & 33/7 \\
         & confidence & 40 & $\mathbf{+0.041}$ & $\mathbf{-0.086}$ & $\mathbf{+0.063}$ & 31/9 & $\mathbf{+0.001}$ & $\mathbf{-0.014}$ & $\mathbf{+0.006}$ & 31/9 \\ \hline
    \end{tabular}
\end{table}

\begin{table}[t]
    \centering
    \small
    \setlength{\tabcolsep}{4pt}
    \caption{Paired differences in $H$ by architecture in both ablations: DISCERN against Gaussian (G) and against CV-only. In bold, differences that are significant under the two-sided paired Wilcoxon test with Holm correction over metrics and architectures within each scenario, loss, and comparison. Each cell contains 40 pairs in single- and multi-class unlearning and 10 in ICU.}
    \label{tab:AblationArch}
    \begin{tabular}{ll|rr|rr|rr} \hline
        \multirow{3}{*}{Architecture} & \multirow{3}{*}{Comparison} & \multicolumn{6}{c}{$\Delta H$} \\ \cline{3-8}
         & & \multicolumn{2}{c|}{Single class} & \multicolumn{2}{c|}{Multiple classes} & \multicolumn{2}{c}{ICU} \\ \cline{3-8}
         & & notop & conf. & notop & conf. & notop & conf. \\ \hline
        \multirow{2}{*}{ViT-B/16} & vs. G & $-0.0001$ & $-0.0282$ & $\mathbf{+0.2204}$ & $\mathbf{+0.1670}$ & $-0.0001$ & $\mathbf{+0.1400}$ \\
         & vs. CV-only & $\mathbf{+0.0001}$ & $\mathbf{+0.0115}$ & $+0.0003$ & $\mathbf{+0.0455}$ & $\mathbf{+0.0015}$ & $\mathbf{+0.0159}$ \\ \hline
        \multirow{2}{*}{Swin-T} & vs. G & $-0.0001$ & $-0.1824$ & $+0.0158$ & $+0.0368$ & $+0.0001$ & $\mathbf{+0.0917}$ \\
         & vs. CV-only & $0.0000$ & $\mathbf{+0.0279}$ & $-0.0004$ & $\mathbf{+0.0102}$ & $\mathbf{+0.0072}$ & $+0.0030$ \\ \hline
        \multirow{2}{*}{ResNet18} & vs. G & $-0.0012$ & $\mathbf{-0.0267}$ & $+0.0023$ & $-0.0359$ & $+0.0039$ & $\mathbf{+0.0300}$ \\
         & vs. CV-only & $-0.0002$ & $-0.0003$ & $+0.0004$ & $+0.0010$ & $+0.0041$ & $0.0000$ \\ \hline
        \multirow{2}{*}{EfficientNetV2-S} & vs. G & $-0.0001$ & $\mathbf{-0.1605}$ & $-0.0014$ & $-0.0815$ & $\mathbf{+0.0134}$ & $-0.0081$ \\
         & vs. CV-only & $+0.0001$ & $+0.0189$ & $-0.0014$ & $\mathbf{+0.0427}$ & $+0.0094$ & $+0.0044$ \\ \hline
    \end{tabular}
\end{table}

\textbf{Results with notop.} The two variants are practically indistinguishable (Table~\ref{tab:AblationGlobal}): in single-class unlearning and in ICU no difference is significant and $|\Delta H|$ does not exceed $0.005$. Multi-class unlearning is the only apparent exception, with $\Delta H=+0.059$ in favor of DISCERN, but this mean comes almost entirely from ViT-B/16 ($+0.220$, Table~\ref{tab:AblationArch}), where Gaussian degenerates on a few class groups, and the aggregate test does not survive the Holm correction. By architecture, only two differences with notop are significant, both in favor of DISCERN: ViT-B/16 in multi-class unlearning and EfficientNetV2-S in ICU ($\Delta H=+0.013$).

\textbf{Results with confidence.} With confidence, the direction of the difference depends on the scenario. In single-class unlearning Gaussian is clearly better ($\Delta H=-0.099$): DISCERN leaves a larger residual $Acc_f$ ($\Delta Acc_f=+0.107$) and retains slightly less ($\Delta Acc_r=-0.026$). By architecture, the difference is significant on EfficientNetV2-S ($-0.161$) and ResNet18 ($-0.027$); on Swin-T it is just as large ($-0.182$) but highly variable across classes and does not survive the correction, and on ViT-B/16 it is not significant. Multi-class unlearning shows a trade-off: DISCERN retains more ($\Delta Acc_r=+0.058$) but forgets less ($\Delta Acc_f=+0.058$), and the net effect on $H$ ($+0.022$) is not significant; only ViT-B/16 shows a clear advantage for DISCERN ($\Delta H=+0.167$).

In ICU, the scenario that motivates this work, the trend reverses. DISCERN improves all three metrics ($\Delta H=+0.063$, $\Delta Acc_r=+0.041$, and $\Delta Acc_f=-0.086$) and is better in 31 of the 40 sequences. The advantage is significant on ViT-B/16 ($\Delta H=+0.140$), Swin-T ($+0.092$), and ResNet18 ($+0.030$). The only exception is EfficientNetV2-S ($-0.008$, not significant), where DISCERN with confidence already collapses in the main results (Table~\ref{tab:ICUefficientnetv2s}) and the difference is minimal.

\begin{figure}[t]
    \centering
    \begin{subfigure}[b]{0.48\linewidth}
        \centering
        \includegraphics[width=\linewidth]{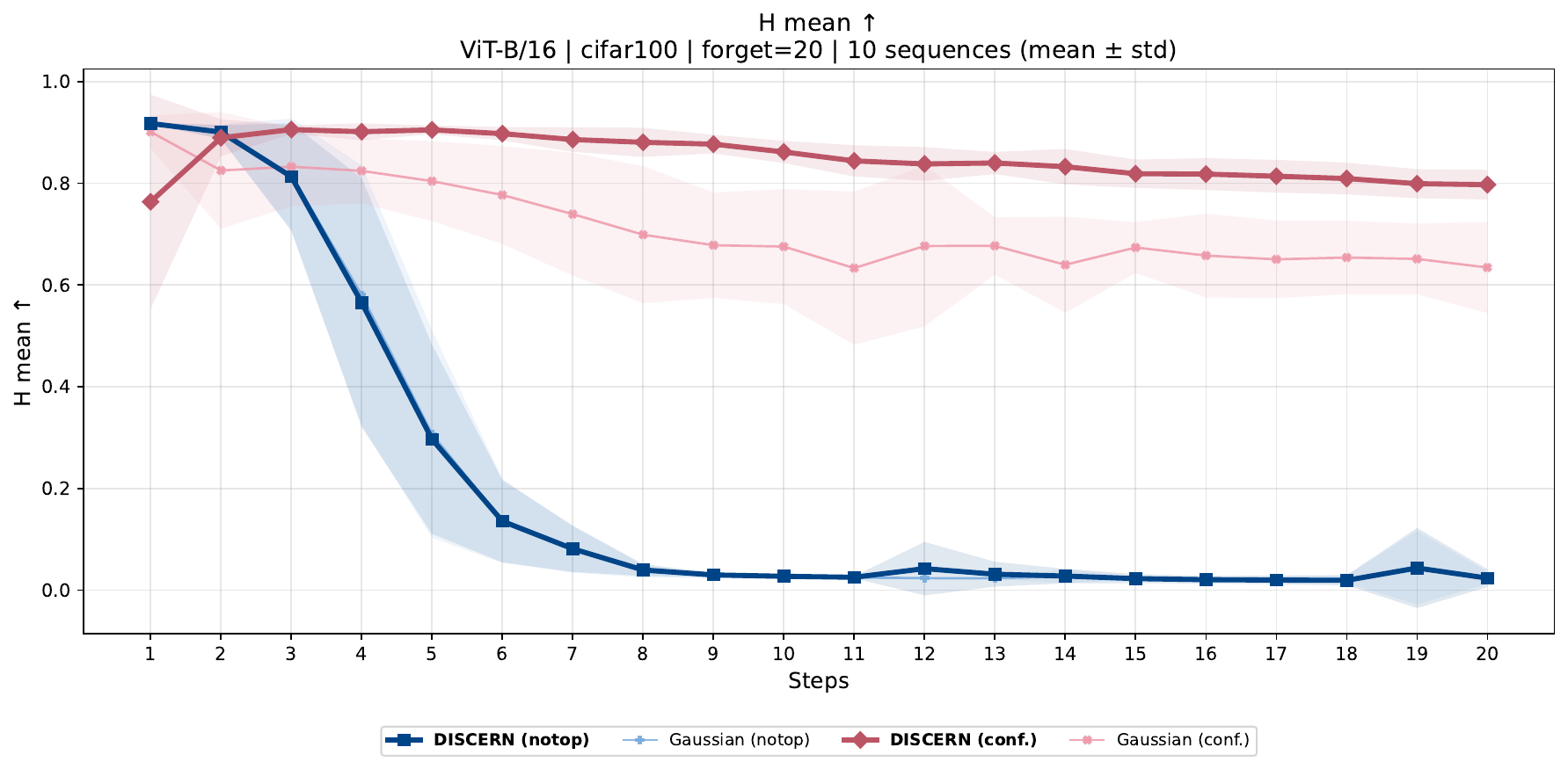}
        \caption{ViT-B/16}
    \end{subfigure}
    \hfill
    \begin{subfigure}[b]{0.48\linewidth}
        \centering
        \includegraphics[width=\linewidth]{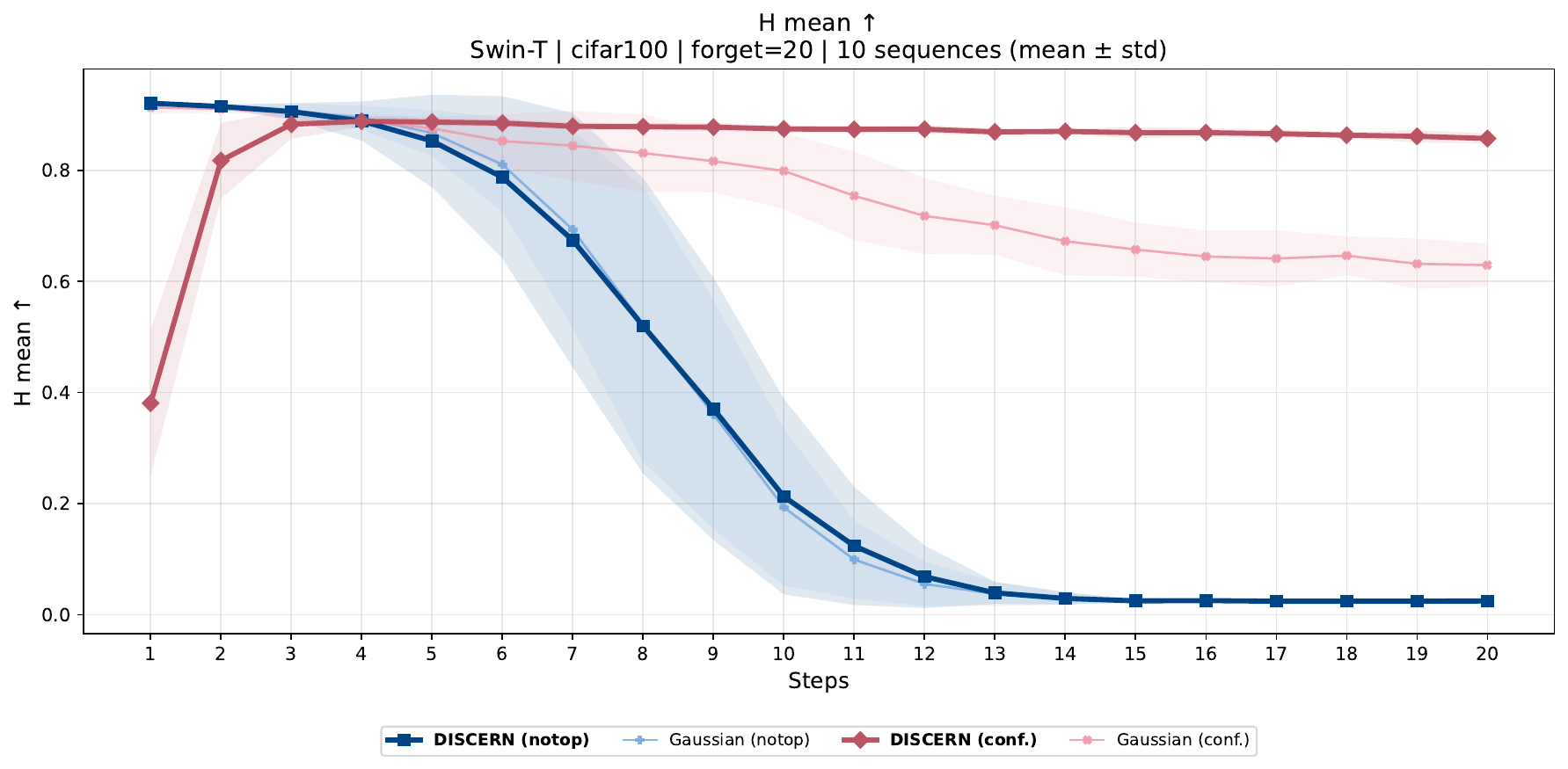}
        \caption{Swin-T}
    \end{subfigure}
    \\[1em]
    \begin{subfigure}[b]{0.48\linewidth}
        \centering
        \includegraphics[width=\linewidth]{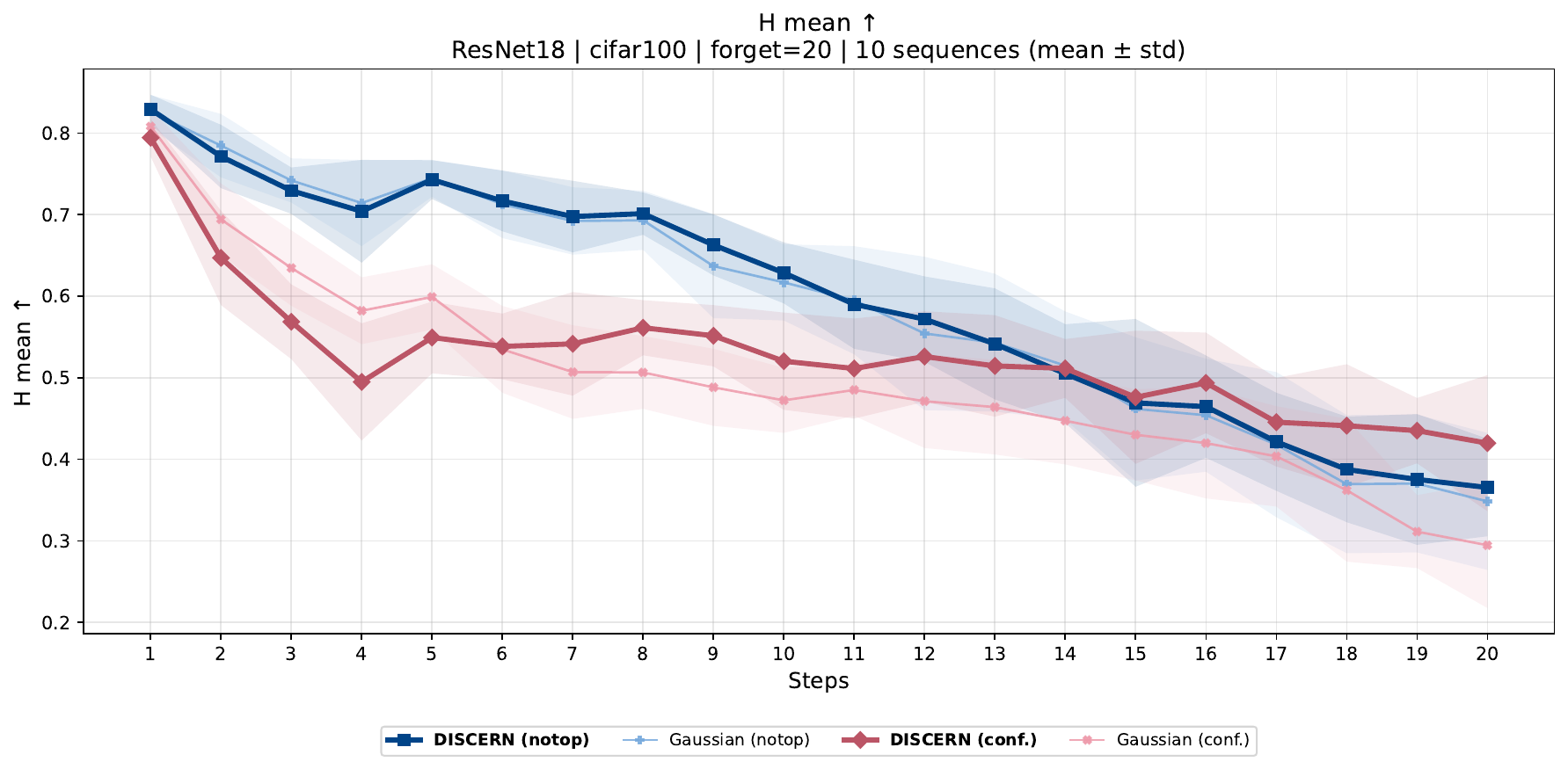}
        \caption{ResNet18}
    \end{subfigure}
    \hfill
    \begin{subfigure}[b]{0.48\linewidth}
        \centering
        \includegraphics[width=\linewidth]{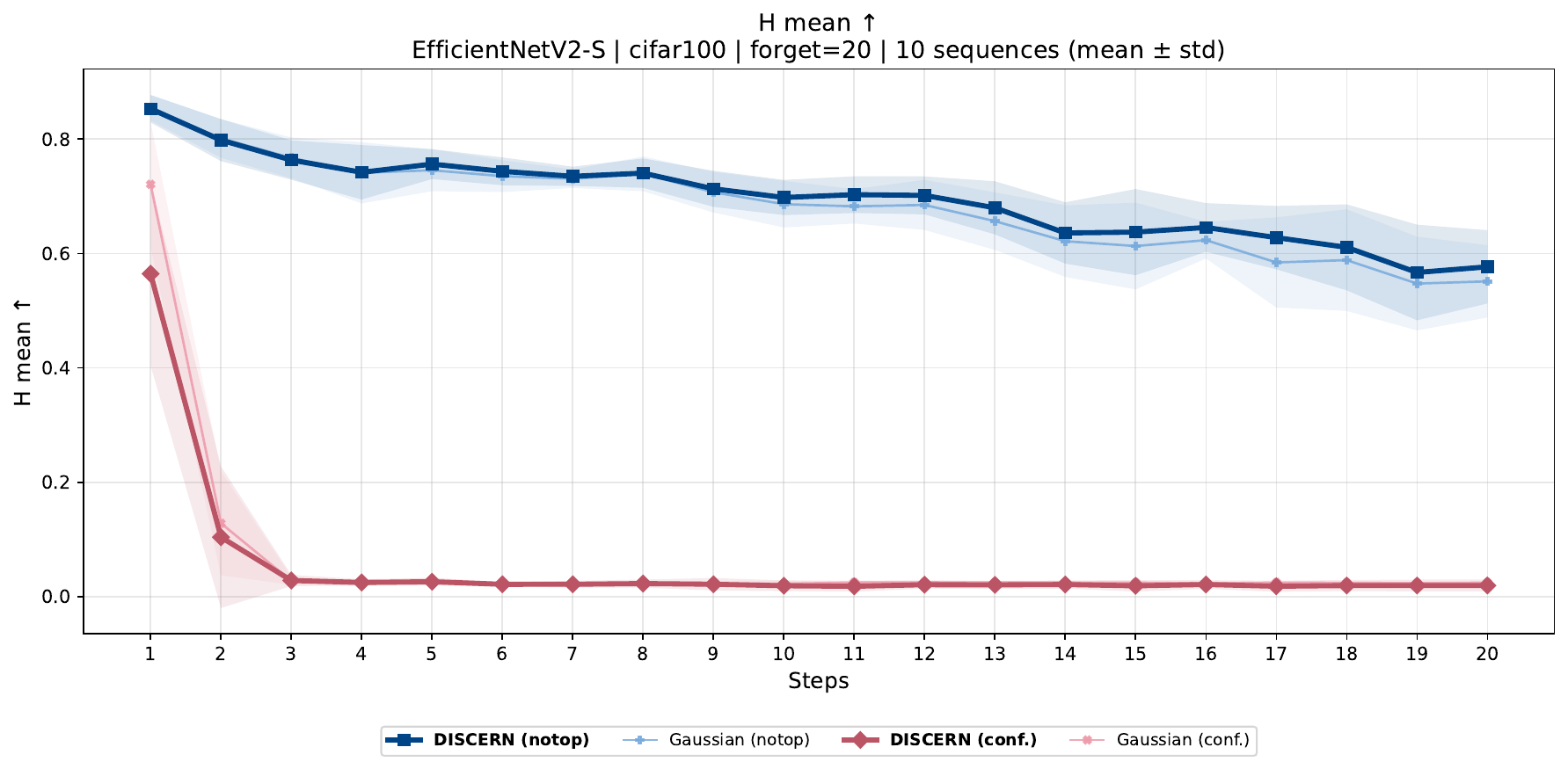}
        \caption{EfficientNetV2-S}
    \end{subfigure}
    \caption{Ablation against Gaussian in ICU (CIFAR-100, 20 sequential classes): mean and standard deviation of $H$ over the 10 sequences at each step, for DISCERN and Gaussian with both losses.}
    \label{fig:AblationNoisyICUH}
\end{figure}

\textbf{Mechanism.} Figure~\ref{fig:AblationNoisyICUH} explains why the sign depends on the scenario. At the first ICU step, which is a single-class removal, Gaussian obtains a higher $H$ with confidence on all four architectures, as in one-shot unlearning. DISCERN takes the lead from the second step on the two transformers and from the fifth on ResNet18, and the gap widens as steps accumulate. Since $Acc_f$ is measured on the cumulative set of forgotten classes, its growth indicates that already removed classes are recognized again, and it accounts for the decline in the $H$ of Gaussian. In aggregate terms, this resurfacing translates into a clearly lower $Acc_f$ for DISCERN in ICU ($\Delta Acc_f=-0.086$), the opposite of single-class unlearning ($\Delta Acc_f=+0.107$, Table~\ref{tab:AblationGlobal}).

The ablation therefore suggests that the main role of distribution selection is not to improve each deletion in isolation, but to limit the replay errors that an inadequate approximation of the geometry of the retained classes accumulates step by step. This holds not only on the transformers, where selection assigns the von Mises--Fisher distribution to every CIFAR-100 class, but also on ResNet18, where only 32 of those 100 classes receive a family other than the Gaussian.

\subsubsection{Test cascade versus cross-validation}\label{sec:ablationFlow}

The selection procedure (Figure~\ref{fig:flowchart}) resolves each class with a cascade of tests (Steps 1 and 2) and falls back on the cross-validation of Step 3 only when no test accepts it. To assess what this cascade contributes compared with a purely data-driven selection, we compare it with a variant, CV-only, that assigns every class through Step 3, with the same folds, tie rule, and margin. The rest of the method is identical, and the statistical design is that of Section~\ref{sec:ablationDist}.

\begin{table}[t]
    \centering
    \small
    \caption{Distribution selected per class by the CV-only variant, which assigns every class through the cross-validation of Step 3, without the test cascade. In parentheses, the number of classes assigned to each family out of the total. The von Mises--Fisher distribution is never selected. The families coincide with those of the full procedure (Table~\ref{tab:selectedDist}) in 105 of the 480 classes.}
    \label{tab:AblationFlowSelection}
    \begin{tabular}{l|c c c} \hline
        Architecture & CIFAR-10 & CIFAR-100 & MNIST \\ \hline
        ResNet18 & Student's $t$ (10/10) & \begin{tabular}[c]{@{}c@{}}Gaussian (74/100), $\mathcal{FB}$ (24/100)\\ Laplace (2/100)\end{tabular} & \begin{tabular}[c]{@{}c@{}}Student's $t$ (8/10)\\ Laplace (1/10), $\mathcal{FB}$ (1/10)\end{tabular} \\ \hline
        EfficientNetV2-S & Student's $t$ (10/10) & Gaussian (58/100), $\mathcal{FB}$ (42/100) & Student's $t$ (10/10) \\ \hline
        Swin-T & \begin{tabular}[c]{@{}c@{}}Laplace (8/10)\\ Student's $t$ (2/10)\end{tabular} & $\mathcal{FB}$ (75/100), Gaussian (25/100) & Laplace (10/10) \\ \hline
        ViT-B/16 & Student's $t$ (10/10) & $\mathcal{FB}$ (81/100), Gaussian (19/100) & \begin{tabular}[c]{@{}c@{}}Laplace (6/10)\\ Student's $t$ (4/10)\end{tabular} \\ \hline
    \end{tabular}
\end{table}

\textbf{Selected distributions.} Table~\ref{tab:AblationFlowSelection} reports how often each family is chosen without the test cascade, and should be read alongside Table~\ref{tab:selectedDist}. The selection changes completely: the two variants agree on 105 of the 480 classes, exactly those that the full procedure already resolved in Step 3, whereas every class decided by a test receives a different family. CV-only never selects the von Mises--Fisher distribution: its 299 classes move to the Fisher--Bingham (193) or the Gaussian (106). The 35 classes for which the Henze--Zirkler test does not reject normality almost all move to Student's $t$ (34), and the 41 Fisher--Bingham classes decided by a test move to the Laplace (25) or Student's $t$ (16). Overall, CV-only selects the Fisher--Bingham distribution for 223 classes, the Gaussian for 176, Student's $t$ for 54, and the Laplace for 27; that is, cross-validation favors more flexible families than those chosen by the cascade. By architecture, no class agrees on the transformers, whereas ResNet18 agrees on 94 of its 100 CIFAR-100 classes.

\begin{figure}[t]
    \centering
    \begin{subfigure}[b]{0.48\linewidth}
        \centering
        \includegraphics[width=\linewidth]{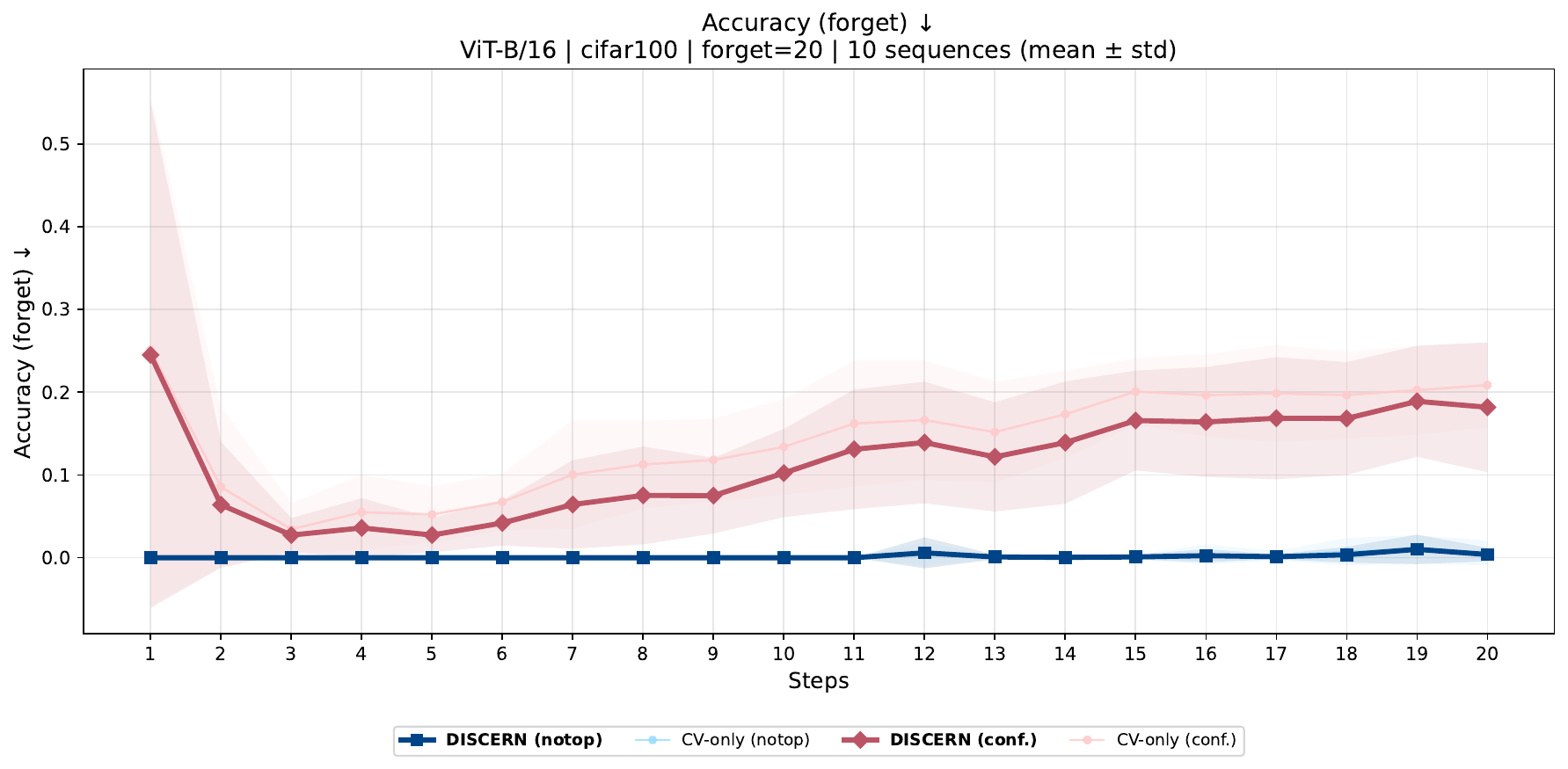}
        \caption{ViT-B/16}
    \end{subfigure}
    \hfill
    \begin{subfigure}[b]{0.48\linewidth}
        \centering
        \includegraphics[width=\linewidth]{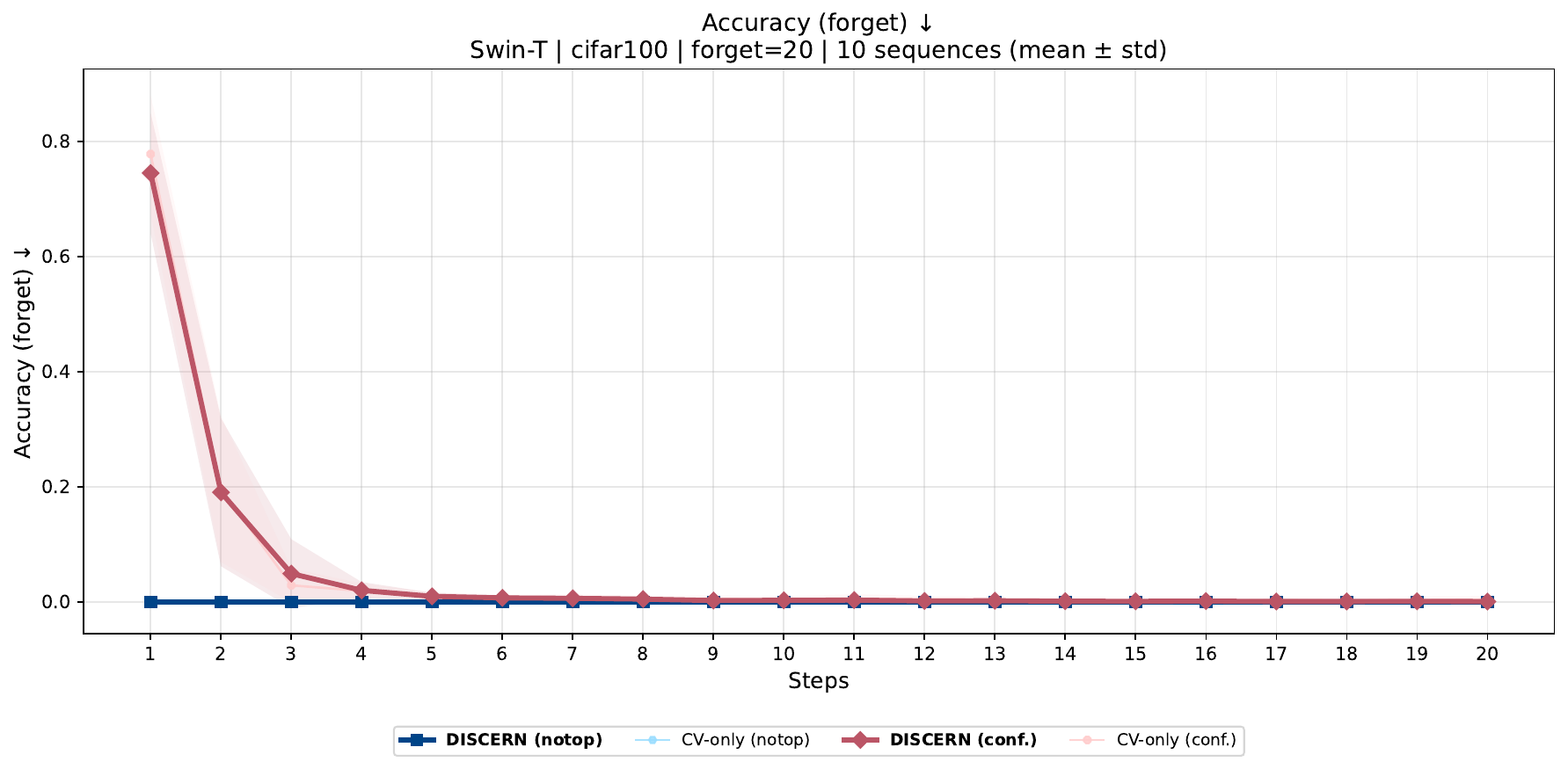}
        \caption{Swin-T}
    \end{subfigure}
    \\[1em]
    \begin{subfigure}[b]{0.48\linewidth}
        \centering
        \includegraphics[width=\linewidth]{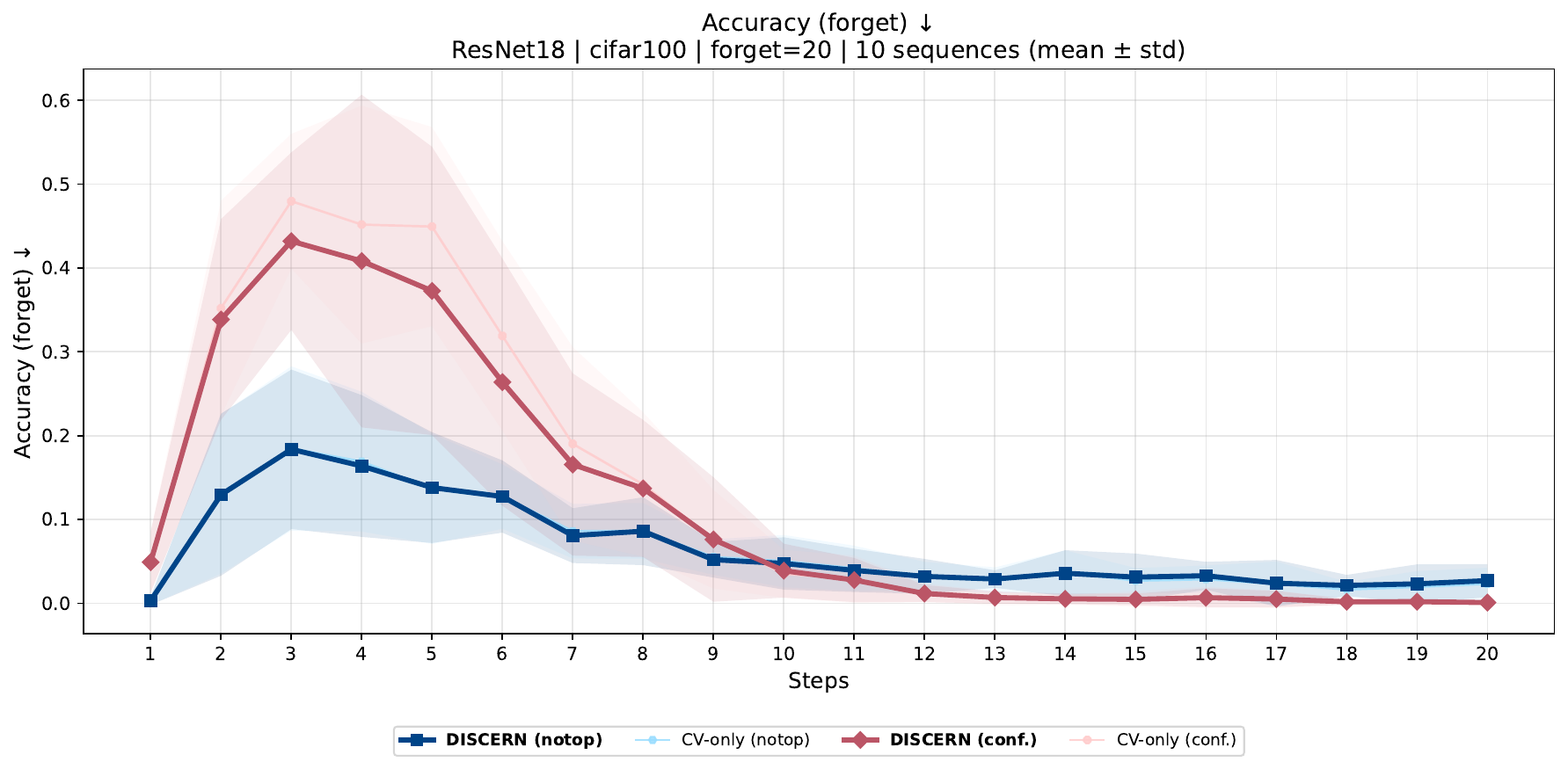}
        \caption{ResNet18}
    \end{subfigure}
    \hfill
    \begin{subfigure}[b]{0.48\linewidth}
        \centering
        \includegraphics[width=\linewidth]{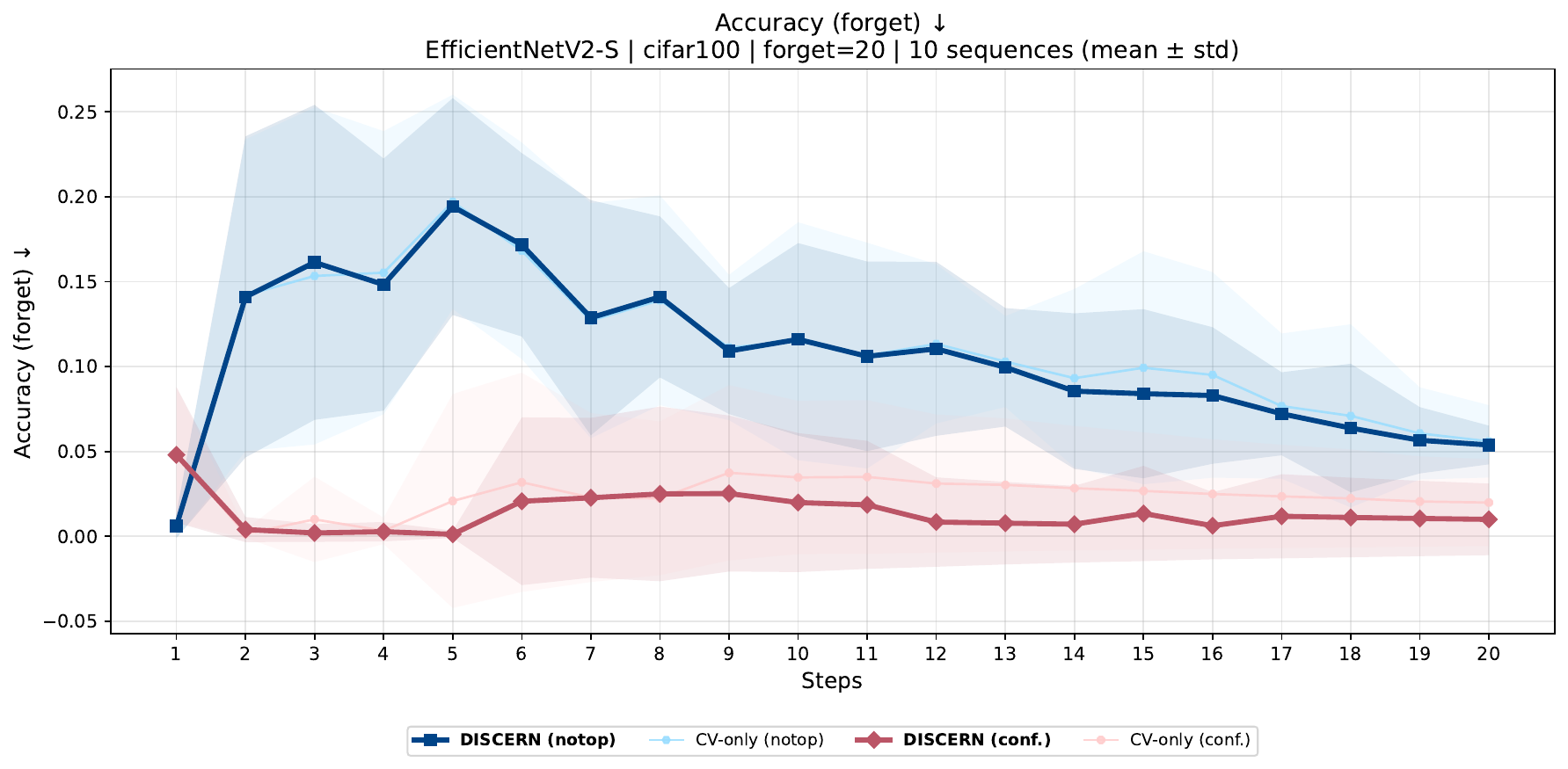}
        \caption{EfficientNetV2-S}
    \end{subfigure}
    \caption{Ablation of the selection flow in ICU (CIFAR-100, 20 sequential classes): mean and standard deviation of $Acc_f$ over the 10 sequences at each step, computed on the cumulative set of forgotten classes, for the full procedure (DISCERN) and CV-only with both losses.}
    \label{fig:AblationFlowICU}
\end{figure}

\textbf{Results.} No aggregate comparison is significantly in favor of CV-only (Table~\ref{tab:AblationGlobal}). With confidence, the full procedure is better in all three scenarios ($\Delta H=+0.014$, $+0.025$, and $+0.006$), mainly because it forgets more ($\Delta Acc_f$ between $-0.014$ and $-0.017$) with equal or slightly higher retention. With notop, the two variants tie in single- and multi-class unlearning, where $Acc_f$ is saturated near 0, and the full procedure wins slightly in ICU ($\Delta H=+0.006$), through $Acc_r$. By architecture (Table~\ref{tab:AblationArch}), the gain concentrates on ViT-B/16 and Swin-T, precisely where the two variants agree on no class, and on EfficientNetV2-S in multi-class unlearning with confidence ($+0.043$). On ResNet18, whose selection barely changes on CIFAR-100, no difference is significant. In ICU with confidence, Figure~\ref{fig:AblationFlowICU} shows that the difference is visible mainly in $Acc_f$: on ViT-B/16, the cumulative $Acc_f$ of CV-only stays above that of the full procedure for virtually the whole sequence, although with a much smaller gap than the one observed against Gaussian.

The differences are small, between $0.006$ and $0.025$ in $H$, but consistent. Since both ablations share the DISCERN runs, the two comparisons in Table~\ref{tab:AblationGlobal} are comparable, and subtracting their mean differences gives the mean difference between CV-only and Gaussian. In ICU with confidence, DISCERN outperforms Gaussian by $0.063$ in $H$ and CV-only by only $0.006$, so CV-only retains about $90\%$ of the advantage over Gaussian. In the scenario that motivates this work, the largest improvement therefore comes from not assuming normality for every class, and the test cascade adds a small but consistent improvement on top of it. Moreover, a higher cross-validated log-likelihood does not translate into better forgetting: the more parsimonious families chosen by the tests, in particular the von Mises--Fisher distribution, work slightly better than the more flexible alternatives preferred by cross-validation. The test cascade is thus supported by the results and, in addition, avoids repeated cross-validation for $78\%$ of the classes (375 of 480).

\subsection{Computational cost}\label{sec:runtime}

Table~\ref{tab:runtime} summarizes how long each method takes to serve a single-class unlearning request, averaged over the four architectures and the three datasets, together with the one-off cost of the prior intervention for the methods that require one.

\begin{table}[htb]
    \centering
    \small
    \caption{Computational cost in single-class unlearning: mean time per forgotten class, averaged over the four architectures and the three datasets, how many times faster each method is than retraining (ratio between the mean time of Retrain and that of the method), and one-off cost of the prior intervention, paid only once per architecture and dataset pair. $^{\dagger}$Median over the architecture and dataset pairs; the maximum, about three hours, corresponds to ResNet18 on CIFAR-100.}
    \label{tab:runtime}
    \begin{tabular}{l|rrr|l} \hline
        Method & s / class & min / class & Faster than Retrain & One-off cost \\ \hline
        EWC & 46 & 0.8 & 120$\times$ & MAS coefficients, $\sim$17 min \\
        RL & 54 & 0.9 & 102$\times$ & -- \\
        DELETE & 56 & 0.9 & 98$\times$ & -- \\
        DISCERN (notop) & 69 & 1.1 & 80$\times$ & distribution fitting, 4.6 min$^{\dagger}$ \\
        BT & 118 & 2.0 & 47$\times$ & -- \\
        LF-SSD & 123 & 2.1 & 45$\times$ & MAS coefficients, $\sim$17 min \\
        DISCERN (conf.) & 298 & 5.0 & 18$\times$ & same fitting as notop \\
        Expand & 397 & 6.6 & 14$\times$ & -- \\
        Shrink & 910 & 15.2 & 6$\times$ & -- \\
        \hline
        Retrain & 5504 & 91.7 & 1$\times$ & -- \\ \hline
    \end{tabular}
\end{table}

DISCERN with notop needs about 69 seconds per class, in the same group as the fastest methods in the comparison, EWC, Random Labeling, and DELETE (between 46 and 56 seconds), and it is 80 times faster than retraining, which takes an hour and a half per class on average. Sampling the synthetic embeddings adds almost no cost, since they are injected at the input of the classifier and do not go through the backbone. The confidence variant is slower, about 5 minutes per class, which is consistent with the slower onset of its forgetting described in Section~\ref{sec:ablationHyper}: the stopping criterion, which requires the validation accuracy on $\mathcal{D}_f$ to drop to chance level, is reached later. Even so, it is still 18 times faster than retraining, on par with Boundary Expand, and about three times faster than Boundary Shrink, which has to compute an adversarial attack for every sample.

Only three of the evaluated methods require a prior intervention: EWC, LF-SSD, and DISCERN (Table~\ref{tab:methods}). EWC and LF-SSD compute the MAS importance coefficients of the parameters, which takes about 17 minutes per architecture and dataset pair. The intervention of DISCERN, fitting one distribution per class, has a median of 4.6 minutes and is shared by both loss functions. In every case it is performed only once per model, before any deletion request, and its cost is amortized over all subsequent requests; in ICU, for instance, the same fitting serves the twenty steps of the sequence. Its duration depends mainly on how many classes the test cascade resolves: when it resolves all of them a few minutes suffice, whereas the most expensive case, about three hours, is ResNet18 on CIFAR-100, where 94 of the 100 classes require the cross-validation of Step~3. This cost reinforces the role of the test cascade, which avoids cross-validation for $78\%$ of the classes (Section~\ref{sec:ablationFlow}).

\subsection{Cohesion and separation in the latent space}\label{sec:clustering-metrics}

This section analyzes how embeddings behave under unlearning. To this end, we use two geometric measures on the latent space, intra-cluster cohesion and inter-cluster separation, computed with both the Euclidean distance (L2) and cosine similarity.

Each class is identified with a cluster, and clusters are grouped into three disjoint blocks:
\begin{itemize}
    \item Forget: the classes whose knowledge is to be removed.
    \item Retain: the classes the model must preserve.
    \item Unknown: classes from a near out-of-distribution (OOD) dataset. For CIFAR-10, CIFAR-100, and MNIST, we use CIFAR-100, CIFAR-10, and EMNIST-Letters, respectively, as near-OOD data.
\end{itemize}

Both quantities derive from the same mean pairwise distance statistic:

$$\rho(C_1, C_2) = \frac{1}{N_{C_1} N_{C_2}} \sum_{c_1 \in C_1} \sum_{c_2 \in C_2} d(c_1, c_2), \qquad \rho(C_1, C_1) = \frac{1}{N_{C_1} (N_{C_1}-1)} \sum_{c_1 \in C_1} \sum_{\substack{c_2 \in C_1\\ c_2\neq c_1}} d(c_1, c_2),$$

where $C_i$ denotes a class, $d$ the metric used, and $N_{C_i}$ the number of elements of $C_i$. Cohesion corresponds to the case $C_1 = C_2$ and separation to $C_1 \neq C_2$. When computing separation, we additionally require $C_2$ to be a Retain class, and for each class $C_1$ we keep only its nearest Retain class: the separation of $C_1$ is $\min_{C_2}\rho(C_1,C_2)$ in L2 and $\max_{C_2}\rho(C_1,C_2)$ in cosine, over the Retain classes $C_2\neq C_1$: the goal is to measure whether forgotten classes move closer to the knowledge the model retains or, conversely, away from it. Finally, cohesion and separation are averaged within each block, yielding six quantities per model: mean cohesion and separation in forget, retain, and unknown.

The metrics of each method are compared against two reference models: the original model and Retrain. For each triplet (architecture, dataset, forgotten class) and each method, we define an increment with respect to each reference:

$$\Delta_{L2} = \rho_{\text{method}} - \rho_{\text{reference}}, \qquad \Delta_{\cos} = \rho_{\text{reference}} - \rho_{\text{method}}.$$

The sign of the cosine term is inverted so that both metrics share the same orientation: $\Delta < 0$ indicates concentration and $\Delta > 0$ indicates dispersion. This yields four groups of increments per combination: L2 vs.\ original, L2 vs.\ retrain, cosine vs.\ original, and cosine vs.\ retrain.

We apply the one-sample Wilcoxon signed-rank test against 0 to the increments, with significance level $\alpha = 0.05$. Although the pooled observations come from different methods, each increment is already a paired difference between a method and its reference evaluated on exactly the same forgotten class, dataset, and architecture; the signed-rank test on the differences is therefore appropriate, rather than an unpaired Mann--Whitney test. Before testing, the increments of each cell (architecture $\times$ dataset $\times$ metric $\times$ reference) are cleaned by discarding those outside $[Q_1 - 1.5\,\mathrm{IQR},\ Q_3 + 1.5\,\mathrm{IQR}]$, computed over the increments of the seven methods in that cell; we refer to the remaining ones as cleaned increments. Zero increments are discarded by the test, and the exact null distribution is used when there are at most 50 nonzero increments and no ties, with the normal approximation otherwise.

The test is run in its two one-sided versions:
\begin{itemize}
    \item $H_0: \text{med}(\Delta) \geq 0$ versus $H_1: \text{med}(\Delta) < 0$ (methods concentrate);
    \item $H_0: \text{med}(\Delta) \leq 0$ versus $H_1: \text{med}(\Delta) > 0$ (methods disperse).
\end{itemize}
Here $\text{med}(\Delta)$ denotes the location parameter targeted by the signed-rank test, that is, the pseudomedian of the increments. The trend assigned to each cell is ``concentrates'' if $H_0$ is rejected only in the first version, ``disperses'' if only in the second, and ``no trend'' if it is rejected in neither. Since the two one-sided $p$-values cannot both fall below $\alpha$, this rule is equivalent to a two-sided test at level $2\alpha=0.10$ whose rejection is read according to the sign of the effect.

\begin{figure}
    \centering
    \includegraphics[width=\linewidth]{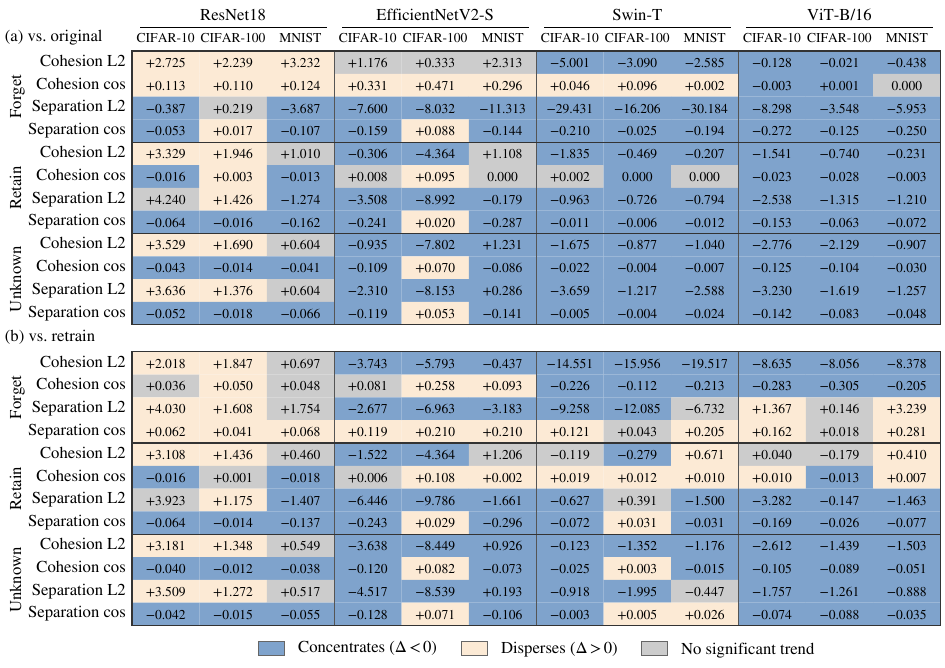}
    \caption{Wilcoxon test trend per architecture--dataset combination against (a) the original model and (b) the retrained model, for each class block (forget, retain, unknown), quantity (cohesion and separation), and metric (L2 and cosine). Color indicates the significant trend ($\alpha=0.05$) and each cell shows the median of the cleaned increments.}
    \label{fig:wilcoxheat}
\end{figure}

The test is performed at two levels, from highest to lowest resolution:
\begin{enumerate}
    \item By individual architecture: one test per architecture $\times$ dataset combination. This is the highest-resolution level and is reported in Figure~\ref{fig:wilcoxheat}.
    \item By architecture family: the two convolutional architectures (ResNet18, EfficientNetV2-S) are pooled on one side and the two transformers (Swin-T, ViT-B/16) on the other. This level is accompanied by an internal consistency check indicating whether the two architectures of the family individually agree in direction, contradict each other, or whether the result is driven by only one of them (Table~\ref{tab:wilcoxFamily}).
\end{enumerate}

The test treats each increment as an independent observation, which is debatable: the same forgotten class appears repeatedly across all methods, sharing the same reference. To check this, we additionally fitted a mixed model with a fixed intercept and a random effect per forgotten class. The intraclass correlation coefficient (ICC) was null or degenerate in 133 of the 144 configurations (three blocks $\times$ twelve combinations $\times$ four increment groups). The eleven exceptions are concentrated mostly in the transformers (eight of them) and almost entirely in the comparison against the retrained model (ten), mostly in cosine (ICC between $0.21$ and $0.82$). In the seven belonging to the forget block, the fixed intercept remains significant in the same direction as the Wilcoxon test ($p<10^{-4}$), so accounting for per-class dependence does not change the conclusion. The pooled test is therefore defensible, although in those cells the variability across forgotten classes is not negligible. 

Table~\ref{tab:wilcoxTrendSummary} and Figure~\ref{fig:wilcoxheat} summarize, for each class block, metric, and reference, how many of the twelve architecture $\times$ dataset combinations exhibit each trend.

\begin{table}[]
    \centering
    \small
    \caption{Summary of Wilcoxon test trends by class block, quantity (intra-cluster cohesion or inter-cluster separation), and metric, against each reference. Each cell indicates how many of the twelve architecture $\times$ dataset combinations concentrate, disperse, or show no significant trend ($\alpha=0.05$).}
    \label{tab:wilcoxTrendSummary}
    \begin{tabular}{ll c ccc ccc} \hline
        \multirow{2}{*}{Block} & \multirow{2}{*}{Quantity} & \multirow{2}{*}{Metric} & \multicolumn{3}{c}{vs. Original} & \multicolumn{3}{c}{vs. Retrain} \\ \cline{4-9}
         & & & Concentrates & Disperses & No trend & Concentrates & Disperses & No trend \\ \hline
        \multirow{4}{*}{Forget} & \multirow{2}{*}{Cohesion} & L2  & 6 & 3 & 3 & 9 & 2 & 1 \\
         & & Cos & 2 & 9 & 1 & 6 & 3 & 3 \\ \cline{2-9}
         & \multirow{2}{*}{Separation} & L2  & 11 & 0 & 1 & 5 & 4 & 3 \\
         & & Cos & 10 & 2 & 0 & 0 & 10 & 2 \\ \hline
        \multirow{4}{*}{Retain} & \multirow{2}{*}{Cohesion} & L2  & 8 & 2 & 2 & 3 & 4 & 5 \\
         & & Cos & 6 & 2 & 4 & 3 & 7 & 2 \\ \cline{2-9}
         & \multirow{2}{*}{Separation} & L2  & 10 & 1 & 1 & 9 & 1 & 2 \\
         & & Cos & 11 & 1 & 0 & 10 & 2 & 0 \\ \hline
        \multirow{4}{*}{Unknown} & \multirow{2}{*}{Cohesion} & L2  & 9 & 2 & 1 & 9 & 2 & 1 \\
         & & Cos & 11 & 1 & 0 & 10 & 2 & 0 \\ \cline{2-9}
         & \multirow{2}{*}{Separation} & L2  & 9 & 2 & 1 & 8 & 2 & 2 \\
         & & Cos & 11 & 1 & 0 & 9 & 3 & 0 \\ \hline
    \end{tabular}
\end{table}

The clearest result concerns the inter-cluster separation of forgotten classes relative to the original model: eleven of the twelve combinations show a significant decrease in L2, and ten of twelve in cosine. The only exception in L2, ResNet18 on CIFAR-100, shows no significant trend. Regardless of architecture and dataset, unlearning methods move each forgotten class closer to its nearest retained class. The pattern repeats in the retain (ten and eleven of twelve) and unknown (nine and eleven of twelve) blocks: the effect is not exclusive to the target classes; every class moves closer to its nearest retained neighbor, so the entire latent space contracts toward the region of retained knowledge.

Intra-cluster cohesion behaves much less homogeneously, and the global aggregation hides two opposite behaviors. Relative to the original model, forgotten-class clusters disperse in cosine in nine of twelve combinations, whereas in L2 the results split into six that concentrate, three that disperse, and three with no trend, so the aggregate allows no conclusion.

Relative to the retrained model, the picture changes in an interpretable way. Inter-cluster separation in cosine now increases in ten of twelve combinations, and in L2 there is no dominant direction: five combinations concentrate, four disperse, and three show no trend. The natural reading is that retraining brings the forgotten class even closer, in angular terms, to its nearest retained class: the approximate methods contract the latent space toward the retained knowledge, but less than the retrained model does.

The per-architecture breakdown in Figure~\ref{fig:wilcoxheat} reveals the main finding: the intra-cluster cohesion of forgotten classes behaves in opposite ways depending on the architecture, which explains the ambiguity of the global aggregate.

Relative to the original model, both convolutional architectures disperse the cluster of the forgotten class. ResNet18 does so in all six available cells (three datasets $\times$ two metrics), with medians from $+0.11$ to $+0.12$ in cosine and from $+2.24$ to $+3.23$ in L2. EfficientNetV2-S disperses in cosine on all three datasets, with the largest medians in the whole study ($+0.30$ to $+0.47$); its L2 medians are also positive ($+0.33$ to $+2.31$), although not significant. The two transformers, in contrast, concentrate in L2 on all three datasets: Swin-T with medians of $-5.001$, $-3.090$, and $-2.585$, and ViT-B/16 with $-0.128$, $-0.021$, and $-0.438$.

The two transformers are not equivalent, however. Swin-T disperses significantly in cosine on all three datasets, exactly like the two convolutional networks, whereas ViT-B/16 concentrates on two of the three; this is the only cell in which Swin-T systematically aligns with the convolutional family. Across the three class blocks, two quantities, and two metrics, Swin-T agrees with ViT-B/16 in $31$ of the $36$ cells and with both convolutional networks simultaneously in $14$, so it behaves mostly as a transformer with a deviation confined to the angular structure. A plausible explanation is its intermediate position between the two families (Section~\ref{sec:setup}): local window attention and a multiscale hierarchy give it convolution-like inductive biases that ViT-B/16 lacks. The cosine medians of both transformers are nevertheless small (between $-0.003$ and $+0.096$) and below all of those of the convolutional networks ($+0.11$ to $+0.47$): transformers barely alter the angular structure of the forgotten cluster, whereas convolutional networks alter it substantially. This interpretation should therefore be taken as a hypothesis suggested by the sign of the tests; the substantive architectural divergence lies in the L2 metric.

Relative to the retrained model, the divergence between architectures persists and grows in magnitude. The two transformers concentrate in L2 on all three datasets, with medians of $-14.6$, $-16.0$, and $-19.5$ for Swin-T and $-8.6$, $-8.1$, and $-8.4$ for ViT-B/16. EfficientNetV2-S concentrates on all three datasets ($-3.7$, $-5.8$, and $-0.44$), with values an order of magnitude smaller. ResNet18 disperses significantly on CIFAR-10 and CIFAR-100 ($+2.0$ and $+1.8$) and shows no trend on MNIST. In other words, the intra-cluster cohesion produced on transformers lies much further from that of the retrained model than on convolutional networks.

In the retain and unknown blocks, the hierarchy persists but weakens. ViT-B/16 concentrates systematically in all twenty-four cells of both blocks relative to the original model, making it the most consistent architecture in the study. Swin-T follows closely, except for cosine cohesion in the retain block. ResNet18 is again the exception: it disperses L2 intra-cluster cohesion in both retain and unknown, opposite to the other three architectures.

\begin{table}[]
    \centering
    \small
    \caption{Wilcoxon test trend for the forget block against the original model, pooling architectures by family (convolutional: ResNet18 and EfficientNetV2-S; transformer: Swin-T and ViT-B/16). $\downarrow$: concentrates; $\uparrow$: disperses; --: no significant trend ($\alpha = 0.05$). Each cell reports the $p$-value of the significant one-sided test. $^\dagger$: the trend is significant in only one of the two members of the family. $^\ddagger$: the two members show significant trends in opposite directions.}
    \label{tab:wilcoxFamily}
    \begin{tabular}{ll cccc} \hline
        \multirow{2}{*}{Family} & \multirow{2}{*}{Dataset} & \multicolumn{2}{c}{Cohesion} & \multicolumn{2}{c}{Separation} \\ \cline{3-6}
         & & L2 & Cos & L2 & Cos \\ \hline
        \multirow{3}{*}{Conv.} & CIFAR-10  & --$\,^\dagger$ & $\uparrow\ 7.9\cdot10^{-16}$ & $\downarrow\ 1.1\cdot10^{-15}$ & $\downarrow\ 7.6\cdot10^{-22}$ \\
         & CIFAR-100 & $\uparrow\ 1.2\cdot10^{-11}\,^\dagger$ & $\uparrow\ 8.0\cdot10^{-33}$ & $\downarrow\ 1.9\cdot10^{-19}\,^\dagger$ & $\uparrow\ 1.5\cdot10^{-18}$ \\
         & MNIST     & $\uparrow\ 4.3\cdot10^{-3}\,^\dagger$ & $\uparrow\ 3.4\cdot10^{-16}$ & $\downarrow\ 6.6\cdot10^{-18}$ & $\downarrow\ 3.3\cdot10^{-23}$ \\ \hline
        \multirow{3}{*}{Transf.} & CIFAR-10  & $\downarrow\ 3.6\cdot10^{-7}$ & --$\,^\ddagger$ & $\downarrow\ 9.4\cdot10^{-25}$ & $\downarrow\ 1.4\cdot10^{-22}$ \\
         & CIFAR-100 & $\downarrow\ 3.0\cdot10^{-2}$ & $\uparrow\ 8.4\cdot10^{-10}\,^\ddagger$ & $\downarrow\ 1.3\cdot10^{-41}$ & $\downarrow\ 9.0\cdot10^{-24}$ \\
         & MNIST     & $\downarrow\ 2.9\cdot10^{-8}$ & --$\,^\dagger$ & $\downarrow\ 6.3\cdot10^{-25}$ & $\downarrow\ 1.2\cdot10^{-22}$ \\ \hline
    \end{tabular}
\end{table}

Table~\ref{tab:wilcoxFamily} pools the architectures into families and confirms the above for the L2 metric: relative to the original model, the transformer family significantly concentrates the cohesion of the forgotten class on all three datasets ($p = 3.6\cdot10^{-7}$, $3.0\cdot10^{-2}$, and $2.9\cdot10^{-8}$), with internal agreement between its two members. The convolutional family, in contrast, significantly disperses on CIFAR-100 and MNIST ($p = 1.2\cdot10^{-11}$ and $4.3\cdot10^{-3}$) and shows no trend on CIFAR-10; on all three datasets L2 dispersion is significant only on ResNet18, whereas EfficientNetV2-S shows positive but non-significant medians. In cosine, by contrast, both of its members disperse on all three. For L2 inter-cluster separation, both families concentrate on all three datasets, and in cosine they agree except on CIFAR-100. The architecture therefore determines how the class to be forgotten is internally reorganized, not how the latent space as a whole is repositioned.

For cosine cohesion relative to the original model, the internal consistency check flags the transformer-family cells as contradictory ($^\ddagger$ in Table~\ref{tab:wilcoxFamily}), owing to the Swin-T/ViT-B/16 discrepancy described above, so the pooled value of that cell should not be read as a family property.

\subsubsection*{Behavior of DISCERN}

\begin{figure}
    \centering
    \includegraphics[width=\linewidth]{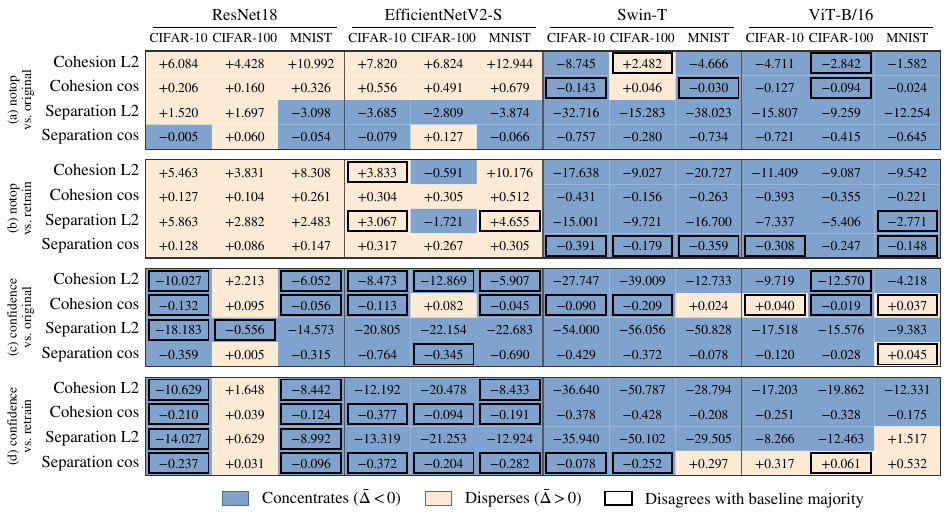}
    \caption{Mean cleaned increment of DISCERN in single-class unlearning for the forget block: (a) notop against the original model, (b) notop against the retrained model, (c) confidence against the original model, and (d) confidence against the retrained model. Columns are organized as in Figure~\ref{fig:wilcoxheat}. Color indicates whether the method concentrates or disperses, and a black border marks the cells in which it disagrees with the majority of the seven baseline methods.}
    \label{fig:distribForget}
\end{figure}

All of the analysis above pools the seven baseline methods and leaves the proposed method out. To position it relative to them, for each cell (architecture $\times$ dataset $\times$ quantity $\times$ metric $\times$ reference) we compute the mean increment of DISCERN under its two losses, notop and confidence, after discarding its own increments outside $1.5\,\mathrm{IQR}$ in that cell, and check whether its direction matches the majority direction of the seven methods in that same cell, that is, the sign shared by most of their mean cleaned increments. The question this answers is not whether the proposal concentrates or disperses the latent space, but whether it does so in the same direction as the other methods or imposes a geometry of its own. DISCERN covers the four architectures and the three class blocks. In multi-class unlearning only CIFAR-100 is available. Table~\ref{tab:refAgreement} summarizes the agreement rates, and Figure~\ref{fig:distribForget} breaks them down cell by cell for the forget block, together with the magnitudes. Cells in which the seven methods are tied have no majority and are excluded from the counts.

\begin{table}[]
    \centering
    \small
    \caption{Agreement of each variant of DISCERN with the majority direction of the seven baseline methods, by class block and quantity. Each cell pools the two metrics (L2 and cosine), the three datasets, and the two references (original and retrained model), so the denominator is at most 12; cells in which the baseline methods are tied are excluded.}
    \label{tab:refAgreement}
    \begin{tabular}{ll cc cc cc} \hline
        \multirow{2}{*}{Loss} & \multirow{2}{*}{Architecture} & \multicolumn{2}{c}{Forget} & \multicolumn{2}{c}{Retain} & \multicolumn{2}{c}{Unknown} \\ \cline{3-8}
         & & Coh. & Sep. & Coh. & Sep. & Coh. & Sep. \\ \hline
        \multirow{4}{*}{notop} & ResNet18 & 12/12 & 12/12 & 10/12 & 12/12 & 10/12 & 10/12 \\
         & EfficientNetV2-S & 11/12 & 10/12 & 8/9 & 12/12 & 8/12 & 12/12 \\
         & Swin-T & 9/12 & 9/12 & 9/11 & 12/12 & 11/12 & 9/12 \\
         & ViT-B/16 & 10/12 & 9/12 & 11/12 & 12/12 & 12/12 & 12/12 \\ \hline
        \multirow{4}{*}{confidence} & ResNet18 & 4/12 & 6/12 & 7/12 & 10/12 & 8/12 & 8/12 \\
         & EfficientNetV2-S & 3/12 & 8/12 & 4/9 & 8/12 & 8/12 & 8/12 \\
         & Swin-T & 10/12 & 10/12 & 7/11 & 9/12 & 7/12 & 8/12 \\
         & ViT-B/16 & 8/12 & 10/12 & 10/12 & 12/12 & 11/12 & 10/12 \\ \hline
    \end{tabular}
\end{table}

The notop variant follows the general trend: it agrees with the majority in 82 of the 96 cells with a defined majority in the forget block, 86 of 92 in retain, and 84 of 96 in unknown. Agreement never falls below $67\%$ in any of the twenty-four cells of Table~\ref{tab:refAgreement}, and it reaches $100\%$ in nine of them. More important than the counts is that it reproduces the architectural divergence described above: it disperses the cohesion of the forgotten class on the two convolutional networks (between $+4.4$ and $+12.9$ in L2 against the original) and concentrates it on the two transformers (Swin-T, $-8.8$ and $-4.7$, with the exception of CIFAR-100, $+2.5$; ViT-B/16, between $-1.6$ and $-4.7$). No cell falls below two thirds; the worst is cohesion in the unknown block on EfficientNetV2-S ($8/12$).

The confidence variant is the one that departs from consensus, and it does so in a structured way. For the cohesion of the forgotten class, agreement drops to $33\%$ on ResNet18 ($4/12$) and $25\%$ on EfficientNetV2-S ($3/12$), against $83\%$ ($10/12$) on Swin-T and $67\%$ ($8/12$) on ViT-B/16. The reason is that it concentrates that cohesion in L2 in eleven of the twelve combinations regardless of the architecture, the only exception being ResNet18 on CIFAR-100. It also does so with magnitudes far above those of the majority: against the original, between $-5.9$ and $-12.9$ on the convolutional networks and between $-12.7$ and $-39.0$ on Swin-T, whereas the median of the seven methods on Swin-T does not go beyond $-5.0$. The same pattern appears in L2 inter-cluster separation, which decreases in all twelve combinations against the original, with drops of up to $-56.1$. A plausible explanation is that the confidence penalty pushes the predictions of the forgotten class toward the uniform distribution, collapsing its \textit{embeddings} into a compact region of maximum uncertainty located among the retained classes. Confidence thus imposes its own geometry, superimposed on the architectural bias. This effect is nevertheless confined to the forgotten class: in the retain and unknown blocks agreement largely recovers, between $44\%$ and $100\%$ with a single cell below one half, and for separation in the retain block ResNet18 and ViT-B/16 reach $10/12$ or more.

Relative to the retrained model, neither variant reproduces its geometry. For the L2 cohesion of the forgotten class, notop disperses more than retraining on the convolutional networks (between $+3.8$ and $+10.2$, the only exception being EfficientNetV2-S on CIFAR-100, $-0.6$) and concentrates more than it on the transformers (between $-9.0$ and $-20.7$). Confidence concentrates more than retraining in eleven of the twelve combinations, with means between $-8.4$ and $-50.8$. DISCERN therefore does not approach the latent-space organization of the retrained model any more closely than the approximate baseline methods: with notop it falls within their typical behavior, and with confidence it exaggerates it in the concentrating direction. Its good accuracy is thus not explained by reproducing the geometry of the retrained model.

In multi-class unlearning the disagreement of confidence drops markedly: it matches the majority in 26 of the 32 cells, $81\%$ against $61\%$ in the forget block of single-class unlearning, whereas notop stays at a level similar to single-class unlearning (27 of 32). The reason is that in this scenario most baseline methods already concentrate the forgotten class in L2 and bring it closer to the retained ones, which is precisely what confidence does systematically.

This geometric analysis should be read as complementary to the functional evaluation of unlearning: its aim is not to propose a new success metric, but to characterize how different methods reorganize internal representations. The main finding is a systematic dependence on the architecture. On the convolutional networks evaluated, the forgotten class tends to disperse in Euclidean distance (on ResNet-18, with medians between $+2.24$ and $+3.23$ in L2 relative to the original model), whereas on the transformers it tends to concentrate, with both members agreeing on all three datasets. DISCERN with notop largely reproduces this architectural trend, whereas the confidence variant imposes its own concentration dynamics, present in eleven of the twelve combinations regardless of the architecture. These results show that methods with similar behavior in terms of accuracy can reach it through substantially different latent reorganizations.

\section{Conclusions}\label{sec:conclusions}

We have presented DISCERN, a retain-free, intervention-based MU method that stands out in ICU. The results show that, on transformer architectures, it achieves good performance compared with the other methods evaluated, while remaining robust and competitive in single-class and simultaneous multi-class unlearning. Since the same group of classes is removed at once in the multi-class scenario and step by step in ICU, comparing both scenarios isolates the effect of the deletion schedule: several of the strongest methods when classes are removed simultaneously degrade sharply when the same deletion is split into successive steps, whereas DISCERN barely changes. This indicates that the main obstacle for these methods is not the amount of knowledge to be forgotten, but the way in which it is forgotten.

Our ablation experiments show that the choice of distribution is relevant, as opposed to directly assuming that every class follows a Gaussian. In ICU, fitting each class with its most suitable family reduces the reappearance of previously forgotten classes along the sequence, so distribution selection acts as a safeguard of accumulated forgetting rather than as a uniform improvement of each isolated deletion. A second ablation, which replaces the test cascade with a selection based solely on cross-validation, separates the two sources of improvement. In ICU, the cross-validation-only selection retains most of the advantage over Gaussian, so the largest improvement comes from not assuming normality for every class. The proposed test cascade adds a small but consistent improvement, mainly with the confidence loss on the transformers, and also avoids cross-validation for most classes.

The method has three main limitations. The first is inherent to its design: DISCERN requires an intervention phase before any deletion request, in which the distribution of each class is estimated from the embeddings of the original model. If a model has been deployed without this phase and its training data are no longer available, the method cannot be applied. The second concerns the confidence penalty, the forgetting loss used in ICU, which shows a high variance across runs on some architectures. The third is that the synthetic embeddings are injected at the input of the classifier, so they do not constrain the last stage of the backbone, which is updated only by the forgetting loss. Since the distributions are fitted once on the original model and reused at every step, they become increasingly outdated as that stage drifts along the ICU sequence.

As future work, we plan to evaluate DISCERN on a wider range of architectures, to extend it to tasks beyond image classification, and to study its application to concept removal in diffusion models, the other main IU scenario discussed in the introduction. We also plan to update the class distributions along the sequence, or to inject the synthetic samples before the trainable stage of the backbone, so that the replay keeps up with the drift of the representation.

Overall, DISCERN offers a practical answer to the setting that motivates this work. It needs no access to the retained data at deletion time: a single intervention on the original model is enough to serve every subsequent request. Its per-class selection adapts the replay to the latent geometry of each backbone, which proves decisive precisely on the transformer architectures that increasingly underpin vision models, and the test cascade resolves most classes without cross-validation. On these architectures, DISCERN achieves the best $H$ after twenty ICU steps and is the method whose performance changes least when the deletion of twenty classes is split into twenty successive requests, while remaining competitive when classes are removed at once. Taken together, these properties make DISCERN particularly well suited to deployed models that must honor a continuing stream of deletion requests without keeping the data they were trained on.

\section*{Acknowledgments}

This publication is part of the grant PID2023-150070NB-I00, funded by MCIU/AEI/10.13039/501100011033 and by ESF+. This research is also part of the project TSI-100927-2023-1, funded by the Recovery, Transformation and Resilience Plan of the European Union--NextGenerationEU through the Spanish Ministry for Digital Transformation and the Civil Service.

\printbibliography

\end{document}